\documentclass{article} % For LaTeX2e
\usepackage{iclr2026_conference,times}

\usepackage{amsmath,amsfonts,bm}

\usepackage[utf8]{inputenc} % allow utf-8 input
\usepackage[T1]{fontenc}    % use 8-bit T1 fonts
\usepackage{hyperref}       % hyperlinks
\usepackage{url}            % simple URL typesetting
\usepackage{booktabs}       % professional-quality tables
\usepackage{nicefrac}       % compact symbols for 1/2, etc.
\usepackage{microtype}      % microtypography
\usepackage{wrapfig}
\usepackage{enumitem}
\usepackage{CJKutf8}
\usepackage{makecell}
\usepackage{tabularx}
\usepackage{colortbl}
\usepackage{amssymb}
\usepackage{subfigure}
\usepackage{multirow}
\usepackage{pifont}
\usepackage{arydshln}
\usepackage{fontawesome5}
\usepackage{tcolorbox}
\tcbuselibrary{breakable,skins}
\usepackage{algorithm}
\usepackage{algpseudocode}
\usepackage[table]{xcolor} % 加载用于颜色的宏包
\usepackage{ulem} % 加载用于下划线的宏包
\usepackage{tikz} % 用于复杂操作

\definecolor{mlb}{RGB}{173,216,230}  % 浅蓝色
\definecolor{mlo}{RGB}{255,223,186}  % 浅橙色

\def\eqref#1{equation~\ref{#1}}
\def\1{\bm{1}}

\DeclareMathAlphabet{\mathsfit}{\encodingdefault}{\sfdefault}{m}{sl}
\SetMathAlphabet{\mathsfit}{bold}{\encodingdefault}{\sfdefault}{bx}{n}

\title{From Utterance to Vividity: Training Expressive Subtitle Translation LLM via Adaptive Local Preference Optimization}

\author{
Chaoqun Cui$^{1,2}$,~Shijing Wang$^3$,~Liangbin Huang$^3$,~Qingqing Gu$^4$,~Zhaolong Huang$^3$,\\
~\textbf{Xiao Zeng}$^3$,~\textbf{Wenji Mao}$^{1,2,*}$\\
$^1$MAIS, Institute of Automation, Chinese Academy of Sciences\\
$^2$School of Artificial Intelligence, University of Chinese Academy of Sciences\\
$^3$Hujing Digital Media \& Entertainment Group\\
$^4$Geely AI lab\\
\texttt{cuichaoqun2025@ia.ac.cn, wenji.mao@ia.ac.cn}
}

\iclrfinalcopy % Uncomment for camera-ready version, but NOT for submission.
\begin{document}

\maketitle

\begingroup
\renewcommand\thefootnote{} % 去掉注脚的编号
% \footnotetext{$^\dagger$ Equal contribution.}
\footnotetext{$^*$ Corresponding author.}
\endgroup

\begin{abstract}

The rapid development of Large Language Models (LLMs) has significantly enhanced the general capabilities of machine translation. However, as application scenarios become more complex, the limitations of LLMs in vertical domain translations are gradually becoming apparent. In this study, we focus on how to construct translation LLMs that meet the needs of domain customization. We take visual media subtitle translation as our topic and explore how to train expressive and vivid translation LLMs. We investigated the situations of subtitle translation and other domains of literal and liberal translation, verifying the reliability of LLM as reward model and evaluator for translation. Additionally, to train an expressive translation LLM, we constructed and released a multidirectional subtitle parallel corpus dataset and proposed the Adaptive Local Preference Optimization (ALPO) method to address fine-grained preference alignment. Experimental results demonstrate that ALPO achieves outstanding performance in multidimensional evaluation of translation quality.

% Large Language Models (LLMs)的快速发展显著提升了机器翻译的通用能力，然而，随着应用场景的深入，LLM在垂直领域翻译中暴露的局限性逐渐显现。In this study，我们关注如何搭建满足领域定制化需求的翻译LLM。我们以visual media subtitle translation为课题，探索如何训练expressive和vivid的翻译LLM。我们调研了subtitle translation和其他领域的literal translation和liberal translation情况，并验证了14B模型作为翻译的reward model和evaluator的可靠性。另外，为了训练一个expressive的翻译LLM，我们构建并开源了一个多方向字幕平行语料数据集，并提出了Adaptive Local Preference Optimization (ALPO)方法以处理细粒度的偏好对齐问题。实验结果验证了ALPO在多维度的翻译质量评估中取得了卓越性能。

\end{abstract}

\section{Introduction}

In recent years, the rapid advancement of Large Language Models (LLMs) has significantly enhanced the general capabilities of machine translation \citep{cpo,sr,ladder}. Representative LLMs such as GPT-4 \citep{gpt4} and Qwen3 \citep{qwen3}, trained on massive cross-lingual data, have demonstrated exceptional contextual comprehension and generation abilities in multilingual translation tasks. Their translation quality in general domains (e.g., news, daily conversations) has approached human-level performance \citep{paradigm,xalma}. However, as applications penetrate vertical domains, the limitations of LLMs in specialized translation scenarios have become increasingly apparent: critical issues include inadequate consistency in specialized terminology \citep{mtchallenge1,mtchallenge2}, deviations from industry-standard expressions \citep{mtchallenge4,mtchallenge3}, and weak style adaptability \citep{mtchallenge5}. Consequently, LLM-based machine translation research has increasingly focused on addressing domain-specific customization requirements, such as developing legislation-focused LLMs that strictly adhere to clause formulation norms or medicine-oriented LLMs that precisely align with medical terminology.

% 近年来，Large Language Models (LLMs)的快速发展显著提升了机器翻译的通用能力\citep{cpo,sr,ladder}。以GPT-4 \citep{gpt4} and Qwen3 \citep{qwen3}等为代表的LLM凭借海量跨语言数据训练，在多语种翻译任务中展现出优异的Context理解与生成能力，尤其在通用领域（如新闻、日常对话）的翻译质量已接近人类水平\citep{paradigm,xalma}。然而，随着应用场景的深入，LLM在垂直领域翻译中暴露的局限性逐渐显现：专业术语一致性不足\citep{mtchallenge1,mtchallenge2}、行业规范表达偏差\citep{mtchallenge4,mtchallenge3}、风格适配性弱\citep{mtchallenge4,mtchallenge5}等问题突出。因此，基于LLM的机器翻译研究更加关注如何满足领域独有的定制化需求，例如搭建严格遵循条款表述规范的legislation翻译LLM、精准匹配医学术语的medicine翻译LLM等。

In this study, we focus on visual media subtitle translation task, which aims to translate the lines in subtitles of visual media programs across genres such as movies, TV series, and documentaries from the source language into the target language. This task plays a crucial role in the localization of entertainment programs and in fostering global cultural dissemination and exchange \citep{vd1,vd2,vd3}, yet it remains an underexplored subfield in machine translation. Similar to literary translation, subtitle translation requires a localized and stylistic \textbf{liberal translation} of program lines to convey the atmosphere, emotions, and tone of the original lines. However, while LLMs can achieve high translation accuracy, they tend to favor \textbf{literal translation} (as we quantitatively examined in Section~\ref{sec:domain}). Therefore, exploring how to train a customized subtitle translation LLM with strong expressiveness and vividness is our primary challenge.

% In this study, 我们focus on visual media subtitle translation task，which aims to将movies, TV series, and documentarie等体裁的visual media program字幕中的角色dialogue从source语言翻译为target语言，是entertainment program本地化和世界文化传播与交流的重要环节\citep{vd1,vd2,vd3}，但也是目前机器翻译中鲜有研究的子领域。Subtitle translation类似于literature翻译，要求对program中的台词进行本地化和风格化的\textbf{liberal translation}以传达原台词的氛围、情感、语气等风格信息。然而尽管LLM能够达到很高的翻译accuracy但是其更偏向于\textbf{literal translation}（我们在Section~\ref{sec:domain}中对此进行了量化的实验调研）。因而，exploring how to train a customized subtitle translation LLM with high expressiveness and vividness是我们要面临的首要挑战。

We employ LLM-as-a-Judge (using LLMs as evaluators) and preference optimization techniques to build a customized subtitle translation LLM. We conduct a quantitative investigation into the degree of liberal translation in human-translated texts across different domains and the liberal translation performance of various LLMs in subtitle translation. The findings reveal that: (1) Compared to the translation of serious texts requiring high accuracy (literal translation), such as legislation, news, and medicine, subtitle translation tends to favor more liberal translation; (2) In subtitle translation, the translations generated by chat LLMs tend to be more literal compared to those produced by reason LLMs and humans. We also experimentally validate the reliability of LLMs as reward models and evaluators for subtitle translation, as well as their alignment with human preferences. This enables automated construction of preference data for preference alignment training, thereby enhancing the expressiveness and vividness of subtitle translation LLMs. Additionally, for the task of translating short subtitle lines, which requires fine-grained local preference alignment, we propose a novel preference alignment strategy called \textbf{A}daptive \textbf{L}ocal \textbf{P}reference \textbf{O}ptimization (ALPO).

% 我们利用LLM-as-a-Judge (the concept of using LLMs as evaluators) 及preference optimization技术来搭建定制化的subtitle translation LLM。我们对不同domain人工翻译的liberal translation程度和不同LLM在subtitle translation上的liberal translation表现做了量化调研，结果显示：(1) subtitle translation相比legislation、news、medicine等要求高accuracy（literal translation）的严肃文本翻译，其更偏向liberal translation；（2）在subtitle translation domain中，chat LLMs的译文相比reason LLMs和human的译文更偏literal translation。我们也通过实验验证了LLM作为subtitle translation的reward model和evaluator的reliability及其与人类偏好的一致性，并依此自动化地构建偏好数据用于LLM的偏好对齐训练以提升subtitle translation LLM的expressiveness和vividness。另外，针对subtitle短句翻译这种需要细粒度地进行local偏好对齐的任务，我们提出了一种novel的偏好对齐策略，称为\textbf{A}daptive \textbf{L}ocal \textbf{P}reference \textbf{O}ptimization (ALPO)。

In summary, the main contributions of this study are as follows:
\begin{itemize}
\item We introduce the visual media subtitle translation task for the first time and investigate the extent of liberal translation and the reliability of LLM-as-a-Judge in this domain.
\item We propose ALPO, a local preference optimization method, provide a formal explanation of its effectiveness, and use it to train a subtitle translation LLM with high vividness.
\item We release subtitle parallel corpora in multiple directions to support community research, and established a multidimensional evaluation framework based on LLM-as-a-Judge.
\item Experimental results demonstrate that our 14B LLM, trained with ALPO, achieves significant improvements across multiple dimensions and outperforms SOTA LLMs.
\end{itemize}

% 总结来说，this study的主要贡献如下。
% \begin{itemize}
% \item 我们第一次提出了visual media subtitle translation task，并对该领域的liberal translation程度和LLM-as-a-Judge的reliability进行了investigation。
% \item 我们提出了ALPO这一local偏好优化方法并形式化地解释了其有效性，并据此训练了高vividness的subtitle translation LLM。
% \item 我们开源了多方向的subtitle parallel corpora以促进社区研究，并基于LLM-as-a-Judge搭建了subtitle translation多维度评估体系。
% \item 实验结果表明经过ALPO训练的14B参数LLM取得了多维度的显著提升，超越了SOTA LLMs。
% \end{itemize}

\section{Related Work}

\subsection{LLM-as-a-Judge}

The capacity of LLMs to emulate human reasoning and evaluate specific inputs against a set of predefined rules has paved the way for "LLM-as-a-Judge" \citep{judge1,judge2}. Existing studies demonstrate that LLMs' scalability, adaptability, and cost-effectiveness render them particularly suitable for handling increasing volumes of assessment tasks that humans conventionally performed \citep{judge3,judge4,judge5,judge6}. These capabilities are crucial for deploying LLMs flexibly across diverse evaluation scenarios and objectives, driving their rapid adoption in practical evaluation scenarios \citep{judge7,judge8,judge9}.

Originally developed for language generation and comprehension tasks, LLMs have evolved significantly through advanced training methodologies such as Reinforcement Learning from Human Feedback (RLHF) \citep{po1}, enhancing their alignment with human values and reasoning processes. This alignment has allowed LLMs to transition from generative tasks to evaluative roles. Fundamentally, LLM-as-a-Judge refers to the deployment of LLMs to assess objects, actions, or decisions according to predefined rules, criteria, or preferences \citep{judge10,judge11}. This framework encompasses a wide range of evaluative roles, including: Graders \citep{judge12,judge13}, Evaluators/Assessors \citep{judge14,judge15}, Critics \citep{judge16,judge17,judge18}, Verifiers \citep{judge20,judge21}, Examiners \citep{judge22}, Reward/Ranking Models \citep{judge23,judge24,judge25}, etc.

\subsection{Language Model Preference Optimization}

Reinforcement learning provides an effective solution for aligning LLMs with human values and controlling text generation \citep{po2,po3,po4,po5,po6,po7}. To this end, RLHF framework based on human feedback reward models has been established \citep{po1,po8,po9,po10,po11,po13}. However, despite its effectiveness, the complexity, instability, and hyperparameter sensitivity of RLHF remain insufficiently addressed \citep{po15,po16}. Recently proposed method named Direct Preference Optimization (DPO) \citep{dpo} simplifies the RLHF framework by eliminating the need for explicit reward model construction or reinforcement learning procedures, thereby avoiding dependence on reward models. Several variants have subsequently emerged, such as SimPO, KTO, and IPO \citep{po17,po18,po19}. Nevertheless, when applied to local preference alignment task, these methods still exhibit limitations, including coarse granularity and gradient dilution (see Appendix~\ref{sec:theory} for formal specification).

% Reinforcement learning是为将LLMs与人类价值观对齐提供了有效解决方案，也成为控制文本生成的有效手段 \citep{po2,po3,po4,po5,po6,po7}。为此，已形成了一套基于人类反馈奖励模型的Reinforcement Learning from Human Feedback（RLHF）框架 \citep{po1,po8,po9,po10,po11,po12,po13,po14}。不过尽管RLHF有效，但其复杂性、不稳定性和超参数敏感性尚未得到充分解决 \citep{po15,po16}。近来提出的DPO\citep{dpo}简化了RLHF框架，其无需显式构建奖励模型或执行强化学习流程，避免了对奖励模型的依赖。在其基础上也出现了多个变体，如SimPO、KTO和IPO \citep{po17,po18,po19}。然而在面对local偏好对齐任务时，这些技术仍然存在粒度过粗、梯度稀释等局限性（see Appendix~\ref{sec:theory} for formal specification）。

\section{Empirical Investigation}

% In this section, we validate the reliability of LLM-as-a-Judge and investigate the degree of liberal translation in subtitle translation.

% In this section，我们验证了LLM-as-a-Judge的reliability并调研了subtitle translation翻译的liberal translation程度。

\subsection{LLM Is Excellent Translation Evaluator}
\label{sec:judge}

We aim to verify whether LLMs can achieve high consistency with human preferences in quality assessment across different dimensions of subtitle translation, thereby enabling LLMs to serve as both reward models and evaluators for subtitle translation model alignment training. We begin with a formal definition of LLM-as-Evaluator:
\begin{equation}
\mathcal{E}\gets \pi _{\text{e}}(\mathcal{C}\oplus s\oplus \{t\}),
\end{equation}
where $\pi _{\text{e}}$ is the LLM used for evaluation, $\mathcal{C}$ is the introduction and instruction in the prompt that guides the LLM’s evaluation, $s$ is the line under evaluation, $\{t\}$ represents multiple translations of $s$, and $\mathcal{E}$ denotes the evaluation score given by $\pi _{\text{e}}$.

% 我们希望验证LLM在subtitle translation的不同维度的质量评估上能否满足与人类偏好的高度一致性，从而期待LLM能够作为subtitle translation模型对齐训练的reward model以及evaluator。首先，我们从一个formal definition of LLM-as-Evaluator开始：
% \begin{equation}
% \mathcal{E}\gets \pi _{\text{e}}(\mathcal{C}\oplus s\oplus \{t\}),
% \end{equation}
% where $\pi _{\text{e}}$是用于评估的LLM，$\mathcal{C}$是prompt中驱动LLM进行评估的introduction和instrustion，$s$是待评估的台词，$\{t\}是$s$的多个译文，$\mathcal{E}$是$\pi _{\text{e}}$给出的评估得分。

We investigated the correlation between human evaluators and LLM evaluators in terms of vividness scores for 10 different translations of 500 lines from our self-constructed \textbf{Mu}ltilingual \textbf{S}ubtitle \textbf{C}orpus (MuSC) dataset in Appendix~\ref{sec:dc}. Each evaluator assigned a score from 0 to 100 for the 10 translations of each line under evaluation, and then we computed the average Spearman rank correlation $\rho$ across evaluators. We present the Spearman rank correlation results in Figure~\ref{fig:preferencesrc}. The detailed experimental settings are provided in Appendix~\ref{sec:settingjudge}. The translation directions include \texttt{en}$\Rightarrow$\texttt{zh}, \texttt{en}$\Rightarrow$\texttt{de}, \texttt{zh}$\Rightarrow$\texttt{en}, and \texttt{zh}$\Rightarrow$\texttt{th}, where Chinese and English, German, and Thai correspond to high-, medium-, and low-resource languages, respectively.

% 我们调研了human evaluator和LLM evaluator对500句台词（来自MuSC dataset in Appendix~\ref{sec:dc}）的10个不同译文的生动性评分的相关性。我们令每个evaluator对每句待评估台词的10个译文打0-100的得分，然后计算多个evaluator得分序列的平均Spearman rank correlation $\rho$。我们在Figure~\ref{fig:preferencesrc}中展示了Spearman rank correlation的结果。具体的实验设置参看Appendix~\ref{sec:settingjudge}。涉及的翻译方向包括\texttt{en}$\Rightarrow$\texttt{zh}、\texttt{en}$\Rightarrow$\texttt{de}、\texttt{zh}$\Rightarrow$\texttt{en}和\texttt{zh}$\Rightarrow$\texttt{th}，其中中文和英文，德语，泰语分别为高中低资源语种。

\begin{figure*}[!h]
  \centering
  \subfigure[\texttt{en}$\Rightarrow$\texttt{zh}]{\includegraphics[width=0.235\textwidth]{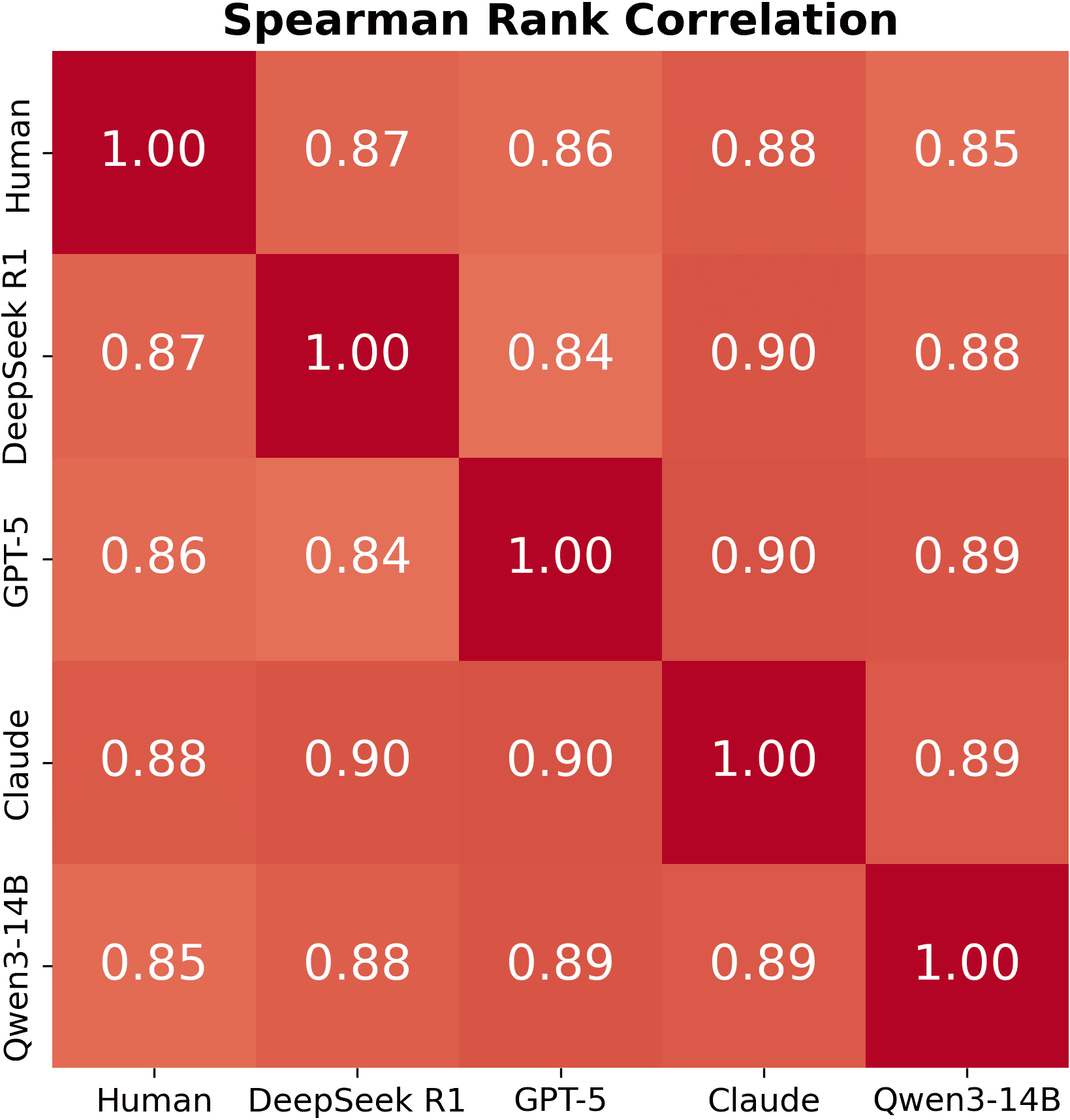}}
  \subfigure[\texttt{en}$\Rightarrow$\texttt{de}]{\includegraphics[width=0.235\textwidth]{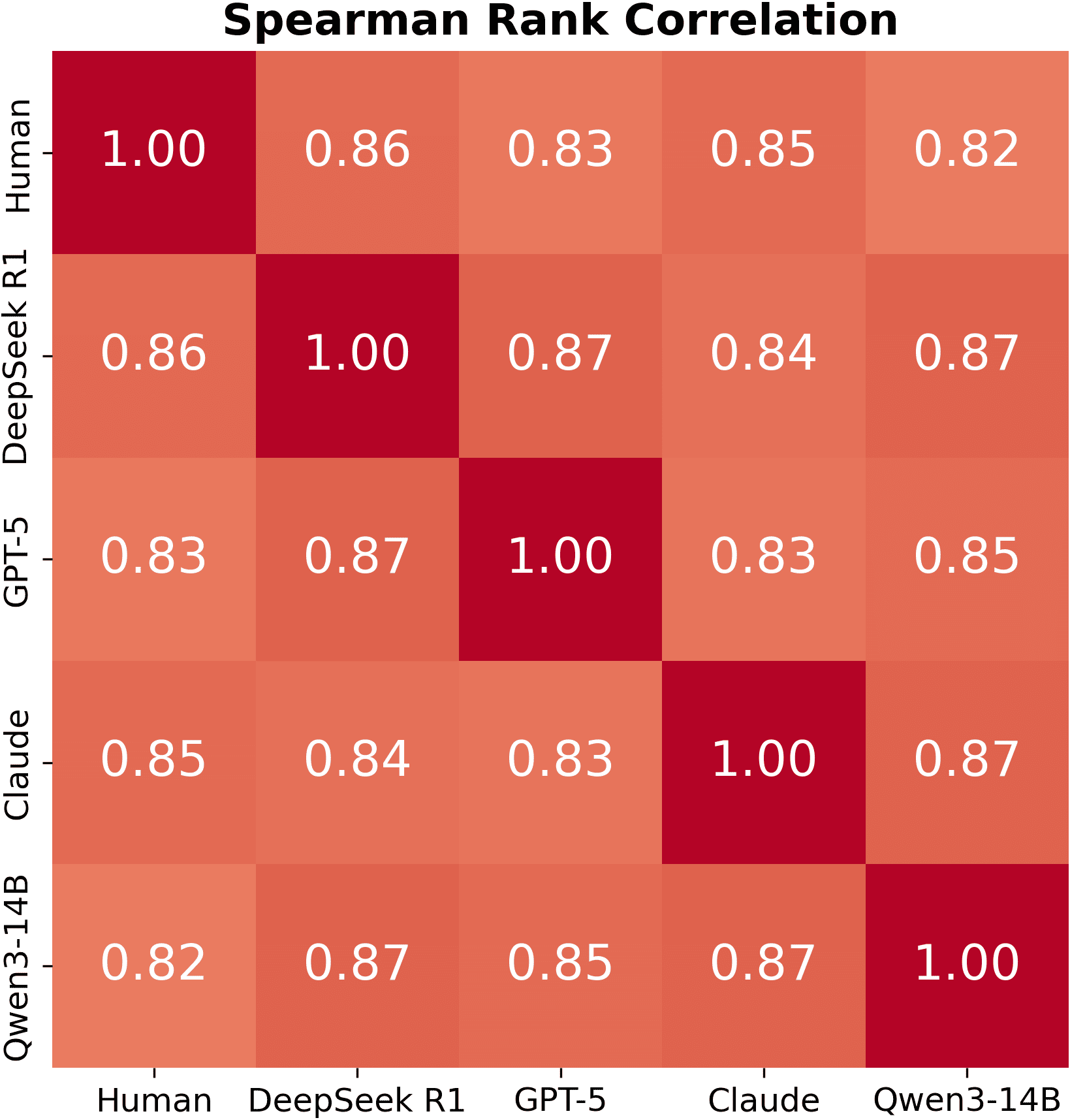}}
  \subfigure[\texttt{zh}$\Rightarrow$\texttt{en}]{\includegraphics[width=0.235\textwidth]{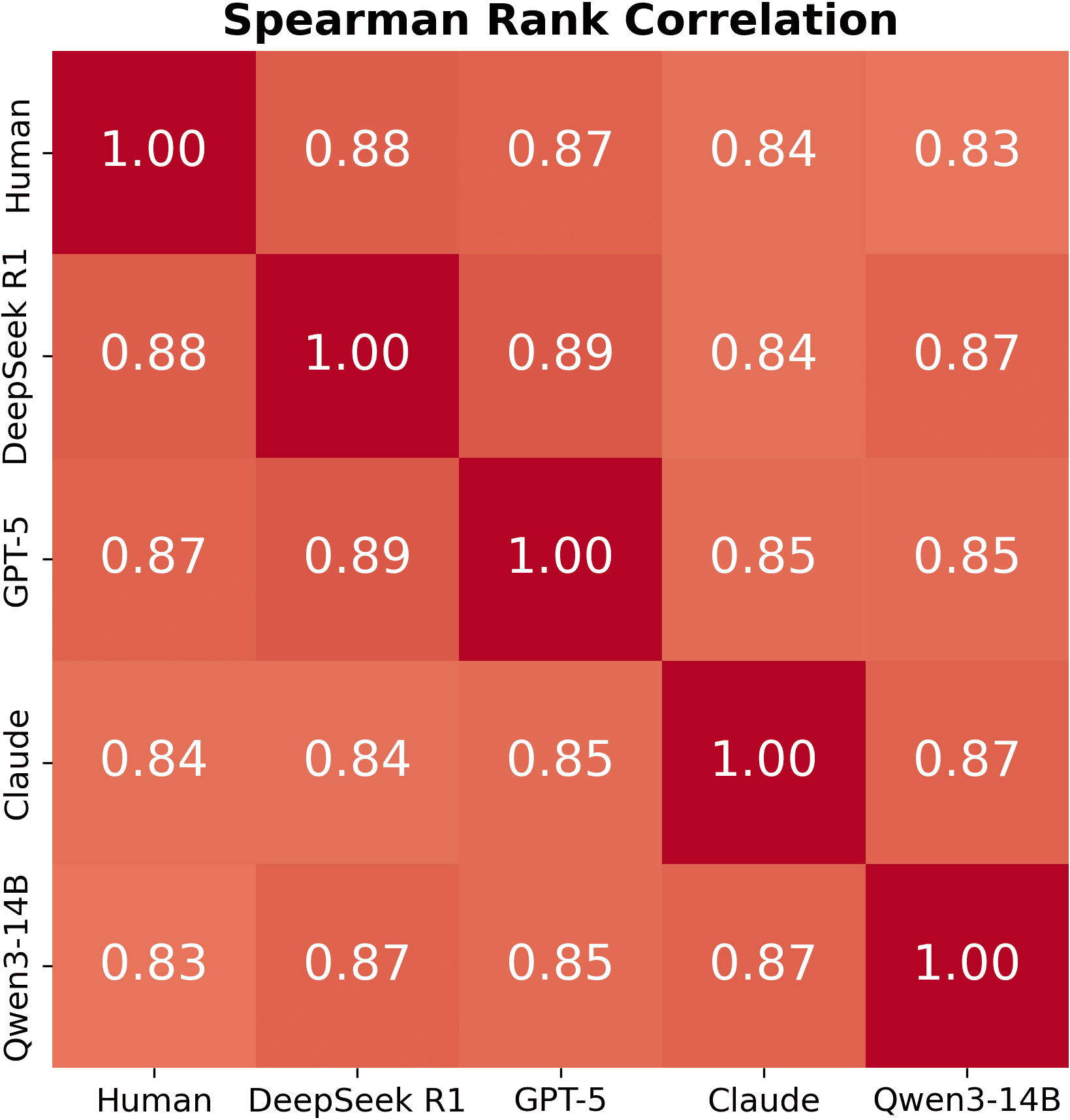}}
  \subfigure[\texttt{zh}$\Rightarrow$\texttt{th}]{\includegraphics[width=0.274\textwidth]{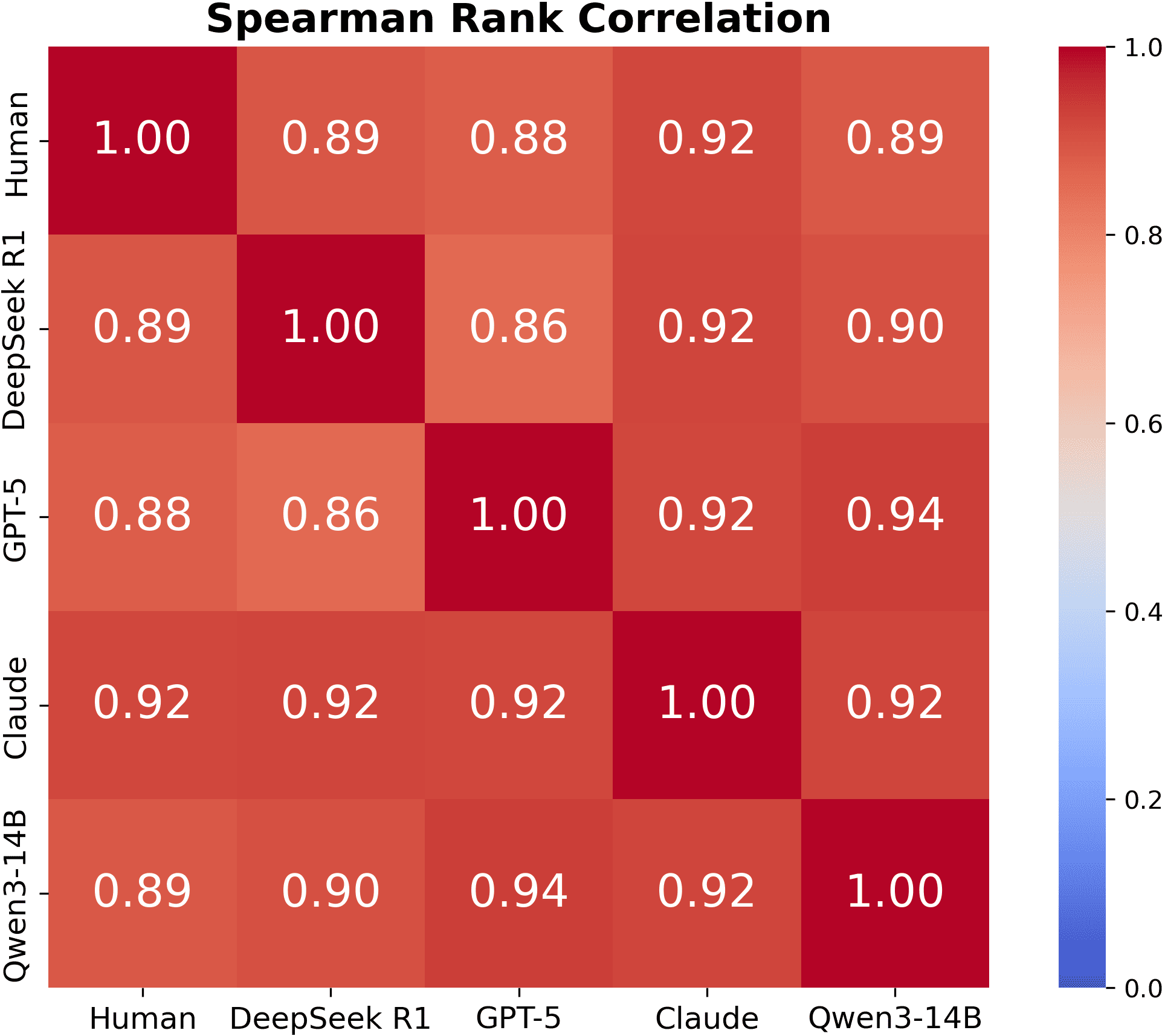}}
  \caption{Investigation of the consistency of multiple evaluators with Spearman rank correlation $\rho$.}
  \label{fig:preferencesrc}
\end{figure*}

\begin{figure*}[!h]
  \centering
  \subfigure[\texttt{en}$\Rightarrow$\texttt{zh}]{\includegraphics[width=0.49\textwidth]{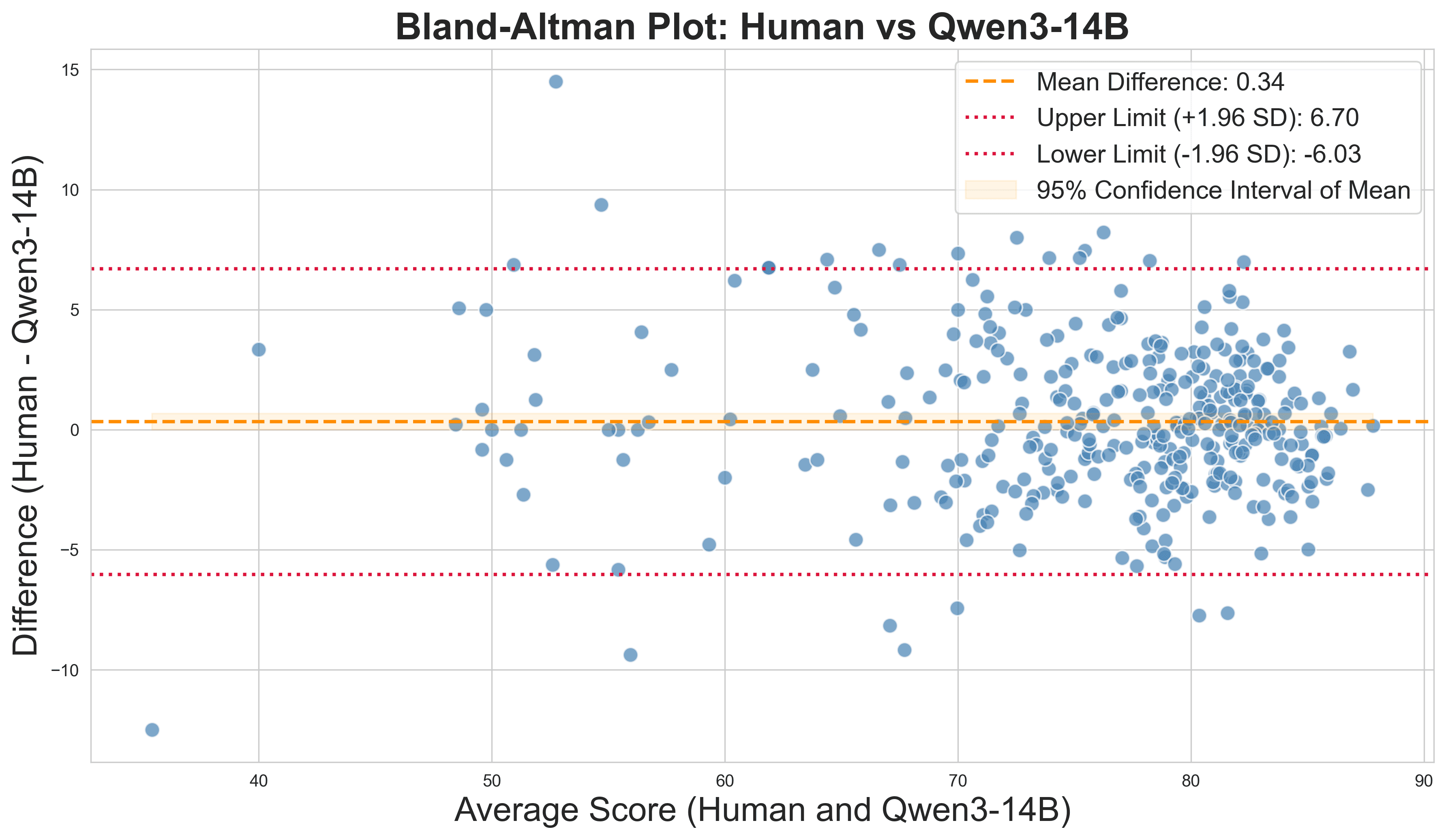}}
  \subfigure[\texttt{zh}$\Rightarrow$\texttt{th}]{\includegraphics[width=0.49\textwidth]{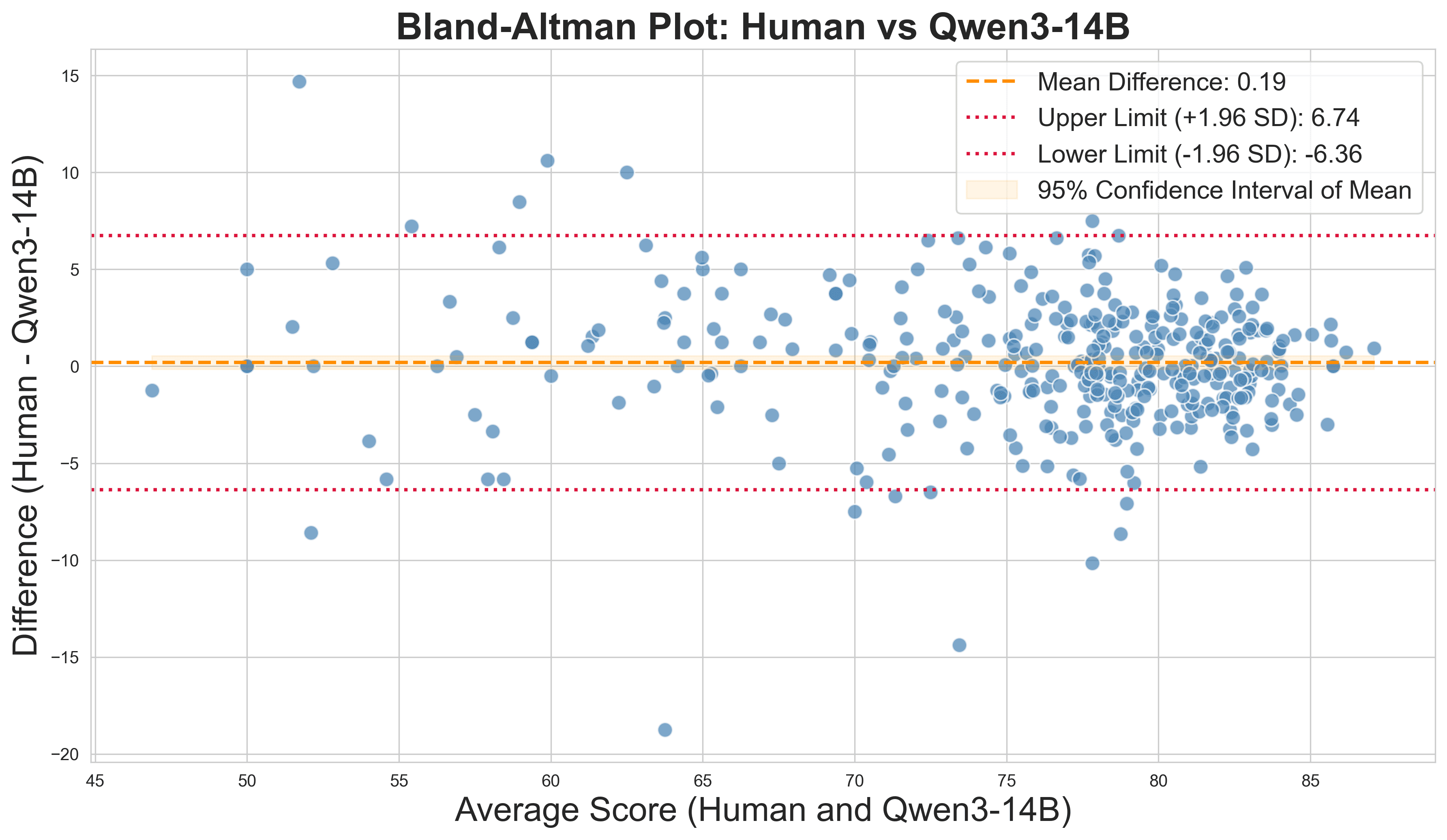}}
  \caption{Bland-Altman plot comparing Qwen3-14B with human evaluator.}
  \label{fig:bland}
\end{figure*}

The results show that the LLM evaluator exhibits high consistency with human evaluators across multiple directions, indicating its ability to handle deviations in cultural and linguistic conventions across languages. Notably, even the Qwen3-14B model, with 14B parameters, achieves high agreement with both human evaluators and other SOTA LLMs across all directions ($\rho\geq 0.82$). This indicates that a 14B model, which incurs relatively low inference cost, can serve as an efficient reward model, forming the foundation of ALPO. Furthermore, Figure~\ref{fig:bland} presents the Bland-Altman plots of Qwen3-14B and human evaluators for identical line translations in \texttt{en}$\Rightarrow$\texttt{zh} and \texttt{zh}$\Rightarrow$\texttt{th}. The results show that the mean difference (MD) between the two is very low, indicating minimal systematic bias, and its limits of agreement (LoA) are within acceptable error margins for scoring on a scale of 100. This further confirms the consistency between the 14B model and human evaluators.

% 结果表明LLM evaluator在多方向上表现出与人类evaluator的高度一致性，这表明LLM能够应对不同语种的cultural and linguistic conventions偏差。值得注意的是，即使是14B参数的Qwen3-14B模型也在所有方向上与人类evaluator及其他前沿LLM表现出了高度的一致性（$\rho\geq 0.82$）。这表明我们可以使用推理成本较低的14B模型来作为一个高效的reward model，这构成了ALPO方法的基础。进一步地，我们对每个evaluator对每句台词的同一译文的评分，在Figure~\ref{fig:bland}中展示了\texttt{en}$\Rightarrow$\texttt{zh}和\texttt{zh}$\Rightarrow$\texttt{th}上Qwen3-14B与人类evaluator的Bland-Altman plot。结果显示两者的mean difference (MD)很低，表明系统性偏差极小，且其limits of agreement (LoA)对于百分制打分来说位于可接受误差以内，这进一步验证了14B模型与human的一致性。

\subsection{Parallel Corpora Are Actually Not Parallel}
\label{sec:domain}

Literal translation focuses on translation accuracy, striving to preserve the original form, while liberal translation emphasizes conveying meaning and flexibly adjusting the expression \citep{translation1,translation2}. Based on our experience and observation, translations in different domains exhibit varying degrees of liberal translation: domains such as subtitle and literature translation often lean towards liberal translation, whereas legislation, medicine, and technical texts typically require highly accurate literal translation.

% literal translation注重翻译accuracy，力求保留原文形式，而liberal translation更侧重传达意义而灵活调整表达方式 \citep{translation1,translation2}。根据我们的经验和observation，不同领域的翻译展现出不同的liberal translation程度：subtitle translation和literature翻译等领域通常会更偏向liberal translation，而legislation、medicine、技术文本的翻译通常要求高准确性的literal translation。

We designed relevant experiments to quantitatively validate this hypothesis. We investigated the back-translation consistency of parallel corpora from different domains. Specifically, for a particular translation direction, we utilized GPT-4o to directly translate the target language translation back to the source language, and then calculated the BLEU and ChrF++ scores between the back-translated text and the original text. Lower scores indicate a higher degree of liberal translation. Detailed experimental settings are provided in Appendix~\ref{sec:settingback}.

% 我们设计了相关实验来量化地验证这一设想。我们调研了不同domain的翻译平行语料的back-translation consistency。具体的，对于一个翻译方向，我们借助GPT-4o将其目标语言译文直译回源语言，然后计算回译文本与原文本的BLEU和ChrF++指标，指标越低则表明liberal translation程度越高。具体的实验设置参看Appendix~\ref{sec:settingback}。

The investigation results in Table~\ref{tab:backtranslation} indicate that the back-translated texts in domains such as visual media, literature, and religion show lower similarity to the original texts compared to those in legislation, news, and medicine. This suggests a higher degree of liberal translation. For visual media subtitle translation, translators integrate video information into their translations, aiming to convey the atmosphere, emotions, and tone of the original text; this results in greater expressiveness and vividness, sometimes at the cost of sacrificing some degree of accuracy.

% Table~\ref{tab:backtranslation}中的调查结果表明visual media、literature、religion等领域相比legislation、news、medicine这些领域的回译文本与原文本展现出更低的相似性，即代表了更高的意译程度。对于visual media subtitle translation来说，译者结合视频信息进行翻译，更倾向于使字幕译文能够传达原文的atmosphere, emotions, and tone，从而展现出更高的expressiveness and vividness，有时甚至可以牺牲一定的准确性。

\begin{table*}[!h]
\centering
\caption{Domain liberal translation investigation. The highest and lowest are in \colorbox{mlb}{blue} and \colorbox{mlo}{orange}.}
\resizebox{\textwidth}{!}{
\begin{tabular}{llcc|cc|cc}
\Xhline{1.0pt}
\rowcolor{gray!20}
 ~ & ~ & \multicolumn{2}{c}{\texttt{en}$\Rightarrow$\texttt{de}} & \multicolumn{2}{c}{\texttt{en}$\Rightarrow$\texttt{fr}} & \multicolumn{2}{c}{\texttt{en}$\Rightarrow$\texttt{es}} \\
\cline{3-8}
\rowcolor{gray!20}
\multirow{-2}{*}{\textbf{Dataset}} & \multirow{-2}{*}{\textbf{Domain}} & \textbf{BLEU} & \textbf{ChrF++} & \textbf{BLEU} & \textbf{ChrF++} & \textbf{BLEU} & \textbf{ChrF++} \\
\hline
OpenSubtitles & Visual Media & \colorbox{mlo}{\textbf{15.00}} & \colorbox{mlo}{\textbf{37.08}} & \colorbox{mlo}{\textbf{17.84}} & \colorbox{mlo}{\textbf{37.80}} & 21.60 & 43.78 \\
\hdashline
Books & Literature & 17.22 & 43.20 & 21.51 & 47.14 & \colorbox{mlo}{\textbf{18.60}} & \colorbox{mlo}{\textbf{43.00}} \\
\hdashline
bible-uedin & Religion & 22.55 & 49.57 & 22.07 & 48.24 & 22.31 & 51.03 \\
\hdashline
DGT & \multirow{2}{*}{Legislation} & 19.85 & 50.05 & 22.44 & 48.10 & 21.88 & 52.24 \\
JRC-Acquis & ~ & \colorbox{mlb}{\textbf{27.55}} & \colorbox{mlb}{\textbf{61.95}} & \colorbox{mlb}{\textbf{28.83}} & \colorbox{mlb}{\textbf{69.65}} & 25.16 & 65.30 \\
\hdashline
News-Commentary & News & 25.00 & 55.23 & 24.87 & 54.37 & \colorbox{mlb}{\textbf{35.90}} & 65.84 \\
\hdashline
ECDC & \multirow{2}{*}{Medicine} & 23.18 & 60.77 & 23.86 & 62.97 & 27.20 & 65.39 \\
EMEA & ~ & 25.20 & 59.36 & 24.73 & 61.65 & 31.46 & \colorbox{mlb}{\textbf{69.15}} \\
\hline
\rowcolor{gray!20}
 ~ & ~ & \multicolumn{2}{c}{\texttt{de}$\Rightarrow$\texttt{en}} & \multicolumn{2}{c}{\texttt{fr}$\Rightarrow$\texttt{en}} & \multicolumn{2}{c}{\texttt{es}$\Rightarrow$\texttt{en}} \\
\cline{3-8}
\rowcolor{gray!20}
\multirow{-2}{*}{\textbf{Dataset}} & \multirow{-2}{*}{\textbf{Domain}} & \textbf{BLEU} & \textbf{ChrF++} & \textbf{BLEU} & \textbf{ChrF++} & \textbf{BLEU} & \textbf{ChrF++} \\
\hline
OpenSubtitles & Visual Media & 15.25 & \colorbox{mlo}{\textbf{39.41}} & \colorbox{mlo}{\textbf{16.60}} & \colorbox{mlo}{\textbf{40.65}} & 22.18 & 47.45 \\
\hdashline
Books & Literature & \colorbox{mlo}{\textbf{14.50}} & 40.50 & 20.12 & 48.29 & \colorbox{mlo}{\textbf{15.84}} & \colorbox{mlo}{\textbf{43.99}} \\
\hdashline
bible-uedin & Religion & 15.86 & 47.14 & 22.79 & 51.85 & 24.55 & 53.77 \\
\hdashline
DGT & \multirow{2}{*}{Legislation} & 19.70 & 47.77 & 23.31 & 48.91 & 22.87 & 52.49 \\
JRC-Acquis & ~ & \colorbox{mlb}{\textbf{27.54}} & \colorbox{mlb}{\textbf{63.43}} & \colorbox{mlb}{\textbf{31.33}} & 64.15 & 27.68 & 64.46 \\
\hdashline
News-Commentary & News & 22.64 & 54.05 & 29.34 & 56.20 & \colorbox{mlb}{\textbf{41.75}} & 66.35 \\
\hdashline
ECDC & \multirow{2}{*}{Medicine} & 21.09 & 57.41 & 29.13 & \colorbox{mlb}{\textbf{64.81}} & 29.62 & 64.20 \\
EMEA & ~ & 24.94 & 60.16 & 26.24 & 62.96 & 29.86 & \colorbox{mlb}{\textbf{67.09}} \\
\Xhline{1.0pt}
\end{tabular}
}
\label{tab:backtranslation}
\end{table*}

\subsection{Chat LLMs Favor Literal, Reason LLMs Excel in Liberal}
\label{sec:chatreason}

Generally, LLMs capable of effectively performing liberal translation tend to generate more expressive translations. We aim to verify the liberal translation capabilities of different LLMs. Our investigation spans multiple translation directions, where we sampled 2,000 lines from MuSC test set for each direction and prompted various LLMs to produce expressive and vivid translations (see the prompt in Appendix~\ref{sec:iosft}). We then computed the pairwise BLEU similarity between human and multiple LLM-generated translations. The LLMs included are chat models such as GPT-4o \citep{gpt4}, Qwen-Max \citep{qwen25}, and Claude Opus 4.1 \citep{claude}, as well as reasoning models like GPT-5 Thinking \citep{gpt5} and DeepSeek-R1 \citep{deepseekr1}. Additionally, we incorporated our supervised fine-tuning (SFT) Qwen2.5-14B model trained on MuSC.

% 通常，能够有效地进行liberal translation的LLM能够生成更expressive的译文，我们希望验证不同的LLMs liberal translation的能力。我们在多个翻译方向上进行了调研，我们从MuSC的测试集中为每个方向抽取了2000句台词，并且令不同的LLM尽可能地生成expressive and vivid的译文（采用的prompt参看Table~\ref{tab:prompt}），然后我们计算human和多个LLM译文的pairwise BLEU相似度。参与对比的LLM包括Chat模型GPT-4o \citep{gpt4}、Qwen-Max \citep{qwen25}、Claude Opus 4.1 \citep{claude}以及reason模型GPT-5 Thinking \citep{gpt5}和DeepSeek-R1 \citep{deepseekr1}，另外也包括我们在MuSC上supervised fine-tuning (SFT) 的Qwen2.5-14B模型。

\begin{figure*}[!h]
  \centering
  \subfigure[\texttt{en}$\Rightarrow$\texttt{zh}]{\includegraphics[width=0.245\textwidth]{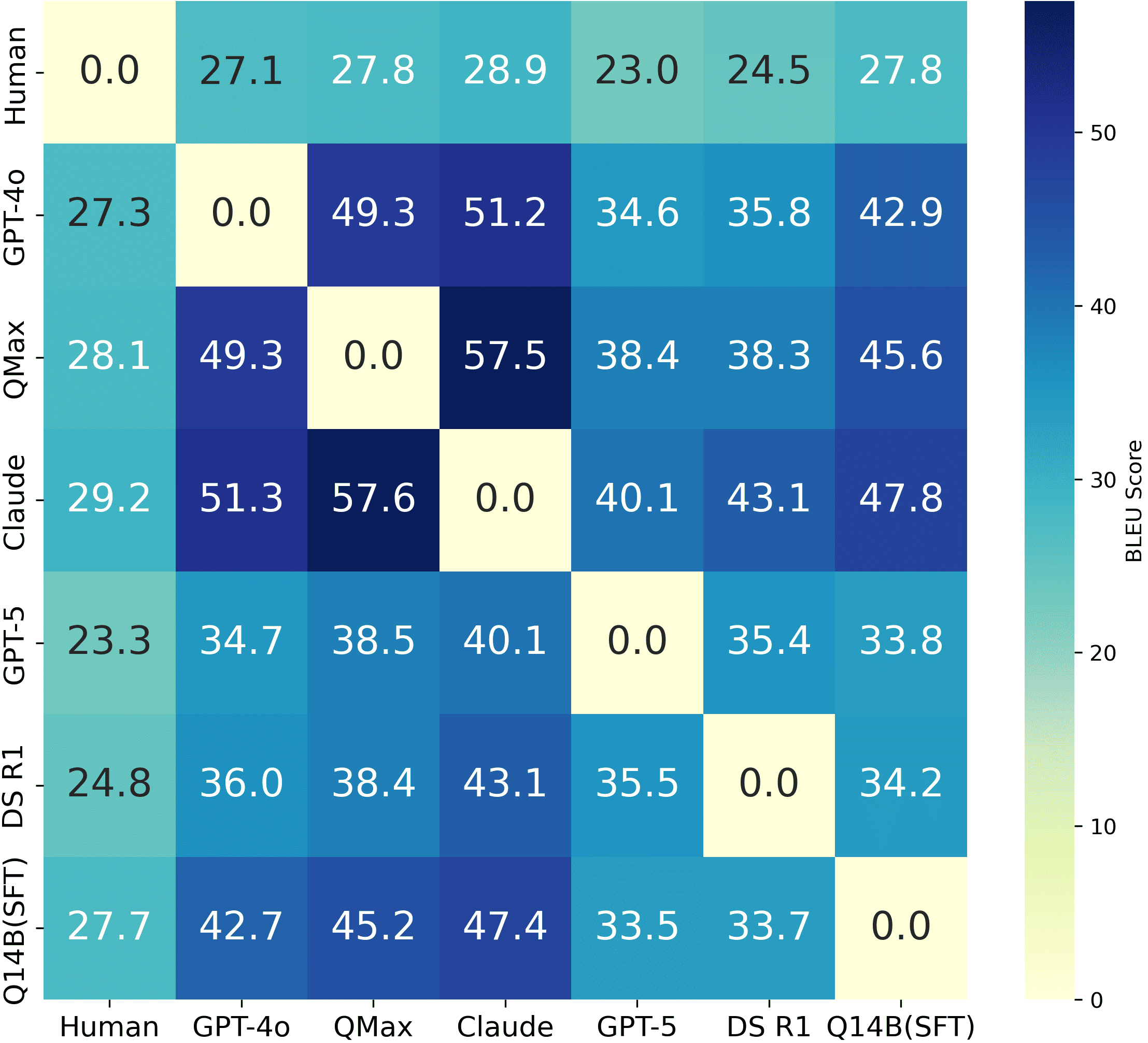}}
  \subfigure[\texttt{en}$\Rightarrow$\texttt{de}]{\includegraphics[width=0.245\textwidth]{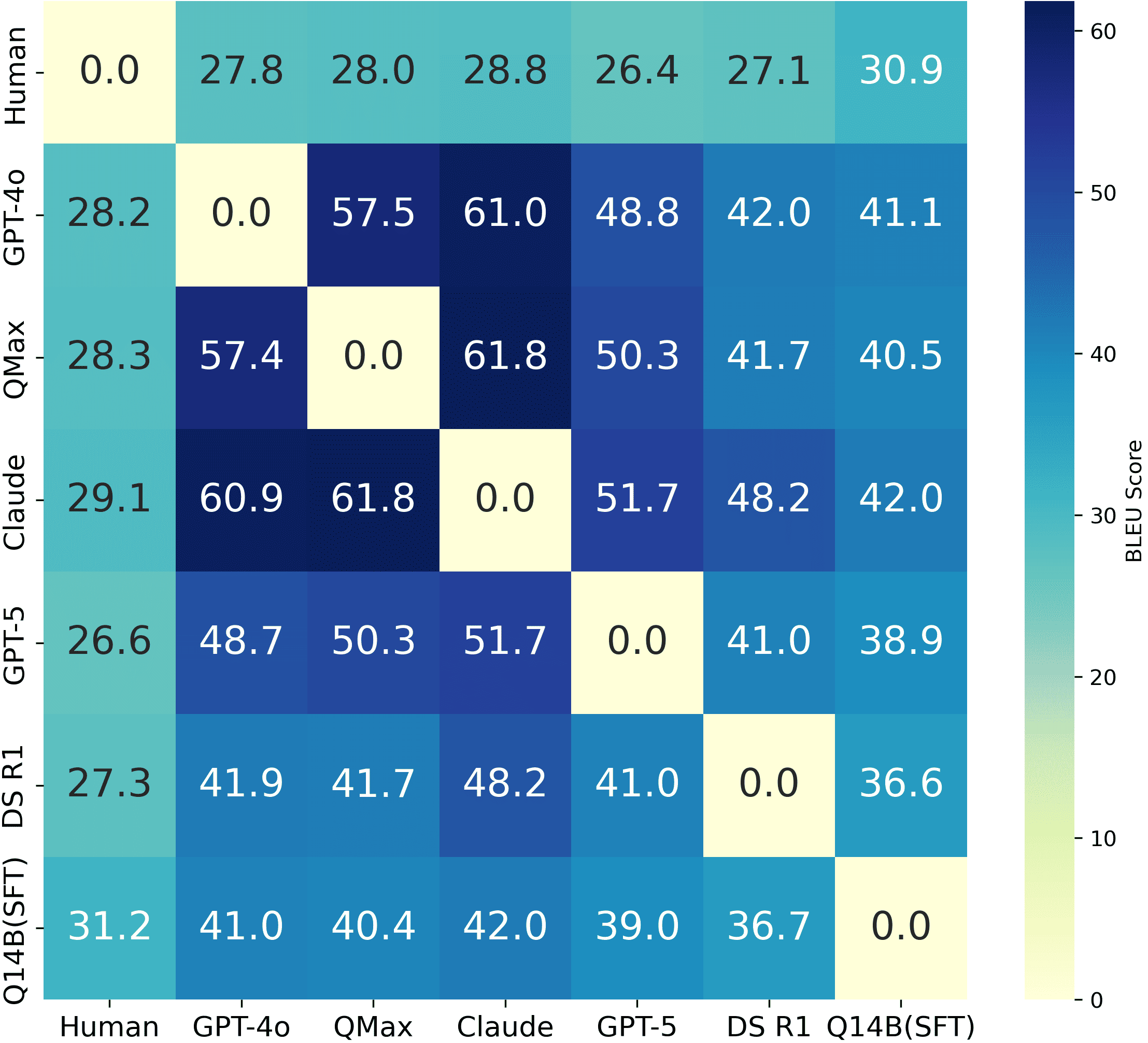}}
  \subfigure[\texttt{zh}$\Rightarrow$\texttt{en}]{\includegraphics[width=0.245\textwidth]{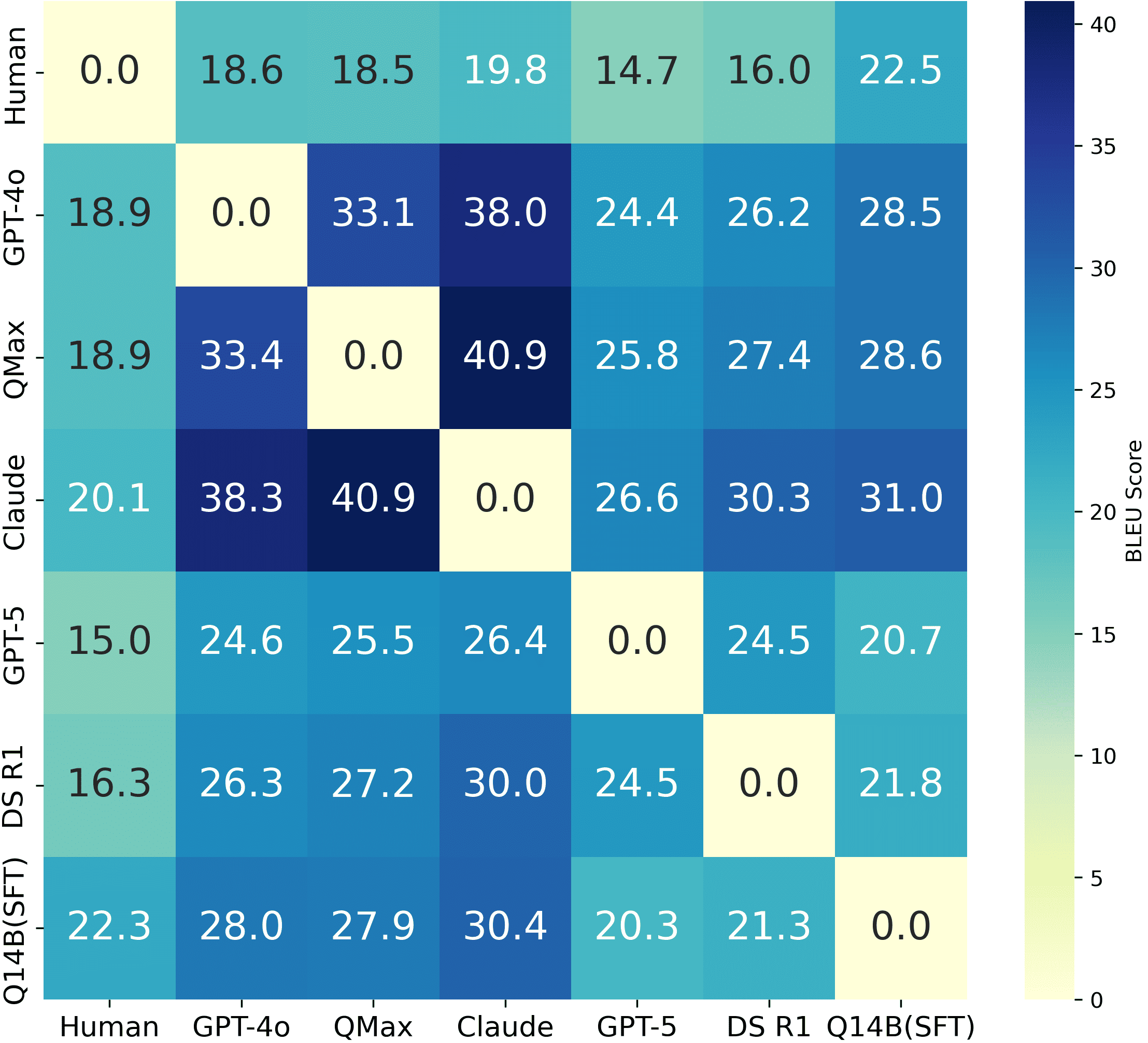}}
  \subfigure[\texttt{zh}$\Rightarrow$\texttt{th}]{\includegraphics[width=0.245\textwidth]{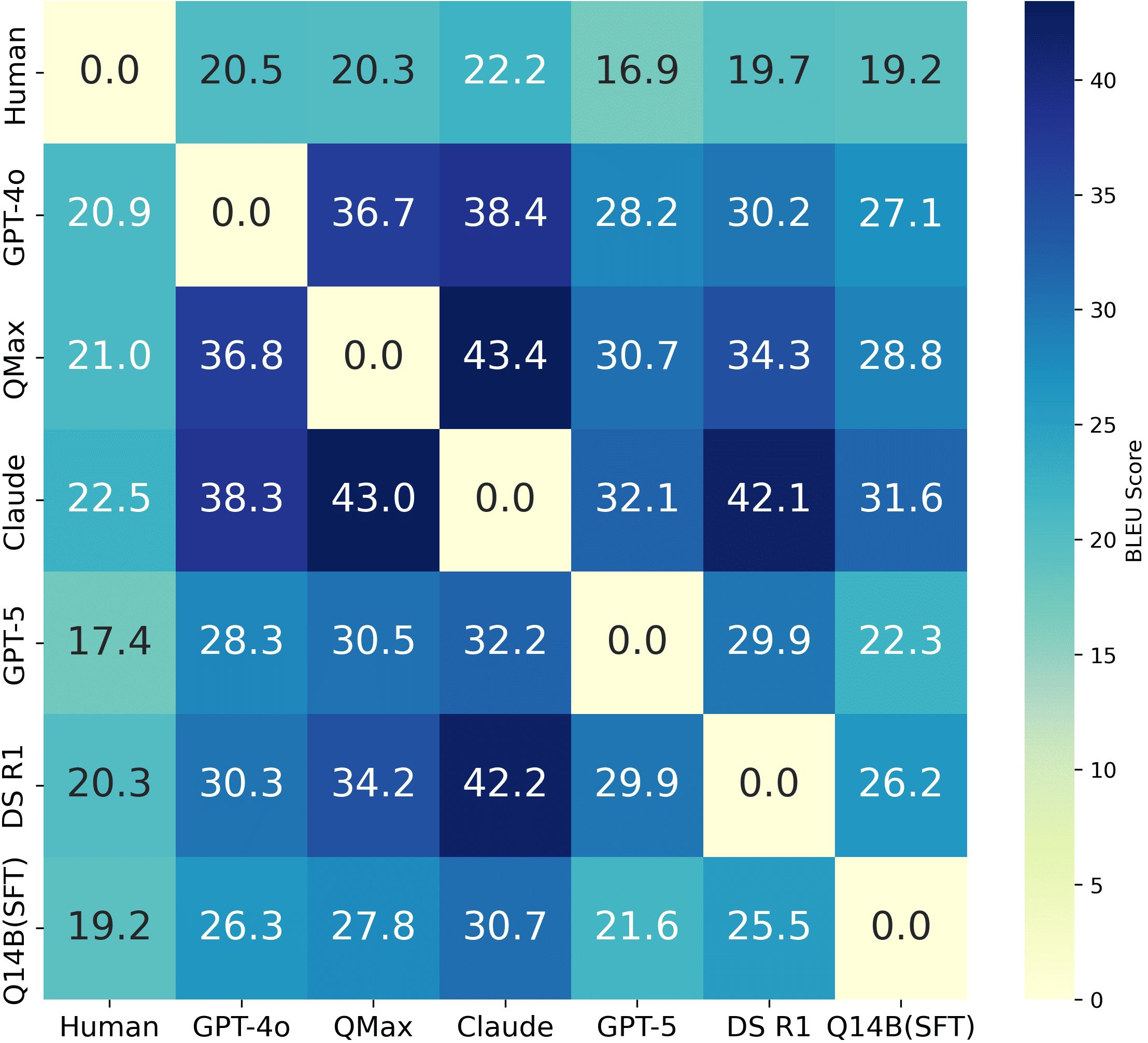}}
  \caption{Pairwise BLEU scores between translations.}
  \label{fig:chatvsreasonbleu}
\end{figure*}

The results in Figure~\ref{fig:chatvsreasonbleu} indicate that translations generated by chat models exhibit higher similarity among themselves, suggesting a tendency toward more literal translation. In contrast, translations produced by reasoning models show lower similarity with other translations, demonstrating that these models, through thinking, better adhere to the instruction of liberal translation. This validates that LLM's inference-time scaling can effectively enhance translation performance. The human translation exhibits lower similarity with other translations, indicating its proficiency in liberal translation. We present demos of different translations in Appendix~\ref{sec:demo} to visually demonstrate these findings.

% Figure~\ref{fig:chatvsreasonbleu}中的实验结果结果显示chat模型的译文之间展现出更高的相似度，这表明chat模型更偏literal translation。而具有思考能力的reason模型译文与其他译文的相似度较低，这表明reason模型经过思考后更能遵从liberal translation的指令，这验证了LLM的inference-time scaling能够很好地提升翻译能力。Human译文与其他译文的相似度也较低，这表明human翻译也能很好地进行liberal translation。我们在Appendix~\ref{sec:demo}中展示了不同译文的demo，可以直观地体现这些结论。

% \begin{figure*}[!h]
%   \centering
%   \subfigure[\texttt{en}$\Rightarrow$\texttt{zh}]{\includegraphics[width=0.245\textwidth]{img/en2zh_chrf_matrix.png}}
%   \subfigure[\texttt{en}$\Rightarrow$\texttt{de}]{\includegraphics[width=0.245\textwidth]{img/en2de_chrf_matrix.png}}
%   \subfigure[\texttt{zh}$\Rightarrow$\texttt{en}]{\includegraphics[width=0.245\textwidth]{img/zh2en_chrf_matrix.png}}
%   \subfigure[\texttt{zh}$\Rightarrow$\texttt{th}]{\includegraphics[width=0.245\textwidth]{img/zh2th_chrf_matrix.png}}
%   \caption{Pairwise ChrF++ scores between translations.}
%   \label{fig:chatvsreasonchrf}
% \end{figure*}

\section{Method}

\subsection{Notations}

A bilingual dataset $\mathbb{D}$ for subtitle translation comprises collections of source language lines $\mathcal{L}_{\text{src}}$ and target language lines $\mathcal{L}_{\text{tgt}}$ from multiple programs of genres such as films or TV series, represented by $\mathbb{D}\equiv \{(\ell_{\text{src}}, \ell_{\text{tgt}})\in \mathcal{L}_{\text{src}}\times \mathcal{L}_{\text{tgt}}\}$. The lines in multilingual subtitles of the same program are often not in one-to-one correspondence. To address this, we have developed a bilingual parallel corpus construction algorithm with $O(N)$ complexity; details can be found in Appendix~\ref{sec:saa}. We will utilize the dataset $\mathbb{D}$ to perform SFT and ALPO training on off-the-shelf LLMs. The input prompt $x$ for the LLM includes the context $\mathcal{C}$ (comprising introduction and instructions) and a set of $n$ source lines $\{s_i\mid i \in [n]\}$ that need to be translated, i.e., $x\equiv \mathcal{C}\oplus \{s_i\}$. The response $y$ from the LLM includes translations for each line $\{t_i\mid i \in [n]\}$, i.e., $y\equiv \{t_i\}$. See Appendix~\ref{sec:iosft} for detailed $x$ and $y$ format.

% 一个用于subtitle翻译的双语数据集$\mathbb{D}$包含film或TV series等genre的多个programs的源语言台词集合$\mathcal{L}_{\text{src}}$与目标语言台词集合$\mathcal{L}_{\text{tgt}}$，即$\mathbb{D}\equiv \{(\ell_{\text{src}}, \ell_{\text{tgt}})\in \mathcal{L}_{\text{src}}\times \mathcal{L}_{\text{tgt}}\}$。同一个节目的多语言字幕的台词之间通常不是一一对应的，为此我们开发了一种$O(N)$复杂度的双语平行语料构建算法，具体请参看Appendix~\ref{sec:saa}。我们将利用数据集$\mathbb{D}$来完成对off-the-shelf LLM的SFT和ALPO训练。LLM的输入prompt$x$包括上下文$\mathcal{C}$（包括introduction和instructions等文本）与需要翻译的一组$n$句source台词$\{s_i\mid i \in [n]\}$，即$x\equiv \mathcal{C}\oplus \{s_i\}$。LLM的response$y$包括对应于每句台词译文$\{t_i\mid i \in [n]\}$，即$y\equiv \{t_i\}$。详细的prompt与response的示例参看Appendix~\ref{sec:iosft}。

\subsection{Overall Framework}

\begin{figure*}[t]
  \centering
  \includegraphics[width=\textwidth]{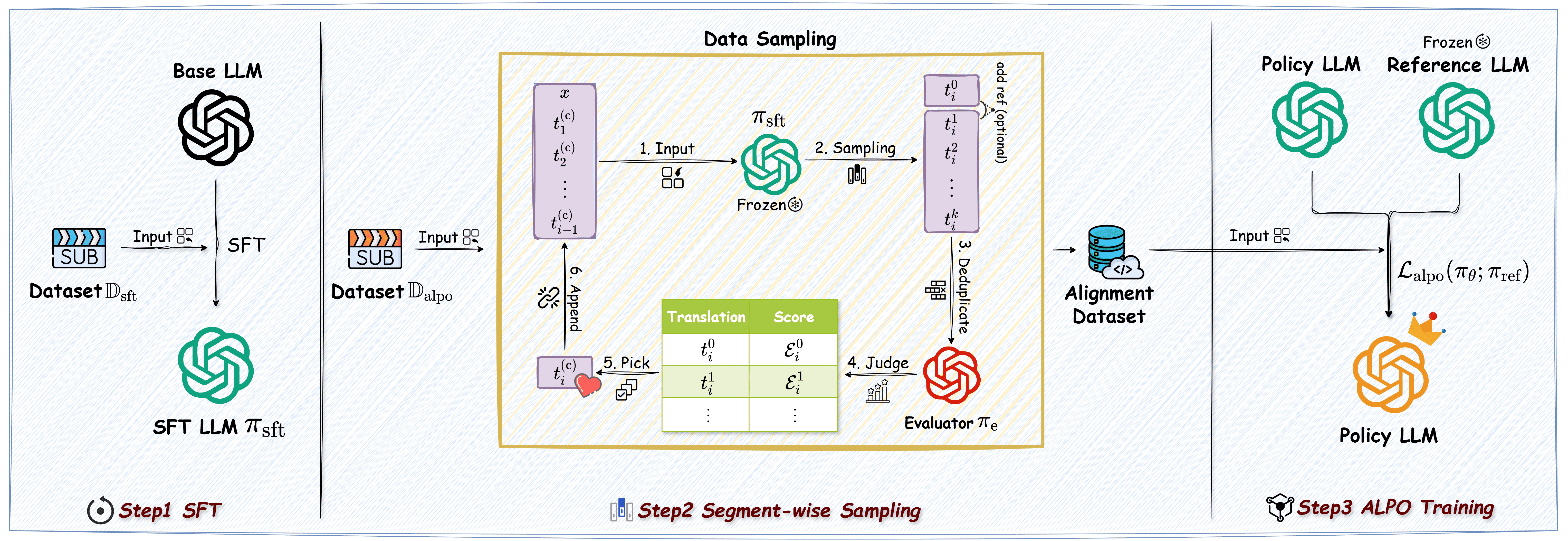}
  \caption{The overall framework of ALPO.}
  \label{fig:framework}
\end{figure*}

Obviously, constructing an expressive translation LLM is a preference optimization task. However, we cannot directly apply general methods (e.g., PPO \citep{po1} or DPO \citep{dpo}), primarily because the translation of each line of subtitle depends on its context. The input $x$ of the SFT model $\pi_{\text{sft}}$ must include multiple lines of subtitle $\{s_i\}$, necessitating finer-grained alignment for each line in the response $y$. Consequently, outcome-supervised methods like PPO and DPO, which optimize the complete output of LLM, are inadequate for this task. We present experimental and theoretical validation of this in Appendix~\ref{sec:furtherexp} and Appendix~\ref{sec:theory}. To address this fine-grained preference optimization task, we propose the process-supervised ALPO method. ALPO introduces a novel paradigm that leverages a segment-wise sampling strategy and fine-grained alignment loss to train a high-quality subtitle translation LLM. The overall framework of ALPO is shown in Figure~\ref{fig:framework}.

% 显然，构造一个expressive的translation LLM是一个preference optimization task，但是我们并不能直接应用一些通用的算法（例如PPO \citep{po1} or DPO \citep{dpo}），这主要是因为对每一句台词的翻译需要依赖于其上下文，SFT模型$\pi _{\text{sft}}$的输入$x$中需要包含多句台词$\{s_i\}$，因而需要对response $y$中的每句台词进行更细粒度的对齐，从而导致PPO和DPO这些针对LLM完整输出进行优化的Outcome-supervised的方法不能胜任。我们在Appendix~\ref{sec:furtherexp}和Appendix~\ref{sec:theory}展示了实验和理论验证了这一点。为了能够解决这个细粒度的preference optimization任务，我们提出了process-supervised的ALPO方法。ALPO提供了一种新的paradigm，能够利用segment-wise sampling strategy和fine-grained alignment loss来训练一个高质量的subtitle translation LLM。我们在Figure~\ref{fig:framework}中展示了ALPO算法的整体框架。

\subsection{Sampling Strategy}

We refer to a task requiring multi-segment local preference alignment of LLM response as a \textbf{local preference optimization} task. ALPO provides an effective paradigm for handling such tasks. We initially divide the parallel corpus $\mathbb{D}$ into a demonstration dataset $\mathbb{D}_{\text{sft}}$ and an alignment dataset $\mathbb{D}_{\text{alpo}}$ (approximately 8:2), using $\mathbb{D}_{\text{sft}}$ to train a SFT model $\pi _{\text{sft}}$ (see Appendix~\ref{sec:iosft} for input-output format).

% 我们将一个需要对LLM的response进行multi-segment local preference alignment的任务称为\textbf{local preference optimization}任务。ALPO提供了处理这样的任务的有效paradigm。我们首先将平行语料数据集$\mathbb{D}$划分为demonstration dataset $\mathbb{D}_{\text{sft}}$和alignment dataset $\mathbb{D}_{\text{alpo}}$（约8:2），并使用$\mathbb{D}_{\text{sft}}$训练一个SFT翻译模型$\pi _{\text{sft}}$（其输入输出格式见Appendix~\ref{sec:iosft}）。

In sampling phase, for a $x\in \mathbb{D}_{\text{alpo}}$, which contains $n$ lines $s_{1}, \dots, s_{n}$. For each $s_{i}$, using $p_{i}=x, t_{1}^{\text{(c)}}, \dots, t_{i-1}^{\text{(c)}}$ as prefix, $k$ translations are sampled ($k=15$ in experiments), i.e., $\pi _{\text{sft}}(t_{i}^{j} \mid p_{i})$, $j=1, \dots, k$, resulting in a candidate set $\{t_{i}^{j} \mid j=1, \dots, k\}$. Then, the candidate set is deduplicated; if human reference $t_{i}^{0}$ is available, it can be added to the candidate set, forming $\mathcal{T}_{i} = \{t_{i}^{j} \mid j=0, 1, \dots \}$. Subsequently, a Qwen3-14B model is deployed as an evaluator $\pi _{\text{e}}$ (or can be regarded as a reward model) to evaluate vividness of $\mathcal{T}_{i}$, obtaining the score sequence $\mathcal{E}_{i}$ (see Table~\ref{tab:daexample} for demonstration). Based on $\mathcal{E}_{i}$, we select a superior translation $t_{i}^{\text{(c)}}$ for $s_{i}$ (randomly chosen from the top 3 scores) to serve as the prefix for the next sampling cycle. Ultimately, for a $x \in \mathbb{D}_{\text{alpo}}$, we obtain the sampling result $\mathcal{S}(x) \equiv \{(s_{i}, \mathcal{T}_{i}, \mathcal{E}_{i}) \mid i=1, \dots, n\}$. Algorithm~\ref{alg:sampling} details the entire process.

% 在ALPO采样阶段，对于一个$x\in \mathbb{D}_{\text{alpo}}$，其中包含$n$句台词$s_{1},s_{2},\dots ,s_{n}$。对于每句台词$s_{i}$，以$p_{i}=x,t_{1}^{\text{(c)}},\dots ,t_{i-1}^{\text{(c)}}$为prefix采样$k$个候选译文（实验中我们采用$k=15$），即$\pi _{\text{sft}}(t_{i}^{j}\mid p_{i}),j=1,\dots ,k$，得到候选集$\{t_{i}^{j}|j=1,2,\dots ,k\}$。接着我们对候选集进行去重，如果human reference译文$t_{i}^{0}$是available的，则可以将其加入候选集中，得到$\mathcal{T}_{i}=\{t_{i}^{j}\mid j=0,1,\dots \}$。然后我们使用一个Qwen2.5-14B-Instruct模型作为evaluator $\pi _{\text{e}}$（或者可视为reward model）对$\mathcal{T}_{i}$进行vividness评估，得到对应的得分序列$\mathcal{E}_{i}$ (see Table~\ref{tab:daexample} for demonstration)，然后我们根据$\mathcal{E}_{i}$选择$s_{i}$的一个superior译文$t_{i}^{\text{(c)}}$（从top 3得分中随机挑选一个）作为下一个sampling cycle的prefix。最终，对于样本$x\in \mathbb{D}_{\text{alpo}}$，我们得到对应的采样结果$\mathcal{S}(x)\equiv \{(s_{i},\mathcal{T}_{i},\mathcal{E}_{i})\mid i=1,2,\dots ,n\}$。Algorithm~\ref{alg:sampling}详细展示了整个流程。

\begin{table*}[!h]
\centering
\caption{The \texttt{zh}$\Rightarrow$\texttt{en} evaluation demonstration of the evaluator $\pi _{\text{e}}$.}
\resizebox{\textwidth}{!}{
\begin{tabular}{lc}
 \Xhline{1.0pt}
 \rowcolor{gray!20}
 \textbf{Line} & \textbf{Evaluation of $\pi _{\text{e}}$} \\
 \hline
 \begin{CJK}{UTF8}{gkai}历史虽然会重演，但是人类是无法回到过去的。\end{CJK} & - \\
 \hdashline
 \texttt{History repeats, but we can’t go back to what was.} & 70 \\
 \texttt{History might replay, but mankind cannot go back in time.} & 85 \\
 \texttt{The wheel of history may turn full circle, but the door to the past stays forever locked.} & 88 \\
 \texttt{Although history may repeat itself, humans cannot return to the past.} & 82 \\
 \texttt{History often echoes, yet there’s no way for us to turn back the clock.} & 92 \\
 \Xhline{1.0pt}
\end{tabular}
}
\label{tab:daexample}
\end{table*}

\begin{algorithm}[!h]
  \caption{ALPO Sampling Strategy.}
  \label{alg:sampling}
  \begin{algorithmic}[1]
    \Require SFT model $\pi _{\text{sft}}$, evaluation LLM $\pi _{\text{e}}$, alignment dataset $\mathbb{D}_{\text{alpo}}$, sample number $k$.
    \Ensure sampled segment-level candidate set $\mathcal{S}(x)$.
    \For {any $x\in \mathbb{D}_{\text{alpo}}$} \quad $//$ Iterate through the alignment dataset $\mathbb{D}_{\text{alpo}}$.
    \For {$i=1$ to $n$} \quad $//$ Iterate through the subtitle lines in $x$.
    \For {$j=1$ to $k$} \quad $//$ Sample multiple translation candidates.
    \State Sample $\pi _{\text{sft}}(t_{i}^{j}\mid x,t_{1}^{\text{(c)}},\dots ,t_{i-1}^{\text{(c)}})$.
    \EndFor
    \State Deduplicate the candidate set $\{t_{i}^{j}\mid j=1,\dots ,k\}$.
    \State (Optional) Add human reference $t_{i}^{0}$ to the candidate set, get $\mathcal{T}_{i}=\{t_{i}^{j}\mid j=0,1,\dots \}$.
    \State Evaluate $\mathcal{T}_{i}$ by $\pi _{\text{e}}$, get score sequence $\mathcal{E}_{i}$.
    \State Select a superior translation $t_{i}^{\text{(c)}}$ (random from top 3) as the prefix of next sampling cycle.
    \EndFor
    \EndFor \\
    \Return $\mathcal{S}(x)\equiv \{(s_{i},\mathcal{T}_{i},\mathcal{E}_{i})\mid i=1,2,\dots ,n\}$.
  \end{algorithmic}
\end{algorithm}

\subsection{Adaptive Alignment Loss}

The reward signal acts on segments of model output (process-supervised) rather than the complete response (outcome-supervised), which constitutes the main challenge for vividness enhancement. ALPO enables process-supervised alignment across multiple segments in translation LLM’s response $y$. Specifically, for all lines $\{s_i\mid i=1,\dots ,n\}$ in an $x\in \mathbb{D}_{\text{alpo}}$, we assign each $s_i$ an adaptive weight $w(s_{i})$, defined as the product of a gating function $\mathbf{1}(s_{i})$ and an importance score $\delta (s_{i})$:
\begin{equation}
w(s_{i})=\mathbf{1}(s_{i})\cdot \delta (s_{i}),
\end{equation}
where $\mathbf{1}(s_{i})$ acts as a gate to determine whether $s_i$ participates in optimization. When the sampled translations lack diversity or clear quality distinction, i.e., $|\mathcal{T}_{i}|\leq 3$ or $\max(\mathcal{E}_{i})-\min(\mathcal{E}_{i})\leq 5$, we set $\mathbf{1}(s_{i})=0$; otherwise, $\mathbf{1}(s_{i})=1$. The importance score $\delta (s_{i})$ depends on the diversity of sampled translations. Since lines with richer translations provide more potential for vividness enhancement, we assign higher $\delta (s_{i})$ to lines with a larger number of distinct sampled translations after deduplication:
\begin{equation}
\delta (s_{i})=\frac{|\mathcal{T}_{i}|}{\sum_{j=1}^{n}|\mathcal{T}_{j}|}.
\end{equation}

% 奖励信号作用在model output的segment上（process-supervised）而非完整的response上（outcome-supervised），这是vividness enhancement面临的主要挑战。ALPO能够对translation LLM的response $y$中的多个segment进行process-supervised的逐个对齐。具体的，对于一个$x\in \mathbb{D}_{\text{alpo}}$中所有台词$\{s_i\mid i=1,\dots ,n\}$，我们会为每个$s_i$设置一个自适应的权重$w(s_{i})$，其由一个门控函数$\mathbf{1}(s_{i})$和重要性分数$\delta (s_{i})$的乘积组成：
% \begin{equation}
% w(s_{i})=\mathbf{1}(s_{i})\cdot \delta (s_{i}),
% \end{equation}
% where $\mathbf{1}(s_{i})作为一个门控来决定每个$s_i$是否参与优化。当采样译文缺乏多样性或者优劣不分明，即$|\mathcal{T}_{i}|\leq 3$或者$\max(\mathcal{E}_{i})-\min(\mathcal{E}_{i})\leq 5$时，我们设置$\mathbf{1}(s_{i})=0$，其余情况设置$\mathbf{1}(s_{i})=1$。而重要性分数$\delta (s_{i})$由采样译文的多样程度决定，我们认为译文丰富的台词比译文单一的台词有更大的提升生动性的空间，因此我们为去重后剩余采样译文数量高的台词分配更高的重要性分数$\delta (s_{i})$：
% \begin{equation}
% \delta (s_{i})=\frac{|\mathcal{T}_{i}|}{\sum_{j=1}^{n}|\mathcal{T}_{j}|}.
% \end{equation}

We then apply a Bradley–Terry (DPO-style) preference optimization loss to achieve fine-grained preference alignment of the translation model:
\begin{equation}
\mathcal{L}_{\text{alpo}}(\pi _{\theta };\pi _{\text{ref}})=-\mathbb{E}_{(x,\mathcal{S}(x))\sim \mathbb{D}_{\text{alpo}}}\Bigg[\sum_{i=1}^{n}w(s_{i})\cdot \text{log}\, \sigma \bigg(\beta _{i}\, \text{log}\frac{\pi _{\theta }(t_{i}^{\text{(c)}}\mid p_{i})}{\pi _{\text{ref}}(t_{i}^{\text{(c)}}\mid p_{i})}-\beta _{i}\, \text{log}\frac{\pi _{\theta }(t_{i}^{\text{(r)}}\mid p_{i})}{\pi _{\text{ref}}(t_{i}^{\text{(r)}}\mid p_{i})}\bigg)\Bigg].
\end{equation}
where $\pi _{\theta }$ is the policy model, and $\pi _{\text{ref}}$ is the reference model. The chosen translation $t_{i}^{\text{(c)}}$ is randomly sampled from the top 3 scored translations in $\mathcal{E}_{i}$, while the rejected translation $t_{i}^{\text{(r)}}$ is selected as the third-lowest scored candidate (excluding the lowest one to avoid overly trivial contrastive pairs). $\beta _{i}$ is a hyperparameter controlling sensitivity to reward differences. Since the reward gap between $t_{i}^{\text{(c)}}$ and $t_{i}^{\text{(r)}}$ also reflects the diversity of translations, we set $\beta _{i}$ dynamically as follows:
\begin{equation}
\beta _{i}=\frac{r(s_{i},t_{i}^{\text{(c)}})-r(s_{i},t_{i}^{\text{(r)}})}{\max\{r(s_{j},t_{j}^{\text{(c)}})-r(s_{j},t_{j}^{\text{(r)}})\mid j \in [n]\}},
\end{equation}
where $r(s_{i},t_{i}^{\text{(c)}})$ denotes the score of $t_{i}^{\text{(c)}}$ in $\mathcal{E}_{i}$. Moreover, instead of directly setting the prefix in $\mathcal{L}_{\text{alpo}}$ as $p_{i}=x,t_{1}^{\text{(c)}},\dots ,t_{i-1}^{\text{(c)}}$, we adopt a scheduled prefix mixing strategy to mitigate exposure bias when non-chosen translations are generated during inference. Specifically, with probability $\lambda $, $t_{i}^{\text{(c)}}$ is appended to the prefix, and with probability $1-\lambda $, a translation sampled from $\mathcal{T}_{i}$ is appended (in experiments, $\lambda $ is increased from 0.2 to 0.6 as training progresses), i.e.,
\begin{equation}
p_{i}=x,\hat{t}_{1},\dots ,\hat{t}_{i-1},\; \; \; \hat{t}_{j} \leftarrow \text{Mix}(t_{j}^{\text{(c)}}, t_{j}\sim \mathcal{T}, \lambda).
\end{equation}

\section{Experiments}

\subsection{Experimental Settings}

We run experiments using our customized MuSC dataset, which comprises subtitle corpora from the online video platform Youku for multiple directions, including \texttt{en}$\Rightarrow$\texttt{de}, \texttt{en}$\Rightarrow$\texttt{fr}, \texttt{en}$\Rightarrow$\texttt{zh}, \texttt{ko}$\Rightarrow$\texttt{zh}, \texttt{zh}$\Rightarrow$\texttt{en}, and \texttt{zh}$\Rightarrow$\texttt{th}. Each direction includes 100–200 programs across various genres, with 10\% of the programs reserved as the test set. See Appendix~\ref{sec:dc} for details of MuSC.

% 我们使用自建的MuSC数据集进行实验，其中包含来自在线视频平台Youku的\texttt{en}$\Rightarrow$\texttt{de}, \texttt{en}$\Rightarrow$\texttt{fr}, \texttt{en}$\Rightarrow$\texttt{zh}, \texttt{ko}$\Rightarrow$\texttt{zh}, \texttt{zh}$\Rightarrow$\texttt{en}, \texttt{zh}$\Rightarrow$\texttt{th}等多方向字幕语料，每个方向均包含100-200部各genre的program，我们将其中10\%的节目保留作为测试集。MuSC数据集的详细介绍参看Appendix~\ref{sec:dc}。

We compare our model with the following baselines:
\begin{itemize}
\item \textbf{VideoDubber} \citep{vd2} constructs a length-controlled translation model (the work most closely related to our newly proposed subtitle translation task).
\item \textbf{NLLB-3.3B} \citep{nllb} is a translation model of Meta AI (>200 languages).
\item \textbf{MADLAD-10B} \citep{madlad} is a translation model of Google (>450 languages).
\item \textbf{Google Translate} \citep{google} is a multilingual translation service of Google.
\item \textbf{GPT-4o}, \textbf{Qwen-Max}, and \textbf{DeepSeek-V3.1} \citep{deepseekv3} are current SOTA chat models.
\item \textbf{DeepSeek-R1} and \textbf{GPT-5 Thinking} are SOTA reasoning models. We use the prompt in Appendix~\ref{sec:iosft} to drive these LLMs for subtitle translation.
\item \textbf{Qwen2.5-14B} serves as the backbone for ALPO.
\end{itemize}

Details of the experimental setup are provided in Appendix~\ref{sec:expdetail}. The source code for ALPO and MuSC dataset are available at \url{https://github.com/CcQunResearch/ALPO}.

% 实验设置的细节参看Appendix~\ref{sec:expdetail}。ALPO的source code和MuSC数据集 is available at \url{https://github.com/CcQunResearch/ALPO}。

\subsection{Translation Quality Evaluation}

Unlike the strict requirements for accuracy in the translation of legal texts and technical documents, subtitle translation focuses more on the localized representation of the translation \citep{avt2,legal3}. Specifically, subtitle translation demands not only the conveyance of the original meaning but also expects to reflect the emotions and tone of the original subtitle, aiming for translations that are more fluid, dynamic, and expressive. Given that we have verified the reliability of LLMs as evaluators in Section~\ref{sec:judge}, we developed a multidimensional quality evaluation system for subtitle translation based on LLM-as-a-Judge scoring. We primarily assess three dimensions of subtitle translation: (1) \textit{Accuracy}: whether the translation accurately conveys the original meaning of the subtitle; (2) \textit{Naturalness}: whether the expression in the translation is natural and fluent, aligning with the grammatical structure and lexical conventions of the target language; (3) \textit{Vividness}: whether the translation is expressive and successfully conveys the emotions and atmosphere of the original subtitle. We employ DeepSeek-V3.1, Claude Sonnet 4, and GPT-5 Instant as evaluation models to score subtitle segments in the test set along three dimensions, assigning a score between 0 and 100 for each. Then, we show the average score across all models in Table~\ref{tab:partialeval}.

% 与法律条文和技术文本的翻译对准确性的严格要求不同，subtitle translation更注重译文的本地化表现\citep{avt2,legal3}。具体而言，subtitle translation不仅要求传达原意，同时也期望能够展现出原台词的情感和语气，希望译文更加流畅和生动，更具表现力。鉴于我们在Section~\ref{sec:judge}已经验证了LLM作为evaluator的可靠性，我们搭建了一套基于LLM-as-a-Judge打分的subtitle translation多维度质量评估体系。我们主要评估subtitle translation的三个维度：(1) \textit{Accuracy}：译文是否准确传达了台词原意；(2) \textit{Naturalness}：译文表达是否自然流畅，是否符合目标语言的语法结构和用词习惯；(3) \textit{Vividness}：译文是否具有表现力，是否传达了原台词的情感和氛围。我们使用DeepSeek-V3.1, Claude Sonnet 4, and GPT-5 Instant作为评估模型在这三个维度上分别对测试集中的台词segment打一个0-100的得分，然后我们在Table~\ref{tab:partialeval}中展示了所有模型的均分。

\begin{table*}[!h]
\centering
\caption{Multidimensional quality evaluation. The 1st and 2nd best results are denoted as \colorbox{mlb}{\textbf{blue}} and \colorbox{mlo}{\textbf{orange}}. ST for supervised training. ICL for in-context learning.}
\resizebox{\textwidth}{!}{
\begin{tabular}{llccc|ccc|ccc}
\Xhline{1.0pt}
\rowcolor{gray!20}
 ~ & ~ & \multicolumn{3}{c}{\texttt{en}$\Rightarrow$\texttt{de}} & \multicolumn{3}{c}{\texttt{en}$\Rightarrow$\texttt{fr}} & \multicolumn{3}{c}{\texttt{en}$\Rightarrow$\texttt{zh}} \\
\cline{3-11}
\rowcolor{gray!20}
\multirow{-2}{*}{\textbf{Models}} & \multirow{-2}{*}{\textbf{Training}} & \textbf{Accuracy} & \textbf{Naturalness} & \textbf{Vividness} & \textbf{Accuracy} & \textbf{Naturalness} & \textbf{Vividness} & \textbf{Accuracy} & \textbf{Naturalness} & \textbf{Vividness} \\
\hline
Gold Reference & Human & 84.8 & 83.8 & \colorbox{mlo}{\textbf{73.1}} & 83.5 & 85.0 & 74.8 & 83.6 & 82.6 & \colorbox{mlo}{\textbf{71.5}} \\
\hdashline
VideoDubber & \multirow{4}{*}{ST} & 41.5 & 35.5 & 41.0 & 48.2 & 47.3 & 48.5 & 46.9 & 51.9 & 49.7 \\
NLLB-3.3B & ~ & 75.8 & 70.1 & 59.2 & 76.9 & 74.6 & 61.8 & 61.4 & 54.0 & 43.7 \\
MADLAD-10B & ~ & 73.2 & 65.6 & 54.5 & 73.6 & 67.8 & 57.4 & 59.7 & 55.5 & 46.3 \\
Google Translate & ~ & 89.9 & 80.6 & 62.6 & 91.9 & 84.8 & 64.3 & 84.2 & 79.7 & 54.4 \\
\hdashline
GPT-4o & \multirow{3}{*}{ICL (C)} & 94.1 & 86.7 & 66.9 & 93.2 & 88.8 & 69.5 & 89.3 & 82.3 & 59.8 \\
Qwen-Max & ~ & \colorbox{mlb}{\textbf{95.5}} & \colorbox{mlb}{\textbf{89.2}} & 68.9 & \colorbox{mlo}{\textbf{94.1}} & 89.9 & 71.6 & \colorbox{mlo}{\textbf{91.9}} & 84.4 & 61.3 \\
DeepSeek-V3.1 & ~ & 94.8 & \colorbox{mlo}{\textbf{89.2}} & 67.4 & \colorbox{mlb}{\textbf{94.9}} & \colorbox{mlb}{\textbf{90.7}} & 72.3 & 91.2 & 85.3 & 63.5 \\
\hdashline
DeepSeek-R1 & \multirow{2}{*}{ICL (R)} & 92.8 & 87.4 & 70.0 & 93.6 & 90.0 & 73.4 & 90.5 & \colorbox{mlo}{\textbf{85.7}} & 70.8 \\
GPT-5 & ~ & 93.6 & 88.6 & 72.7 & 92.2 & \colorbox{mlo}{\textbf{90.4}} & \colorbox{mlo}{\textbf{75.8}} & \colorbox{mlb}{\textbf{92.4}} & \colorbox{mlb}{\textbf{87.0}} & 71.1 \\
\hdashline
Qwen2.5-14B & SFT & 87.5 & 83.6 & 64.4 & 86.2 & 85.4 & 67.7 & 86.4 & 82.0 & 59.1 \\
Qwen2.5-14B & \textbf{ALPO} & \colorbox{mlo}{\textbf{95.4}} & 88.4 & \colorbox{mlb}{\textbf{74.8}} & 94.1 & 89.2 & \colorbox{mlb}{\textbf{78.8}} & 90.6 & 84.3 & \colorbox{mlb}{\textbf{76.6}} \\
\hline
\rowcolor{gray!20}
~ & ~ & \multicolumn{3}{c}{\texttt{ko}$\Rightarrow$\texttt{zh}} & \multicolumn{3}{c}{\texttt{zh}$\Rightarrow$\texttt{en}} & \multicolumn{3}{c}{\texttt{zh}$\Rightarrow$\texttt{th}} \\
\cline{3-11}
\rowcolor{gray!20}
\multirow{-2}{*}{\textbf{Models}} & \multirow{-2}{*}{\textbf{Training}} & \textbf{Accuracy} & \textbf{Naturalness} & \textbf{Vividness} & \textbf{Accuracy} & \textbf{Naturalness} & \textbf{Vividness} & \textbf{Accuracy} & \textbf{Naturalness} & \textbf{Vividness} \\
\hline
Gold Reference & Human & 78.0 & 77.8 & \colorbox{mlo}{\textbf{65.8}} & 83.0 & 80.3 & 73.3 & 76.6 & 75.1 & 66.3 \\
\hdashline
VideoDubber & \multirow{4}{*}{ST} & 39.6 & 45.2 & 48.2 & 53.6 & 54.8 & 50.1 & 34.1 & 34.9 & 41.5 \\
NLLB-3.3B & ~ & 33.1 & 26.1 & 25.4 & 29.1 & 21.7 & 20.8 & 42.6 & 33.9 & 40.5 \\
MADLAD-10B & ~ & 44.9 & 42.9 & 46.7 & 45.1 & 38.9 & 37.6 & 47.9 & 50.8 & 51.0 \\
Google Translate & ~ & 54.9 & 52.8 & 52.0 & 79.8 & 66.3 & 50.2 & 55.2 & 56.2 & 54.5 \\
\hdashline
GPT-4o & \multirow{3}{*}{ICL (C)} & 80.0 & 79.9 & 58.1 & 88.5 & 83.0 & 64.6 & 88.0 & 84.4 & 67.9 \\
Qwen-Max & ~ & 83.7 & 82.5 & 61.8 & \colorbox{mlb}{\textbf{90.0}} & 85.0 & 66.8 & \colorbox{mlo}{\textbf{91.3}} & \colorbox{mlb}{\textbf{85.8}} & 69.1 \\
DeepSeek-V3.1 & ~ & 83.1 & 82.2 & 57.2 & \colorbox{mlo}{\textbf{89.5}} & 84.1 & 63.0 & 89.9 & 84.6 & 67.1 \\
\hdashline
DeepSeek-R1 & \multirow{2}{*}{ICL (R)} & 79.8 & 81.6 & 65.6 & 88.5 & 85.6 & 73.5 & 87.6 & 84.0 & 71.0 \\
GPT-5 & ~ & \colorbox{mlb}{\textbf{84.5}} & \colorbox{mlo}{\textbf{82.6}} & 65.0 & 89.1 & \colorbox{mlo}{\textbf{86.1}} & \colorbox{mlo}{\textbf{75.2}} & 88.7 & 83.9 & \colorbox{mlo}{\textbf{73.0}} \\
\hdashline
Qwen2.5-14B & SFT & 80.9 & 76.1 & 53.9 & 85.2 & 80.1 & 54.8 & 87.3 & 82.6 & 66.0 \\
Qwen2.5-14B & \textbf{ALPO} & \colorbox{mlo}{\textbf{84.3}} & \colorbox{mlb}{\textbf{83.3}} & \colorbox{mlb}{\textbf{70.5}} & 88.3 & \colorbox{mlb}{\textbf{86.8}} & \colorbox{mlb}{\textbf{81.7}} & \colorbox{mlb}{\textbf{91.9}} & \colorbox{mlo}{\textbf{84.7}} & \colorbox{mlb}{\textbf{74.2}} \\
\Xhline{1.0pt}
\end{tabular}
}
\label{tab:partialeval}
\end{table*}

The results indicate that ALPO and cutting-edge LLMs like GPT-4o significantly outperform traditional translation models like MADLAD, and surpass human translation in accuracy and naturalness. However, human translation excelled in vividness, possibly suggesting that human translations often incorporate more liberal translation by integrating video content. Reason models achieved a better vividness score compared to chat models, although its accuracy and naturalness were slightly lower in some aspects. This aligns with the conclusions in Section~\ref{sec:chatreason}. Models trained with ALPO demonstrated marked improvement not only in vividness but also in accuracy and naturalness compared to the SFT model. ALPO attained the highest vividness scores across all directions and occasionally surpassed other cutting-edge LLMs in other dimensions, particularly excelling in relatively low-resource language directions such as \texttt{ko}$\Rightarrow$\texttt{zh} and \texttt{zh}$\Rightarrow$\texttt{th}. For the complete results of quality assessment and the prompt design of the LLM evaluator, please refer to Appendix~\ref{sec:fulleval} and Appendix~\ref{sec:ioqe}.

% 结果表明ALPO和GPT-4o等前沿LLM明显优于MADLAD等传统翻译模型，并且在翻译accuracy和naturalness上能够优于human翻译，然而human翻译在vividness上取得了优异的效果，这或许表明人工译文通常会结合视频内容进行更多liberal translation。reason models相比chat models取得了更优的vividness得分，不过在某些方向上accuracy和naturalness会略低于chat模型。这与Section~\ref{sec:chatreason}中结论一致。经过ALPO训练的模型相比SFT模型不仅在vividness上取得了大幅的提升，在accuracy和naturalness上也进步明显。ALPO在所有翻译方向上均取得了最高的vividness得分，在其他维度上也偶有超越前沿LLM，尤其是在\texttt{ko}$\Rightarrow$\texttt{zh}和\texttt{zh}$\Rightarrow$\texttt{th}这些相对低资源语种的方向上取得了全维度的领先。质量评估的完整结果以及LLM evaluator的prompt设计请参看Appendix~\ref{sec:fulleval}和Appendix~\ref{sec:ioqe}。

\subsection{Human Evaluation of Translation Quality}
\label{sec:humaneval}

We present the human evaluation results of the translation quality of ALPO for \texttt{en}$\Rightarrow$\texttt{zh} and \texttt{zh}$\Rightarrow$\texttt{th} in Table~\ref{tab:humaneval}. Due to varying subjective preferences among evaluators, we do not adopt a scoring system. Instead, we perform pairwise comparisons of different translations to assess the win rate metric. We evaluate the ALPO model alongside four baselines across dimensions of accuracy, naturalness, and vividness, and conduct a comprehensive assessment. The human evaluation results are consistent with those shown in Table~\ref{tab:partialeval}, which supports the reliability of our multidimensional evaluation system based on LLM-as-a-Judge. Specific experimental settings can be found in Appendix~\ref{sec:humanevalsetting}.

% 我们在Table~\ref{tab:humaneval}中展示了在\texttt{en}$\Rightarrow$\texttt{zh}和\texttt{zh}$\Rightarrow$\texttt{th}上ALPO翻译质量的human evaluation结果。由于不同评估人员的主观偏好不同，我们不采用打分的形式，而是对不同的翻译进行两两的对比，从而评估win rate指标。我们将ALPO模型与四个baseline在accuracy、naturalness、vividness维度进行评估，并且进行comprehensive评估。human evaluation表现出与Table~\ref{tab:partialeval}一致的结果，这验证了我们基于LLM-as-a-Judge的multidimensional评估体系的可靠性。具体的实验设置参看Appendix~\ref{sec:humanevalsetting}。

\begin{table*}[!h]
\centering
\caption{Human translation quality evaluation, reporting win rate (win:tie:loss). The winning and losing contrasts are marked in \colorbox{mlb}{blue} and \colorbox{mlo}{orange}, respectively.}
\resizebox{\textwidth}{!}{
\begin{tabular}{clcccc|cccc}
\Xhline{1.0pt}
\rowcolor{gray!20}
~ & ~ & \multicolumn{4}{c}{\texttt{en}$\Rightarrow$\texttt{zh}} & \multicolumn{4}{c}{\texttt{zh}$\Rightarrow$\texttt{th}} \\
\cline{3-10}
\rowcolor{gray!20}
\multirow{-2}{*}{\textbf{Challenger}} & \multirow{-2}{*}{\textbf{Competitors}} & \textbf{Accuracy} & \textbf{Naturalness} & \textbf{Vividness} & \textbf{Comprehensive} & \textbf{Accuracy} & \textbf{Naturalness} & \textbf{Vividness} & \textbf{Comprehensive} \\
\hline
\multirow{4}{*}{\makecell{ALPO \\ (Qwen2.5-14B)}} & Gold Reference & \colorbox{mlb}{29:49:22} & \colorbox{mlb}{28:50:22} & \colorbox{mlb}{32:42:26} & \colorbox{mlb}{31:46:23} & \colorbox{mlb}{28:48:24} & \colorbox{mlb}{26:52:22} & \colorbox{mlb}{32:39:29} & \colorbox{mlb}{29:49:22} \\
~ & SFT Model $\pi _{\text{sft}}$ & \colorbox{mlb}{26:50:24} & \colorbox{mlb}{31:48:21} & \colorbox{mlb}{38:41:21} & \colorbox{mlb}{37:43:20} & \colorbox{mlb}{26:55:19} & \colorbox{mlb}{27:51:22} & \colorbox{mlb}{39:42:19} & \colorbox{mlb}{30:50:20} \\
~ & GPT-4o & \colorbox{mlo}{22:54:24} & \colorbox{mlo}{20:57:23} & \colorbox{mlb}{29:51:20} & \colorbox{mlb}{26:54:23} & \colorbox{mlb}{23:57:20} & \colorbox{mlo}{23:51:26} & \colorbox{mlb}{36:37:27} & \colorbox{mlb}{30:45:25} \\
~ & DeepSeek-R1 & \colorbox{mlo}{22:55:23} & \colorbox{mlo}{19:57:24} & \colorbox{mlb}{22:58:20} & \colorbox{mlo}{20:59:21} & \colorbox{mlb}{23:55:22} & \colorbox{mlo}{18:62:20} & \colorbox{mlb}{28:50:22} & \colorbox{mlb}{27:47:24} \\
\Xhline{1.0pt}
\end{tabular}
}
\label{tab:humaneval}
\end{table*}

\subsection{Ablation Study}

\subsubsection{Adaptive Strategy}

% \begin{wraptable}{r}{0.7\textwidth}
% \vspace{-2em}
% \centering
% \caption{Impact of adaptive strategies.}
% \resizebox{0.65\textwidth}{!}{
% \begin{tabular}{lccc|ccc}
% \Xhline{1.0pt}
% \rowcolor{gray!20}
% ~ & \multicolumn{3}{c}{\texttt{en}$\Rightarrow$\texttt{zh}} & \multicolumn{3}{c}{\texttt{zh}$\Rightarrow$\texttt{th}} \\
% \cline{2-7}
% \rowcolor{gray!20}
% ~ & \textbf{Accuracy} & \textbf{Naturalness} & \textbf{Vividness} & \textbf{Accuracy} & \textbf{Naturalness} & \textbf{Vividness} \\
% \hline
% SFT & \textit{86.5} & \textit{82.1} & \textit{59.2} & \textit{87.2} & \textit{82.4} & \textit{66.0} \\
% ALPO & \textit{90.6} & \textit{84.2} & \textit{76.6} & \textit{91.9} & \textit{84.7} & \textit{74.1} \\
% \hdashline
% w/o $w(s_{i})$ & 88.1 & 83.2 & 67.4(↓9.2) & 89.2 & 83.4 & 70.3(↓3.8) \\ 
% \quad w/o $\mathbf{1}(s_{i})$ & 89.1 & 83.4 & 70.2(↓6.4) & 89.9 & 83.9 & 71.2(↓2.9) \\
% \quad w/o $\delta (s_{i})$ & 89.4 & 83.5 & 72.4(↓4.2) & 90.2 & 83.1 & 71.9(↓2.2) \\
% fixed $\beta _{i}$ ($\beta=0.5$) & 90.0 & 83.4 & 74.7(↓1.9) & 89.8 & 83.4 & 72.1(↓2.0) \\
% w/o prefix mixing & 89.5 & 83.7 & 73.4(↓3.2) & 88.8 & 83.7 & 71.6(↓2.5) \\
% \Xhline{1.0pt}
% \end{tabular}
% }
% \label{tab:adaptive}
% \end{wraptable}

We validate the impact of several adaptive strategies in ALPO on model performance in Table~\ref{tab:adaptive}. Results show that the gating function $\mathbf{1}(s_{i})$ has the greatest effect, indicating that the gating mechanism effectively reduces noise from low-diversity irrelevant lines (e.g., simple lines such as “Good morning.”). Moreover, ablating the importance score $\delta(s_{i})$, dynamic $\beta$, and scheduled prefix mixing also leads to varying degrees of performance drop, highlighting the necessity of fine-grained control in training.

% 我们在Table~\ref{tab:adaptive}中验证了ALPO中的多个自适应策略对模型性能的影响。实验结果表明gating function $\mathbf{1}(s_{i})$对性能的影响最大，这表明门控机制能够显著降低优化过程中低多样性无关台词（例如一些简单台词，如“早上好”）的噪声影响。另外，在消融重要性得分$\delta (s_{i})、动态$\beta$以及scheduled prefix mixing等策略后，模型性能也出现不同程度的下降，这表明对模型训练过程进行细粒度处理的必要性。

\begin{table*}[!h]
\centering
\caption{Impact of adaptive strategies.}
\resizebox{\textwidth}{!}{
\begin{tabular}{lccc|ccc}
\Xhline{1.0pt}
\rowcolor{gray!20}
~ & \multicolumn{3}{c}{\texttt{en}$\Rightarrow$\texttt{zh}} & \multicolumn{3}{c}{\texttt{zh}$\Rightarrow$\texttt{th}} \\
\cline{2-7}
\rowcolor{gray!20}
~ & \textbf{Accuracy} & \textbf{Naturalness} & \textbf{Vividness} & \textbf{Accuracy} & \textbf{Naturalness} & \textbf{Vividness} \\
\hline
SFT & \textit{86.5} & \textit{82.1} & \textit{59.2} & \textit{87.2} & \textit{82.4} & \textit{66.0} \\
ALPO & \textit{90.6} & \textit{84.2} & \textit{76.6} & \textit{91.9} & \textit{84.7} & \textit{74.1} \\
\hdashline
w/o $w(s_{i})$ & 88.1 & 83.2 & 67.4(↓9.2) & 89.2 & 83.4 & 70.3(↓3.8) \\ 
\quad w/o $\mathbf{1}(s_{i})$ & 89.1 & 83.4 & 70.2(↓6.4) & 89.9 & 83.9 & 71.2(↓2.9) \\
\quad w/o $\delta (s_{i})$ & 89.4 & 83.5 & 72.4(↓4.2) & 90.2 & 83.1 & 71.9(↓2.2) \\
fixed $\beta _{i}$ ($\beta=0.5$) & 90.0 & 83.4 & 74.7(↓1.9) & 89.8 & 83.4 & 72.1(↓2.0) \\
w/o prefix mixing & 89.5 & 83.7 & 73.4(↓3.2) & 88.8 & 83.7 & 71.6(↓2.5) \\
\Xhline{1.0pt}
\end{tabular}
}
\label{tab:adaptive}
\end{table*}

\subsubsection{Backbone Models}

We compare the performance of different backbone models in Table~\ref{tab:backbone}. The results demonstrate that ALPO enables all backbone models to achieve significant performance improvements compared to the SFT model (Qwen2.5-14B). Notably, the larger Qwen2.5-14B model with more parameters delivers superior and stable outcomes. LLaMA-3.1-8B exhibits underperformance in Chinese-related tasks, resulting in inferior results compared to the other two models in both \texttt{en}$\Rightarrow$\texttt{zh} and \texttt{zh}$\Rightarrow$\texttt{en} translation directions. Nevertheless, it still achieves measurable improvements over the SFT baseline.

% 我们在Table~\ref{tab:backbone}中对比了采用不同的backbone模型的性能对比。结果表明ALPO训练使得不同的backbone模型均取得了相较于SFT模型(Qwen2.5-14B)明显的性能提升。其中，相对更大参数的Qwen2.5-14B模型取得了较为优越且稳定的结果。LLaMA-3.1-8B模型由于在中文上性能欠缺因而在\texttt{en}$\Rightarrow$\texttt{zh}和\texttt{zh}$\Rightarrow$\texttt{en}方向上性能不及另外两个模型，但相比SFT模型也取得了一定提升。

\begin{table*}[!h]
\centering
\caption{Impact of backbone models. The 1st and 2nd best results are marked as \colorbox{mlb}{\textbf{blue}} and \colorbox{mlo}{\textbf{orange}}.}
\resizebox{\textwidth}{!}{
\begin{tabular}{llccc|ccc|ccc}
\Xhline{1.0pt}
\rowcolor{gray!20}
~ & ~ & \multicolumn{3}{c}{\texttt{en}$\Rightarrow$\texttt{de}} & \multicolumn{3}{c}{\texttt{en}$\Rightarrow$\texttt{zh}} & \multicolumn{3}{c}{\texttt{zh}$\Rightarrow$\texttt{en}} \\
\cline{3-11}
\rowcolor{gray!20}
\multirow{-2}{*}{\textbf{Method}} & \multirow{-2}{*}{\textbf{Backbone}} & \textbf{Accuracy} & \textbf{Naturalness} & \textbf{Vividness} & \textbf{Accuracy} & \textbf{Naturalness} & \textbf{Vividness} & \textbf{Accuracy} & \textbf{Naturalness} & \textbf{Vividness} \\
\hline
SFT & - & \textit{87.7} & \textit{83.4} & \textit{64.4} & \textit{86.5} & \textit{82.1} & \textit{59.2} & \textit{85.2} & \textit{80.1} & \textit{54.9} \\
\hdashline
\multirow{3}{*}{ALPO} & LLaMA-3.1-8B \citep{llama} & \colorbox{mlo}{\textbf{94.7}} & \colorbox{mlo}{\textbf{87.2}} & \colorbox{mlo}{\textbf{73.9}} & 88.0 & 83.2 & 72.3 & 87.2 & 85.3 & 77.6\\
~ & GLM4-9B \citep{glm} & 93.2 & 86.1 & 73.2 & \colorbox{mlo}{\textbf{88.2}} & \colorbox{mlb}{\textbf{84.4}} & \colorbox{mlo}{\textbf{74.4}} & \colorbox{mlb}{\textbf{89.2}} & \colorbox{mlo}{\textbf{85.7}} & \colorbox{mlo}{\textbf{78.2}} \\
~ & Qwen2.5-14B \citep{qwen25} & \colorbox{mlb}{\textbf{95.2}} & \colorbox{mlb}{\textbf{88.3}} & \colorbox{mlb}{\textbf{74.8}} & \colorbox{mlb}{\textbf{90.6}} & \colorbox{mlo}{\textbf{84.2}} & \colorbox{mlb}{\textbf{76.6}} & \colorbox{mlo}{\textbf{88.3}} & \colorbox{mlb}{\textbf{86.8}} & \colorbox{mlb}{\textbf{81.6}} \\
\Xhline{1.0pt}
\end{tabular}
}
\label{tab:backbone}
\end{table*}

\subsection{Additional Experiments}

We perform the following extended experiments in the appendix:
1) In Appendix~\ref{sec:fulleval}, we present full results of quality evaluation with further analysis.
2) In Appendix~\ref{sec:demo}, we show demonstration examples of ALPO translations.
3) In Appendix~\ref{sec:furtherab}, We conduct ablation studies for other factors like samping and model size.
4) In Appendix~\ref{sec:furtherexp}, we perform performance evaluation comparing DPO and PPO.
5) In Appendix~\ref{sec:multiturn}, we validate the effectiveness of ALPO in other application task.

% \begin{itemize}
% \item In Appendix~\ref{sec:fulleval}, we present full results of quality evaluation with further analysis.
% \item In Appendix~\ref{sec:demo}, we provide demonstration examples of ALPO translations.
% \item In Appendix~\ref{sec:furtherab}, we investigate the impact of model size and training data volume.
% \item In Appendix~\ref{sec:furtherexp}, we perform performance evaluation comparing vanilla DPO and PPO.
% \item In Appendix~\ref{sec:multiturn}, we validate the effectiveness of ALPO in other application tasks.
% \end{itemize}

% 我们在附录中进行了以下extended experiments of ALPO：
% \begin{itemize}
% \item In Appendix~\ref{sec:fulleval}, 我们展示了质量评估的full result并做了进一步分析。
% \item In Appendix~\ref{sec:demo}, 我们展示了ALPO译文的demo。
% \item In Appendix~\ref{sec:furtherab}, 我们进行了samping size和model size等其他因素的ablation studies
% \item In Appendix~\ref{sec:furtherexp}, 我们进行了vanilla DPO和PPO的性能评估实验。
% \item In Appendix~\ref{sec:multiturn}, 我们验证了ALPO在其他应用任务上的有效性。
% \end{itemize}

\section{Conclusion}

In this study, we investigate literal and liberal translation in subtitle translation as well as other domains, and validate the reliability of large language models as evaluators and reward models for translation quality. Building upon this, we propose the Adaptive Local Preference Optimization (ALPO) method for training expressive and vivid subtitle translation. Our experiments, conducted under the established translation quality evaluation framework, validate the effectiveness of ALPO.

% In this study, 我们调研了subtitle translation领域以及其他领域的literal and liberal translation情况，并验证了LLM作为翻译质量的evaluator和reward model的可靠性。在此基础上我们提出了Adaptive Local Preference Optimization (ALPO)方法，用于训练expressive和vivid的subtitle translation。我们依照我们搭建的翻译质量评估体系进行的实验验证了ALPO方法的有效性。

\subsubsection*{Ethics Statement}

This study utilizes subtitle data obtained exclusively from the online video platform Youku. All data employed in this study were acquired with proper authorization and comply with the terms and conditions set forth by Youku. The data were used solely for academic purposes and were handled in a manner that ensures compliance with relevant ethical and legal standards. 

No personal or sensitive information was collected or processed in this research. All analyses were performed on accessible content provided by Youku, and no attempts were made to infer or reveal any private or identifying information about individuals or entities featured in the materials. 

We affirm that this study adheres to the principles of responsible and ethical research and complies with all applicable institutional and regulatory guidelines.

\subsubsection*{Reproducibility Statement}

We release all code, training scripts, and the MuSC dataset used in this study to facilitate reproducibility and further research. Detailed experimental settings, model configurations, and evaluation protocols are provided in the main text and appendices. The source code and dataset are publicly accessible at \url{https://github.com/CcQunResearch/ALPO}.

\subsubsection*{Acknowledgments}

The authors would like to thank the anonymous reviewers for their helpful comments and suggestions, which have improved the quality of this manuscript.

\bibliography{iclr2026_conference}
\bibliographystyle{iclr2026_conference}

\appendix

\section*{Appendix Contents}

\begin{table}[ht]
    \centering
    \footnotesize
    \begin{tabular}{cl}
    \textbf{Appendix Sections} & \textbf{Contents} \\ \toprule
    \autoref{sec:dc} & \begin{tabular}[c]{@{}l@{}} Data Source and Dataset Construction \end{tabular} \\ \midrule
    \autoref{sec:expdetail} & \begin{tabular}[c]{@{}l@{}} Additional Details of Experimental Settings \end{tabular} \\ \midrule
    \autoref{sec:extendedexp} & \begin{tabular}[c]{@{}l@{}} Extended Evaluation Experiments for ALPO \end{tabular} \\ \midrule
    \autoref{sec:theory} & \begin{tabular}[c]{@{}l@{}} Theory: Adaptive Local Preference Optimization \end{tabular} \\ \midrule
    \autoref{sec:discussion} & \begin{tabular}[c]{@{}l@{}} Further Discussions on ALPO \end{tabular} \\ \midrule
    \autoref{sec:pi} & \begin{tabular}[c]{@{}l@{}} Prompts and Instructions in ALPO \end{tabular} \\
    \bottomrule
    \end{tabular}    
\end{table}

% \clearpage

\section{Dataset Construction}
\label{sec:dc}

In this section, we will introduce the details related to the datasets used for model training.

% In this section, 我们将介绍用于模型训练的数据集相关的细节。

\subsection{Data Sources}

In this study, we utilize the MuSC dataset from the online video platform Youku for experiments, which includes source language subtitles and multilingual translated subtitles of the platform's programs. Table~\ref{tab:subtitle} presents samples of English and Chinese subtitles from the MuSC dataset (in the ass file format), where the "Start" and "End" columns indicate the start and end times of the subtitle within the program. The translation directions covered in MuSC include \texttt{en}$\Rightarrow$\texttt{de}, \texttt{en}$\Rightarrow$\texttt{fr}, \texttt{en}$\Rightarrow$\texttt{zh}, \texttt{ko}$\Rightarrow$\texttt{zh}, \texttt{zh}$\Rightarrow$\texttt{en}, and \texttt{zh}$\Rightarrow$\texttt{th}. These translation directions encompass both cross-language family and cross-language branch scenarios. Subtitle translation, as a data-rich task, makes it easy for leading online video platforms to gather a substantial amount of multilingual subtitle corpus. The MuSC dataset includes over 100 programs in each direction, spanning genres such as film, TV series, documentary, and animation over multiple years. Statistics for MuSC dataset are provided in Table~\ref{tab:sta}.

% In this study，我们使用在线视频网站Youku的影视program的MuSC数据集来进行实验，其中包含站内program的源语言字幕及多语言译文字幕。我们在Table~\ref{tab:subtitle}中展示了MuSC数据集的英文和中文字幕样例（ass文件格式），其中"Start"和"End"列标识了台词在剧集中的起止时间。MuSC中涵盖的翻译方向包括\texttt{en}$\Rightarrow$\texttt{de}、\texttt{en}$\Rightarrow$\texttt{fr}、\texttt{en}$\Rightarrow$\texttt{zh}、\texttt{ko}$\Rightarrow$\texttt{zh}、\texttt{zh}$\Rightarrow$\texttt{en}和\texttt{zh}$\Rightarrow$\texttt{th}。这些翻译方向中既包含cross language family的，也包含cross language branch的。Subtitle translation作为一种data-rich的任务，对于头部在线视频平台而言，收集充足规模的多语言字幕语料是容易的，MuSC数据集中每个方向的节目数量均100+部，包含film, TV series, documentary, animation等多种genre，且横跨多年。MuSC数据集的统计参看Table~\ref{tab:sta}。

Unlike the translation of serious texts in fields such as legislation and medicine, which require a high degree of accuracy \citep{legal1,legal2,legal3}, subtitle text in visual media programs is deeply bound to video and audio modal information, allowing for some tolerance of accuracy loss in translation. However, it is crucial that the translated subtitle accurately conveys the emotions, tone, atmosphere, and cultural context of the original text. Therefore, human translation does not simply involve direct translation of the original subtitle but rather includes polishing, rewriting, and adaptation to meet the localization needs of the target audience \citep{avt1,avt2}. As a result, the gold reference translations produced by human translators are refined versions of the original text, implying that the parallel corpus used for training is not truly "parallel" (the conclusion addressed in Section~\ref{sec:domain}), and there is information asymmetry between the original text and its translation (see Table~\ref{tab:subtitle} for an example). Consequently, while constructing the training set, each direction's program is a native-language program.

% 不同于legislation和medicine领域等严肃文本在翻译时需要高度的准确性\citep{legal1,legal2,legal3}，visual media program中的subtitle文本是与视频、音频模态信息深度绑定的文本，在翻译时可以容忍一定的准确性损失，不过需要译文台词能够精准地传达原台词中的情感、语气、氛围、文化背景等信息，因此人工翻译并非对原台词进行直译，而是经过一定的润色、改写、意译，来适应译文所面向的观众群体的本地化需求\citep{avt1,avt2}。因此，这就使得人工翻译的gold reference译文是在原文基础上经过润色的，即我们训练利用的平行语料并非真的“平行”（Section~\ref{sec:domain}中的结论），其原文与译文是信息不对等的(see Table~\ref{tab:subtitle} for an example)。因此，我们在构造训练集时，每个方向的program都是源语言的本土program。

%% \footnote{\url{https://www.youku.com}}

\begin{table*}[!h]
\centering
\caption{Examples of multilingual subtitles.}
\resizebox{\textwidth}{!}{
\begin{tabular}{ccl}
\Xhline{1.0pt}
\rowcolor{gray!20}
\textbf{Start} & \textbf{End} & \textbf{Text} \\
\hline
0:17:33.25 & 0:17:34.50 & \texttt{Please, let me speak!} \\
0:17:39.54 & 0:17:42.83 & \texttt{As a lower house, my voice doesn't carry much weight here.} \\
0:17:42.91 & 0:17:46.45 & \texttt{But as a mother, I have a voice that matters deeply.} \\
0:17:47.50 & 0:17:50.25 & \texttt{My son isn't in his right mind.} \\
0:17:51.04 & 0:17:53.95 & \texttt{His entire life, he's chased an impossible dream.} \\
0:17:54.91 & 0:17:59.00 & \texttt{What he did was, uh, foolish and unwise.} \\
0:17:59.83 & 0:18:04.20 & \texttt{But he has a good heart. Please, let him come home.} \\
0:18:04.29 & 0:18:08.66 & \texttt{A crime like this can't be overlooked. The boy must be punished.} \\
0:18:08.75 & 0:18:12.16 & \texttt{A violation of the Ethos calls for banishment,} \\
0:18:12.25 & 0:18:16.25 & \texttt{but I can sympathize with a young man's dream to change the world.} \\
0:18:16.83 & 0:18:20.58 & \texttt{Perhaps in this matter, a lesser sentence may suffice.} \\
\hdashline
0:17:42.91 & 0:17:46.50 & \begin{CJK}{UTF8}{gkai}我以母亲的身份 恳请在座的各位议员们听一听\end{CJK} \\
0:17:47.00 & 0:17:50.25 & \begin{CJK}{UTF8}{gkai}我这儿子确实犯了不可饶恕的错误\end{CJK} \\
0:17:51.04 & 0:17:53.95 & \begin{CJK}{UTF8}{gkai}他这辈子 都在追逐一个不可能实现的梦想\end{CJK} \\
0:17:54.91 & 0:17:59.33 & \begin{CJK}{UTF8}{gkai}他的所作所为是很愚蠢 很不明智的\end{CJK} \\
0:17:59.83 & 0:18:04.25 & \begin{CJK}{UTF8}{gkai}但他并没有坏心 求你们 放他回家吧\end{CJK} \\
0:18:04.33 & 0:18:06.70 & \begin{CJK}{UTF8}{gkai}这样严重的罪行不能轻易放过\end{CJK} \\
0:18:06.79 & 0:18:08.66 & \begin{CJK}{UTF8}{gkai}这小子必须受到惩罚\end{CJK} \\
0:18:08.75 & 0:18:12.20 & \begin{CJK}{UTF8}{gkai}违反社会共识的人确实应该遭到驱逐\end{CJK} \\
0:18:12.29 & 0:18:16.25 & \begin{CJK}{UTF8}{gkai}但我也能体会一个年轻人梦想改变世界的雄心\end{CJK} \\
0:18:16.33 & 0:18:20.58 & \begin{CJK}{UTF8}{gkai}就这个案子来说 还是酌情予以轻判为好\end{CJK} \\
\Xhline{1.0pt}
\end{tabular}
}
\label{tab:subtitle}
\end{table*}

\begin{table*}[!h]
\centering
\caption{Statistics of the datasets. FM, TV, DO, and AN are abbreviations for film, TV series, documentary, and animation, respectively.}
\resizebox{\textwidth}{!}{
\begin{tabular}{ccccccc}
\Xhline{1.0pt}
\rowcolor{gray!20}
\textbf{Statistic} & \texttt{en}$\Rightarrow$\texttt{de} & \texttt{en}$\Rightarrow$\texttt{fr} & \texttt{en}$\Rightarrow$\texttt{zh} & \texttt{ko}$\Rightarrow$\texttt{zh} & \texttt{zh}$\Rightarrow$\texttt{en} & \texttt{zh}$\Rightarrow$\texttt{th} \\
\hline
\textbf{cross language family} & \ding{55} & \ding{55} & \ding{51} & \ding{51} & \ding{51} & \ding{51} \\
\textbf{cross language branch} & \ding{55} & \ding{51} & \ding{51} & \ding{51} & \ding{51} & \ding{51} \\
\hdashline
\textbf{period} & 2013-2024 & 2015-2024 & 2020-2024 & 2017-2024 & 2021-2024 & 2021-2024\\
\textbf{\# programs} & 210 & 207 & 161 & 139 & 225 & 218\\
\textbf{program types} & FM,TV,AN & FM,TV,AN & FM,TV,AN & TV & FM,TV,DO & FM,TV,DO\\
\textbf{\# lines} & 1.87M & 1.72M & 1.02M & 1.54M & 1.26M & 1.21M\\
\textbf{total length (h)} & 1156 & 1033 & 852 & 802 & 501 & 479\\
\hdashline
\textbf{\# avg source token} & 7.82 & 7.67 & 7.77 & 9.87 & 6.19 & 6.08 \\
\textbf{\# avg target token} & 10.83 & 10.64 & 6.41 & 6.51 & 8.17 & 21.11 \\
\Xhline{1.0pt}
\end{tabular}
}
\label{tab:sta}
\end{table*}

\subsection{Bilingual Parallel Corpus Construction}
\label{sec:saa}

Typically, the multilingual subtitles for the same program do not correspond line by line. This is mainly due to linguistic differences, which may lead to the merger or splitting of original line during human translation. Additionally, subtitle files may contain noise such as scene titles and annotations. To automate the line-by-line alignment of multilingual subtitles, we designed a subtitle alignment algorithm that utilizes subtitle timing.

% 通常，同一个program的多语言字幕不是逐句对应的，这主要是因为不同语言的信息密度不同，在人工翻译时会对原文台词做合并或者拆分，另外字幕文件中也包含一些场景标题、注释信息等噪声。为了自动化地完成对多语言字幕的逐句对应，我们设计了一种利用台词时间进行逐句对应的subtitle alignment algorithm。

For the source language and target language line sets $\mathcal{L}_{\text{src}}$ and $\mathcal{L}_{\text{tgt}}$, we first determine the row margin $\text{M}=\text{abs}(|\mathcal{L}_{\text{src}}|-|\mathcal{L}_{\text{tgt}}|)$. Then, for each line $\ell_{\text{src}}$ in $\mathcal{L}_{\text{src}}$, we search within a window of size $2\text{M}$ around the corresponding index in $\mathcal{L}_{\text{tgt}}$ for a line whose start time differs from that of $\ell_{\text{src}}$ by within 0.7 seconds to serve as $\ell_{\text{tgt}}$. The reason for choosing 0.7 seconds as the threshold is that the shortest sentence in various languages (such as "Hello") typically has a duration of about 0.7 seconds. Through this process, we can automatically gather a training dataset $\mathbb{D} \equiv \{(\ell_{\text{src}}, \ell_{\text{tgt}}) \in \mathcal{L}_{\text{src}} \times \mathcal{L}_{\text{tgt}}\}$ from bilingual subtitles in programs. The complexity of this process is $O(N)$.

% 对于源语言和目标语言台词集合$\mathcal{L}_{\text{src}}$和$\mathcal{L}_{\text{tgt}}$，我们首先确定其行数margin$\text{M}=\text{abs}(|\mathcal{L}_{\text{src}}|-|\mathcal{L}_{\text{tgt}}|)$，然后对于$\mathcal{L}_{\text{src}}$中的每句台词$\ell_{\text{src}}$，在$\mathcal{L}_{\text{tgt}}$中对应index的上下$2\text{M}$大小的窗口内查找开始时间与$\ell_{\text{src}}$相差0.7s以内的台词，以作为$\ell_{\text{tgt}}$。之所以选择0.7s作为阈值，是因为各种语言中最短的一句话（例如“Hello”）的duration通常在0.7s左右。经过以上过程，我们可以从programs中的双语字幕中全自动地获取到训练数据集$\mathbb{D} \equiv \{(\ell_{\text{src}}, \ell_{\text{tgt}}) \in \mathcal{L}_{\text{src}} \times \mathcal{L}_{\text{tgt}}\}$。该过程的复杂度为$O(N)$。

\section{Additional Details of Experimental Settings}
\label{sec:expdetail}

In this section, we introduce the main settings used in the experiments of this paper. Except for certain experiments with special hyperparameters, the same hyperparameters are adopted to ensure consistency and fairness in experimental comparisons.

% In this section，我们主要介绍本文实验采用的主要设置，除非某些实验存在特殊的超参数，否则均采用相同的超参数以保持实验对比的一致性与公平性。

\subsection{Main Setting}

We primarily employ the PyTorch\footnote{\url{https://github.com/pytorch/pytorch}}, vllm\footnote{\url{https://github.com/vllm-project/vllm}}, and Transformers\footnote{\url{https://github.com/huggingface/transformers}} libraries to implement our approach, while utilizing DeepSpeed\footnote{\url{https://github.com/microsoft/DeepSpeed}} for multi-GPU parallel training. All experiments are conducted on a single machine equipped with eight A800 80GB SXM GPUs. The ALPO sampling phase for each translation direction takes approximately 1.5 hours, while the training phase requires about 2 hours. To ensure fairness and consistency in comparisons, we maintain identical irrelevant parameters across the main experiments, ablation studies, and extended experiments wherever possible. All critical hyperparameter settings are presented in Table~\ref{tab:hp}.

% 我们主要利用PyTorch\footnote{\url{https://github.com/pytorch/pytorch}}、vllm\footnote{\url{https://github.com/vllm-project/vllm}}和Transformers\footnote{\url{https://github.com/huggingface/transformers}}库来implement我们的方法，同时利用DeepSpeed\footnote{\url{https://github.com/microsoft/DeepSpeed}}来进行多GPU并行训练。我们在单机8个A800 80GB SXM GPUs上进行我们的全部实验。每个翻译方向的ALPO采样阶段耗时约1.5小时，训练阶段约2小时。我们在主要实验、ablation studies、extended experiments中尽可能采用一致的无关参数来保证对比的公平性与一致性，所有重要的超参数设置展示在Table~\ref{tab:hp}中。

\begin{table}[h]
\centering
\caption{Hyperparameter configuration in experiments.}
% \resizebox{\textwidth}{!}{
\begin{tabular}{cccl}
 \Xhline{1.0pt}
 \rowcolor{gray!20}
 \textbf{Phase} & \textbf{Hyperparameter} & \textbf{Value} & \textbf{Remark} \\
 \hline
 \multirow{5}{*}{\textbf{SFT}} & n & 35 & Number of lines in a prompt $x$ \\
 ~ & optimizer & AdamW & - \\
 ~ & learning rate & 1e-6 & - \\
 ~ & epoch & 4 & - \\
 ~ & batch size & 96 & - \\
 \hline
 \multirow{8}{*}{\textbf{ALPO}} & $k$ & 15 & Sampling size \\
 ~ & temperature & 1.0 & \multirow{3}{*}{Sampling generation hyperparameters} \\
 ~ & top k & 40 & ~ \\
 ~ & top p & 0.9 & ~ \\
 \cdashline{2-4}
 ~ & optimizer & AdamW & - \\
 ~ & learning rate & 1e-6 & - \\
 ~ & epoch & 1 & - \\
 ~ & batch size & 96 & Segment-level batch size  \\
 % \cdashline{2-4}
 % ~ & $\beta$ & 0.5 & hyperparameter in $\mathcal{L}_{\text{dpo}}$ loss \\
 % ~ & $\eta$ & 0.001 & hyperparameter in $\mathcal{L}_{\text{grpo}}$ loss \\
 \Xhline{1.0pt}
\end{tabular}
% }
\label{tab:hp}
\end{table}

We re-implemented the VideoDubber method and utilized paid APIs from Alibaba Cloud, DeepSeek, OpenAI, and Anthropic to obtain experimental and evaluation results for models such as Qwen, DeepSeek, GPT, and Claude. When eliciting translations from these models on the test set via ICL, we employed prompt in Appendix~\ref{sec:iosft}, with a one-shot example to ensure structured output formatting. All LLMs used in our experiments were the latest versions as of Sep 1, 2025, specifically: qwen-max-2025-01-25\footnote{\url{https://help.aliyun.com/zh/model-studio/what-is-qwen-llm}}, DeepSeek-V3.1\footnote{\url{https://api-docs.deepseek.com/news/news250821}}, gpt-4o-2024-08-06\footnote{\url{https://platform.openai.com/docs/models}}, GPT-5\footnote{\url{https://openai.com/gpt-5}}, and Claude 4/4.1\footnote{\url{https://docs.anthropic.com/en/api/models-list}}.

% 我们re-implement了VideoDubber方法，并且利用Alibaba Cloud、DeepSeek、OpenAI和Anthropic公司提供的付费API来获取Qwen, DeepSeek, GPT, and Claude等模型相关的实验和评估结果。在对这些模型通过ICL获取其在测试集上的译文时，我们采用Table~\ref{tab:prompt}中的prompt，为了使其输出特定格式我们会提供一个one-shot的例子。我们实验应用的多个LLMs均使用截止至2025年9月1日的最新版本，即qwen-max-2025-01-25\footnote{\url{https://help.aliyun.com/zh/model-studio/what-is-qwen-llm}}、DeepSeek-V3-0324\footnote{\url{https://api-docs.deepseek.com}}、gpt-4o-2024-08-06\footnote{\url{https://platform.openai.com/docs/models}}、GPT-5\footnote{\url{https://openai.com/gpt-5}}、Claude 4/4.1\footnote{\url{https://docs.anthropic.com/en/api/models-list}}.

For translation quality evaluation, we reserve 10\% of each dataset's programs as the test set. Due to constraints including paid API costs and inference time, we perform sampling-based evaluation rather than full-scale assessment. From these test programs, we randomly select 2,000 subtitle segments, each containing 35 lines of subtitles, thereby constructing a test set of approximately 70,000 sentences per direction. The final evaluation results represent the average scores across these 2,000 segments.

% 在进行翻译质量评估时，我们将每个翻译方向的数据集的10\%的program保留作为测试集。限于付费API的调用成本以及推理时间等因素，我们并不进行全量evaluation，我们从这些test program中随机挑选2000个台词segment，每个segment包含35句台词，以此为每个方向分别构造一个包含约70000句台词的测试集，最终evaluation结果是2000个segment的平均得分。

\subsection{Experiments in Empirical Investigation}

\subsubsection{Investigation on LLM-as-a-Judge}
\label{sec:settingjudge}

In \texttt{en}$\Rightarrow$\texttt{zh}, \texttt{en}$\Rightarrow$\texttt{de}, \texttt{zh}$\Rightarrow$\texttt{en} and \texttt{zh}$\Rightarrow$\texttt{th} directions, we investigated the evaluation consistency of different LLMs $\pi _{\text{e}}$ (including DeepSeek-R1, GPT-5 Thinking, Claude Opus 4.1, and Qwen3-14B) compared to human evaluators on multiple translations of lines. Specifically, for each direction, each evaluator scored 500 lines $\{s_{i}\mid i\in [n], 1\le i\le 500\}$ across 10 different translations on a scale of 0-100. These translations included human references, Google Translate, an SFT model $\pi _{\text{sft}}$, and multiple LLMs (including DeepSeek-R1, GPT-5 Thinking, Qwen3-235B-A22B Thinking, GPT-4o, DeepSeek-V3, Claude Sonnet 4, and Qwen-Max). The SFT model $\pi _{\text{sft}}$ was trained using the MuSC dataset. The evaluation LLMs assessed translations based on ICL, and since lines are typically closely related to their contextual environment, we also provided the context lines of $s_{i}$ to enable the LLM to reference the semantic environment of $s_{i}$. The prompt is detailed in Appendix~\ref{sec:inputjudge}. The scoring criteria are:
\begin{itemize}
\item Principle 1 (Accuracy): The translation must ensure basic accuracy (Weight: 30\%);
\item Principle 2 (Colloquial Appropriateness): The translation must adhere to the expression habits of human colloquial speech (Weight: 30\%);
\item Principle 3 (Expressive Power): The translation needs to be expressive and vivid (Weight: 40\%).
\end{itemize}
The term "weight" here serves to provide the model with an indicative reference for importance, rather than being utilized for explicit weighting operations.

% 我们在\texttt{en}$\Rightarrow$\texttt{zh}、\texttt{en}$\Rightarrow$\texttt{de}、\texttt{zh}$\Rightarrow$\texttt{en}和\texttt{zh}$\Rightarrow$\texttt{th}翻译方向上调研了不同的评估LLM$\pi _{\text{e}}$（包括DeepSeek-R1, GPT-5 Thinking, Claude Opus 4.1, Qwen3-14B）相比人类evaluator对台词的多个译文的评估一致性。具体的，对于每个方向，每个evaluator对500句台词$\{s_{i}\mid i\in [n], 1\le i\le 500\}$的10个不同译文进行0-100的打分，这些译文包括human reference、Google Translate、SFT模型$\pi _{\text{sft}}$以及多个LLM（包括DeepSeek-R1、GPT-5 Thinking、Qwen3-235B-A22B Thinking、GPT-4o、DeepSeek-V3、Claude Sonnet 4、Qwen-Max）的翻译。其中SFT模型$\pi _{\text{sft}}$使用MuSC dataset训练获得。评估LLM基于ICL实现对译文的评分，由于台词通常与其上下文环境紧密关联，因此我们也会提供$s_{i}$的context lines以供LLM参考$s_{i}$的语义环境，具体的prompt参看Appendix~\ref{sec:inputjudge}。评分标准为：
% \begin{itemize}
% \item Principle 1 (Translation Accuracy): 译文需保证最基本的准确性（权重：30%）；
% \item Principle 2 (Colloquial Appropriateness): 译文需符合人类口语的表达习惯（权重：30%）；
% \item Principle 3 (Expressive Power): 译文需要是expressive和vivid的 (Weight:40%)。
% \end{itemize}
% 这里的weight是为模型提供一个重要性参考，而非用于具体的加权计算。

\begin{wrapfigure}{r}{0.55\textwidth}
    \vspace{-1.3em}
    \subfigure[\texttt{en}$\Rightarrow$\texttt{zh}]{\includegraphics[width=0.27\textwidth]{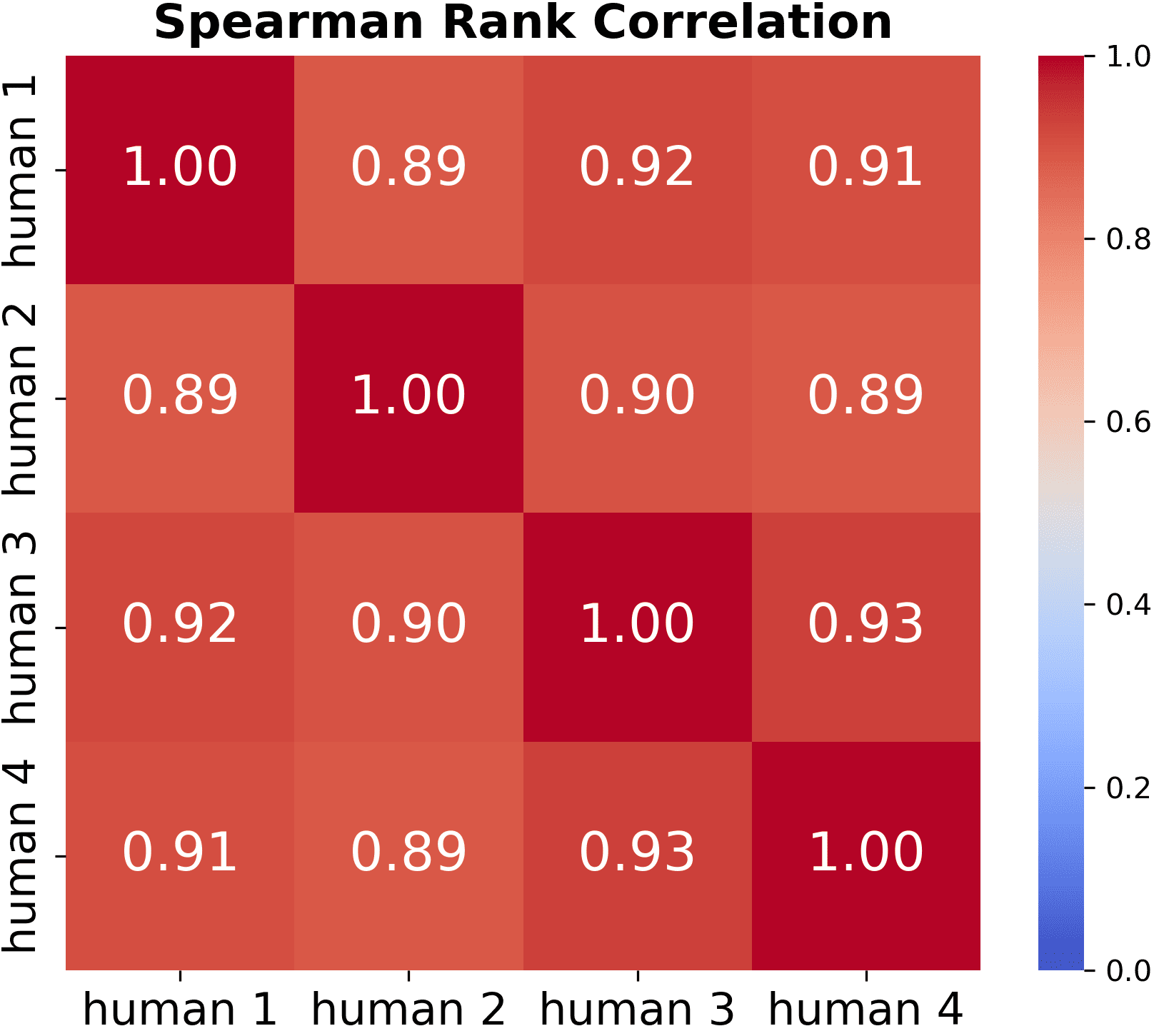}}
    \subfigure[\texttt{zh}$\Rightarrow$\texttt{en}]{\includegraphics[width=0.27\textwidth]{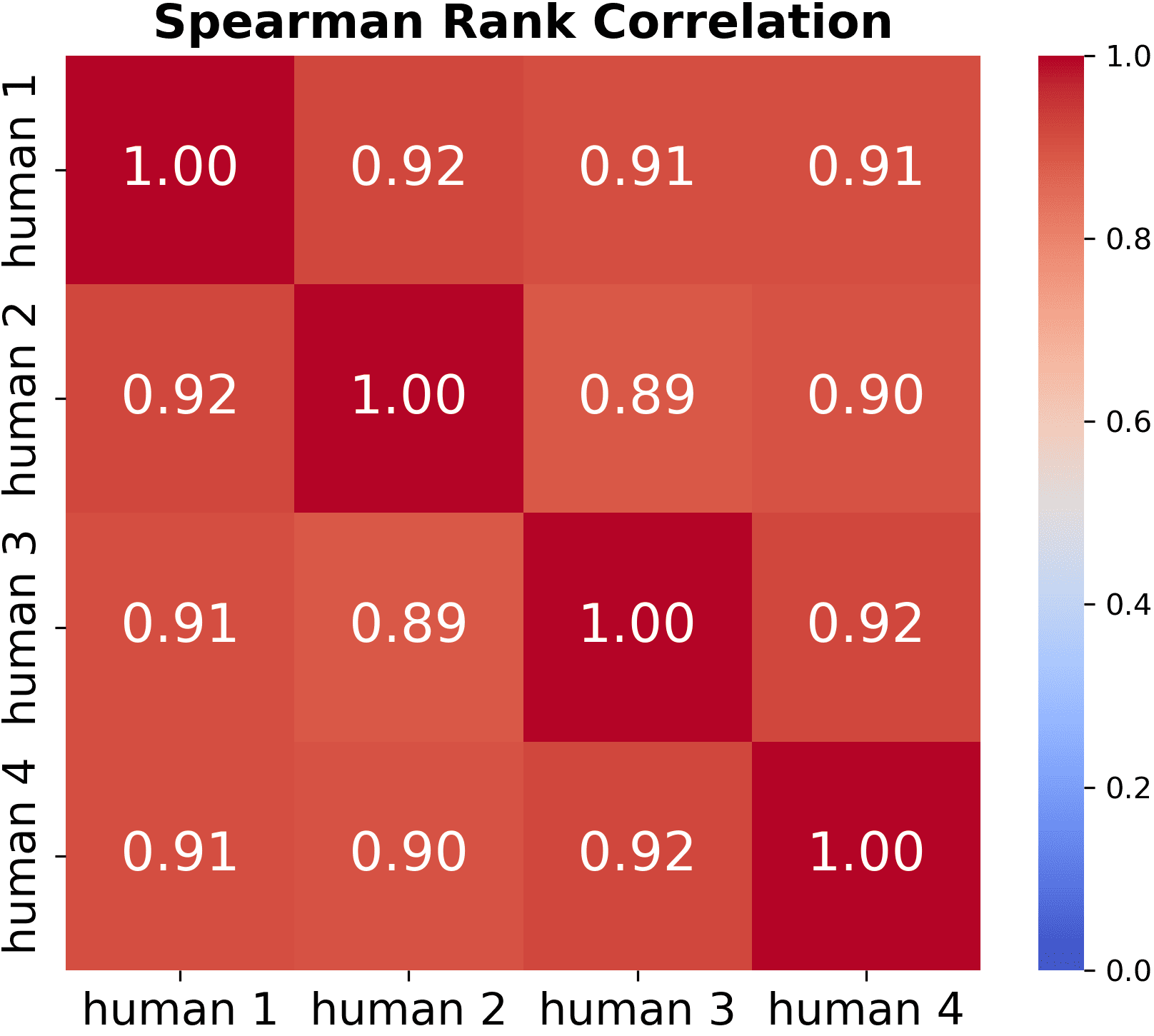}}
    \caption{Human assessment consistency verification.}
    \vspace{-1em}
    \label{fig:humanconsis}
\end{wrapfigure} 

For the \texttt{en}$\Rightarrow$\texttt{zh} and \texttt{zh}$\Rightarrow$\texttt{en} directions, the human evaluator group consisted of four evaluators (the same four individuals), while the \texttt{en}$\Rightarrow$\texttt{de} and \texttt{zh}$\Rightarrow$\texttt{th} directions each included two evaluators. All evaluators were undergraduate or master’s students specializing in English, German, or Thai translation, with Chinese as their native language. The human scores were averaged across evaluators. In Figure~\ref{fig:humanconsis}, we verify the consistency among the four evaluators for \texttt{en}$\Rightarrow$\texttt{zh} and \texttt{zh}$\Rightarrow$\texttt{en}, confirming the reliability of the human evaluation results. For each evaluation LLM $\pi _{\text{e}}$, we shuffled the order of translations four times and took the average across four scores. Each evaluator assigns a score sequence $\mathcal{E}$ to multiple translations of $s_{i}$. Since the scores from different evaluators often fall into different ranges within the 0–100 scale, we focus only on the relative ranking of $\mathcal{E}$ rather than the absolute values. Therefore, we evaluate the Spearman rank correlation $\rho$ between different evaluators instead of using metrics such as ICC that are sensitive to absolute scores.

% \texttt{en}$\Rightarrow$\texttt{zh}和\texttt{zh}$\Rightarrow$\texttt{en}方向的human evaluator包含4名评估人员（相同的四人），\texttt{en}$\Rightarrow$\texttt{de}和\texttt{zh}$\Rightarrow$\texttt{th}方向各包含2名评估人员。这些评估人员均为英语、德语或泰语翻译的本科或硕士专业人员，其母语均为中文，人类打分为多个评估人员的均分。我们在Figure~\ref{fig:humanconsis}中验证\texttt{en}$\Rightarrow$\texttt{zh}和\texttt{zh}$\Rightarrow$\texttt{en}的4名评估人员评估一致性，这表明人类评估结果是可靠的。而对于每个评估LLM$\pi _{\text{e}}$，我们会shuffle译文的顺序四次然后取四次打分的均分。每个evaluator会对$s_{i}$的多个译文给出其打分序列$\mathcal{E}$，在0到100的跨度上由于不同的evaluator的得分通常会位于不同的区间内，因此我们仅关注$\mathcal{E}$的相对排序而不关注其绝对分值。因而我们评估不同evaluator之间的Spearman rank correlation $\rho$，而没有采用受绝对score影响的ICC这类指标。

% \begin{figure*}[!h]
%   \centering
%   \subfigure[\texttt{en}$\Rightarrow$\texttt{zh}]{\includegraphics[width=0.235\textwidth]{img/en2zh_icc_heatmap.png}}
%   \subfigure[\texttt{en}$\Rightarrow$\texttt{de}]{\includegraphics[width=0.235\textwidth]{img/en2de_icc_heatmap.png}}
%   \subfigure[\texttt{zh}$\Rightarrow$\texttt{en}]{\includegraphics[width=0.235\textwidth]{img/zh2en_icc_heatmap.png}}
%   \subfigure[\texttt{zh}$\Rightarrow$\texttt{th}]{\includegraphics[width=0.274\textwidth]{img/zh2th_icc_heatmap.png}}
%   \caption{Investigation of the consistency of multiple evaluators with ICC(3,1).}
%   \label{fig:preferenceicc}
% \end{figure*}

\subsubsection{Investigation on Back-translation Consistency}
\label{sec:settingback}

% \begin{wrapfigure}{r}{0.5\textwidth}
%     % \small
%     \vspace{-1.3em}
%     \includegraphics[width=1.0\linewidth]{img/back_prompt.png}
%     \caption{The \texttt{en}$\Rightarrow$\texttt{de} back-translation prompt demonstration for GPT-4o.}
%     % \vspace{0em}
%     \label{fig:back}
% \end{wrapfigure} 

We investigated the back-translation consistency of parallel corpora from different domains. Specifically, for a translation direction, we utilized GPT-4o with a simple prompt in the following text box to translate the target language text back into the source language. We then calculated the BLEU and ChrF++ scores between the back-translated text and the original text. Lower scores indicate a higher degree of liberal translation. We evaluated 1000 sentence pairs per domain. The datasets from different domains involved in the evaluation include:
\begin{itemize}
\item \textbf{OpenSubtitles} \citep{opensubtitles} is a collection of translated movie subtitles across 60 languages.
\item \textbf{Books} \citep{opus} is a collection of copyright free books in the literature domain.
\item \textbf{bible-uedin} \citep{bible} is a multilingual parallel corpus created from translations of the Bible.
\item \textbf{DGT} \citep{dgt} is a publicly accessible multilingual translation memory of the Acquis Communautaire.
\item \textbf{JRC-Acquis} \citep{jrc} is a collection of legislative texts of the European Union, comprising selected documents written from the 1950s to the present.
\item \textbf{News-Commentary} \citep{opus} is a parallel corpus in the news domain provided by WMT.
\item \textbf{ECDC} \citep{ecdc} is a translation memory from the European Centre for Disease Prevention and Control.
\item \textbf{EMEA} \citep{opus} is a parallel corpus made from documents of the European Medicines Agency.
\end{itemize}

\begin{tcolorbox}[
    title={The \texttt{en}$\Rightarrow$\texttt{de} back-translation prompt demonstration for GPT-4o.},
    colback=blue!5!white,       % 背景：极淡的蓝色
    colframe=blue!75!black,     % 边框：深蓝色
    coltitle=white,             % 标题文字：白色
    breakable,
    fontupper=\small
]
\texttt{Please literally translate the following English text into German, maintaining the highest degree of literal correspondence without any localized rewriting:\newline Now the earth was formless and empty. Darkness was on the surface of the deep. God's Spirit was hovering over the surface of the waters.}
\end{tcolorbox}

\subsection{Human Translation Quality Evaluation}
\label{sec:humanevalsetting}

We conducted human evaluation of ALPO translation quality in the \texttt{en}$\Rightarrow$\texttt{zh} and \texttt{zh}$\Rightarrow$\texttt{th} directions. For each direction, we hired four evaluators, all of whom are undergraduate or master's degree professionals specializing in English or Thai translation, and whose native language is Chinese. We assessed the ALPO model against four baselines (gold reference, SFT model $\pi _{\text{sft}}$, GPT-4o, and DeepSeek-R1) across dimensions of accuracy, naturalness, and vividness, as well as through a comprehensive evaluation. The comprehensive evaluation instructions provided to the evaluators are detailed in Appendix~\ref{sec:ioqe}, with similar instructions for other dimensions. There were a total of 32 evaluation tasks in the multidimensional assessment across both directions. Each evaluator was assigned 4 evaluation tasks. In a single evaluation task, we provided evaluators with challenger and competitor translations of 400 subtitle segments sourced from the MuSC test set, with each segment consisting of 20 subtitle lines. Evaluators were required to choose the better translation between the two options, or mark them as "no significant difference". The subtitle segments selected for different evaluation tasks were random subsets from the test set.

% 我们进行了在\texttt{en}$\Rightarrow$\texttt{zh}和\texttt{zh}$\Rightarrow$\texttt{th}方向上ALPO翻译质量的human evaluation。我们为每个方向雇佣了4名评估人员，均为从事英语或泰语翻译的本科或硕士专业人员，其母语均为中文。我们将ALPO模型与gold reference、SFT model $\pi _{\text{sft}}$、GPT-4o、DeepSeek-R1共四个baseline在accuracy、naturalness、vividness维度进行评估，并且进行comprehensive评估。其中提供给评估人员的comprehensive评估指令参看Appendix~\ref{sec:ioqe}，其他维度也类似。在两个方向上的multidimensional评估一共有32个评估任务，每个评估人员将被分配4个评估任务。在一个评估任务中，我们提供给评估人员来自于MuSC测试集的400个台词segment的challenger译文和competitor译文，每个segment包含20句台词。评估人员需要从中选择两者之间更优的译文，或者两者标记为“no significant difference”。不同的评估任务选取的台词segment均为从测试集中随机选取的子集。

\section{Extended Experiments}
\label{sec:extendedexp}

In this section, we will present additional evaluation experiments for ALPO.

% 在本节中，我们将介绍ALPO的其他评估实验。

\subsection{Full Evaluation of Translation Quality}
\label{sec:fulleval}

We present the full results of the translation quality evaluation in Table~\ref{tab:fulleval}. Our assessment of accuracy includes results from three LLM evaluators as well as traditional translation evaluation model XCOMET \citep{xcomet}. The results in Table~\ref{tab:fulleval} show that Google Translate achieved substantial performance in translation directions within the same language family, such as \texttt{en}$\Rightarrow$\texttt{de} and \texttt{en}$\Rightarrow$\texttt{fr}. However, its performance was poor for other languages, which might be due to the relatively simpler nature of translations between languages of the same language family. Traditional sequence-to-sequence architectures like MADLAD performed significantly worse than modern LLMs, and even VideoDubber, specifically designed for subtitle translation, did not demonstrate noticeably better performance than NLLB and MADLAD models. The comparison between ALPO and cutting-edge LLMs led to several clear conclusions:
\begin{itemize}
\item Different LLM evaluators provide consistent ranking of translations by varying models across multiple dimensions, further validating the conclusions in Section~\ref{sec:judge}.
\item Despite the receiving lower accuracy and naturalness scores, human translations excelled in the vividness dimension, which could indicate that human translators incorporate more interpretive translations based on the video content.
\item Reason models scored higher than chat models in vividness but lower in other dimensions, suggesting that reason models adhere more to liberal translation instructions compared to chat models. This aligns with the conclusion in Section~\ref{sec:chatreason}, as performing liberal translations can compromise accuracy to some extent.
\end{itemize}

% 我们在Table~\ref{tab:fulleval}中展示了翻译质量evaluation的full results。我们在accuracy维度上的评估除了使用三个LLM evaluator以外，也加入了传统翻译评估模型XCOMET \citep{xcomet}的结果。Table~\ref{tab:fulleval}中结果表明Google Translate在\texttt{en}$\Rightarrow$\texttt{de}和\texttt{en}$\Rightarrow$\texttt{fr}等相同language family翻译方向下取得了可观的性能，然而在其他语种上表现均较差，这或许是因为相同language family的语言之间的翻译相对较简单。而MADLAD这些传统的sequence-to-sequence架构模型则远远不如现代LLM，即使是专门面向subtitle translation的VideoDubber也并未展现出明显优于NLLB和MADLAD模型的性能表现。而关于ALPO与前沿LLM的结果对比，有以下几个明显的结论：
% \begin{itemize}
% \item 不同的LLM evaluator在多个维度上对不同模型的译文的评分排序是一致的，这进一步验证了Section~\ref{sec:judge}中的结论。
% \item 尽管人工译文accuracy和naturalness得分较低，但在vividness维度上表现优异，这或许表明人工译文会结合视频内容进行更多意译。
% \item reason模型相比chat模型在vividness维度上得分更高，但在其他维度上相对低一点，这说明reason模型相比chat模型更能服从liberal translation的指令，这与Section~\ref{sec:chatreason}中的结论是一致的，而进行liberal translation也会损耗一定的accuracy。
% \end{itemize}

\begin{table*}[p]
\centering
\caption{Full quality evaluation results. The 1st and 2nd best results are marked as \colorbox{mlb}{\textbf{blue}} and \colorbox{mlo}{\textbf{orange}}, respectively.}
\resizebox{\textwidth}{!}{
\begin{tabular}{llccccc|cccc|cccc}
\Xhline{1.0pt}
\rowcolor{gray!20}
~ & ~ & \multicolumn{13}{c}{\texttt{en}$\Rightarrow$\texttt{de}} \\
\cline{3-15}
\rowcolor{gray!20}
~ & ~ & \multicolumn{5}{c}{\textbf{Accuracy}} & \multicolumn{4}{c}{\textbf{Naturalness}} & \multicolumn{4}{c}{\textbf{Vividness}} \\
\cline{3-15}
\rowcolor{gray!20}
\multirow{-3}{*}{\textbf{Models}} & \multirow{-3}{*}{\textbf{Training}} & \textbf{XCOMET} & \textbf{DeepSeek} & \textbf{Claude} & \textbf{GPT-5} & \textbf{Avg.} & \textbf{DeepSeek} & \textbf{Claude} & \textbf{GPT-5} & \textbf{Avg.} & \textbf{DeepSeek} & \textbf{Claude} & \textbf{GPT-5} & \textbf{Avg.} \\
\hline
Gold Reference & Human & 81.4 & 83.4 & 88.6 & 85.6 & \textsf{84.8} & 83.6 & 85.5 & 82.2 & \textsf{83.8} & 64.4 & \colorbox{mlo}{\textbf{77.1}} & \colorbox{mlo}{\textbf{77.7}} & \colorbox{mlo}{\textbf{\textsf{73.1}}} \\
\hdashline
VideoDubber & \multirow{4}{*}{ST} & 43.2 & 46.1 & 55.1 & 21.4 & \textsf{41.5} & 44.7 & 40.2 & 21.6 & \textsf{35.5} & 42.2 & 47.1 & 33.7 & \textsf{41.0} \\
NLLB-3.3B & ~ & 70.3 & 73.5 & 88.1 & 71.4 & \textsf{75.8} & 72.2 & 75.2 & 63.0 & \textsf{70.1} & 58.2 & 60.4 & 58.9 & \textsf{59.2} \\
MADLAD-10B & ~ & 68.9 & 70.0 & 87.3 & 66.4 & \textsf{73.2} & 68.4 & 71.1 & 57.2 & \textsf{65.6} & 55.2 & 55.7 & 52.6 & \textsf{54.5} \\
Google Translate & ~ & 88.3 & 89.9 & 89.9 & 91.5 & \textsf{89.9} & 80.4 & 81.7 & 79.8 & \textsf{80.6} & 60.9 & 64.2 & 62.8 & \textsf{62.6} \\
\hdashline
GPT-4o & \multirow{3}{*}{ICL (C)} & 91.2 & 96.8 & 96.5 & 91.8 & \textsf{94.1} & 89.1 & 88.7 & 82.2 & \textsf{86.7} & 56.2 & 71.3 & 73.3 & \textsf{66.9} \\
Qwen-Max & ~ & \colorbox{mlo}{\textbf{91.7}} & \colorbox{mlo}{\textbf{97.8}} & \colorbox{mlb}{\textbf{97.9}} & 94.7 & \colorbox{mlb}{\textbf{\textsf{95.5}}} & \colorbox{mlo}{\textbf{89.6}} & \colorbox{mlb}{\textbf{90.1}} & \colorbox{mlo}{\textbf{87.8}} & \colorbox{mlb}{\textbf{\textsf{89.2}}} & 58.0 & 74.0 & 74.7 & \textsf{68.9} \\
DeepSeek-V3.1 & ~ & 90.2 & 97.3 & 96.9 & \colorbox{mlb}{\textbf{94.9}} & \textsf{94.8} & \colorbox{mlb}{\textbf{89.8}} & \colorbox{mlo}{\textbf{89.9}} & \colorbox{mlb}{\textbf{87.9}} & \colorbox{mlo}{\textbf{\textsf{89.2}}} & 56.5 & 71.4 & 74.2 & \textsf{67.4} \\
\hdashline
DeepSeek-R1 & \multirow{2}{*}{ICL (R)} & 89.3 & 94.4 & 94.4 & 93.0 & \textsf{92.8} & 88.0 & 88.7 & 85.5 & \textsf{87.4} & 61.1 & 74.9 & 74.1 & \textsf{70.0} \\
GPT-5 & ~ & 90.7 & 95.7 & 94.6 & 93.4 & \textsf{93.6} & 89.1 & 89.3 & 87.5 & \textsf{88.6} & \colorbox{mlo}{\textbf{64.8}} & 76.5 & 76.9 & \textsf{72.7} \\
\hdashline
Qwen2.5-14B & SFT & 84.7 & 87.6 & 91.1 & 86.7 & \textsf{87.5} & 83.9 & 86.3 & 80.6 & \textsf{83.6} & 56.0 & 68.9 & 68.2 & \textsf{64.4} \\
Qwen2.5-14B & \textbf{ALPO} & \colorbox{mlb}{\textbf{91.9}} & \colorbox{mlb}{\textbf{98.0}} & \colorbox{mlo}{\textbf{97.0}} & \colorbox{mlo}{\textbf{94.8}} & \colorbox{mlo}{\textbf{\textsf{95.4}}} & 88.9 & 89.3 & 87.1 & \textsf{88.4} & \colorbox{mlb}{\textbf{68.3}} & \colorbox{mlb}{\textbf{78.0}} & \colorbox{mlb}{\textbf{78.1}} & \colorbox{mlb}{\textbf{\textsf{74.8}}} \\
\hline
\rowcolor{gray!20}
~ & ~ & \multicolumn{13}{c}{\texttt{en}$\Rightarrow$\texttt{fr}} \\
\cline{3-15}
\rowcolor{gray!20}
~ & ~ & \multicolumn{5}{c}{\textbf{Accuracy}} & \multicolumn{4}{c}{\textbf{Naturalness}} & \multicolumn{4}{c}{\textbf{Vividness}} \\
\cline{3-15}
\rowcolor{gray!20}
\multirow{-3}{*}{\textbf{Models}} & \multirow{-3}{*}{\textbf{Training}} & \textbf{XCOMET} & \textbf{DeepSeek} & \textbf{Claude} & \textbf{GPT-5} & \textbf{Avg.} & \textbf{DeepSeek} & \textbf{Claude} & \textbf{GPT-5} & \textbf{Avg.} & \textbf{DeepSeek} & \textbf{Claude} & \textbf{GPT-5} & \textbf{Avg.} \\
\hline
Gold Reference & Human & 76.4 & 84.4 & 89.0 & 84.1 & \textsf{83.5} & 85.5 & 86.8 & 82.7 & \textsf{85.0} & 68.1 & 78.6 & 77.6 & \textsf{74.8} \\
\hdashline
VideoDubber & \multirow{4}{*}{ST} & 42.3 & 46.9 & 62.1 & 41.7 & \textsf{48.2} & 48.1 & 52.5 & 41.4 & \textsf{47.3} & 47.4 & 53.4 & 44.8 & \textsf{48.5} \\
NLLB-3.3B & ~ & 62.2 & 76.9 & 89.6 & 78.7 & \textsf{76.9} & 75.0 & 77.9 & 70.8 & \textsf{74.6} & 61.4 & 63.3 & 60.7 & \textsf{61.8} \\
MADLAD-10B & ~ & 58.1 & 75.6 & 87.5 & 73.3 & \textsf{73.6} & 70.3 & 71.4 & 61.7 & \textsf{67.8} & 55.7 & 59.4 & 57.1 & \textsf{57.4} \\
Google Translate & ~ & 81.4 & 93.5 & 95.9 & \colorbox{mlb}{\textbf{96.7}} & \textsf{91.9} & 84.5 & 85.8 & 84.2 & \textsf{84.8} & 64.3 & 64.6 & 64.0 & \textsf{64.3} \\
\hdashline
GPT-4o & \multirow{3}{*}{ICL (C)} & 86.4 & \colorbox{mlo}{\textbf{97.7}} & 96.3 & 92.6 & \textsf{93.2} & 89.8 & 90.3 & 86.4 & \textsf{88.8} & 59.0 & 74.9 & 74.7 & \textsf{69.5} \\
Qwen-Max & ~ & 86.1 & 97.5 & \colorbox{mlo}{\textbf{97.3}} & 95.6 & \colorbox{mlo}{\textbf{\textsf{94.1}}} & \colorbox{mlo}{\textbf{90.4}} & 90.4 & 89.0 & \textsf{89.9} & 61.1 & 76.8 & 76.8 & \textsf{71.6} \\
DeepSeek-V3.1 & ~ & \colorbox{mlb}{\textbf{87.6}} & 97.2 & \colorbox{mlb}{\textbf{98.9}} & \colorbox{mlo}{\textbf{95.9}} & \colorbox{mlb}{\textbf{\textsf{94.9}}} & \colorbox{mlb}{\textbf{90.8}} & 90.6 & \colorbox{mlb}{\textbf{90.7}} & \colorbox{mlb}{\textbf{\textsf{90.7}}} & 61.7 & 77.3 & 77.9 & \textsf{72.3} \\
\hdashline
DeepSeek-R1 & \multirow{2}{*}{ICL (R)} & \colorbox{mlo}{\textbf{87.1}} & 96.7 & 95.3 & 95.1 & \textsf{93.6} & 89.7 & \colorbox{mlo}{\textbf{90.8}} & 89.6 & \textsf{90.0} & 64.5 & 77.3 & 78.3 & \textsf{73.4} \\
GPT-5 & ~ & 83.8 & 96.4 & 95.0 & 93.8 & \textsf{92.2} & 90.1 & 90.8 & \colorbox{mlo}{\textbf{90.4}} & \colorbox{mlo}{\textbf{\textsf{90.4}}} & \colorbox{mlo}{\textbf{68.4}} & \colorbox{mlo}{\textbf{79.3}} & \colorbox{mlo}{\textbf{79.8}} & \colorbox{mlo}{\textbf{\textsf{75.8}}} \\
\hdashline
Qwen2.5-14B & SFT & 77.7 & 89.7 & 90.6 & 86.7 & \textsf{86.2} & 85.9 & 86.8 & 83.6 & \textsf{85.4} & 57.4 & 71.7 & 74.0 & \textsf{67.7} \\
Qwen2.5-14B & \textbf{ALPO} & 86.5 & \colorbox{mlb}{\textbf{97.9}} & 96.7 & 95.4 & \textsf{94.1} & 88.7 & \colorbox{mlb}{\textbf{91.5}} & 87.4 & \textsf{89.2} & \colorbox{mlb}{\textbf{73.7}} & \colorbox{mlb}{\textbf{82.0}} & \colorbox{mlb}{\textbf{80.7}} & \colorbox{mlb}{\textbf{\textsf{78.8}}} \\
\hline
\rowcolor{gray!20}
~ & ~ & \multicolumn{13}{c}{\texttt{en}$\Rightarrow$\texttt{zh}} \\
\cline{3-15}
\rowcolor{gray!20}
~ & ~ & \multicolumn{5}{c}{\textbf{Accuracy}} & \multicolumn{4}{c}{\textbf{Naturalness}} & \multicolumn{4}{c}{\textbf{Vividness}} \\
\cline{3-15}
\rowcolor{gray!20}
\multirow{-3}{*}{\textbf{Models}} & \multirow{-3}{*}{\textbf{Training}} & \textbf{XCOMET} & \textbf{DeepSeek} & \textbf{Claude} & \textbf{GPT-5} & \textbf{Avg.} & \textbf{DeepSeek} & \textbf{Claude} & \textbf{GPT-5} & \textbf{Avg.} & \textbf{DeepSeek} & \textbf{Claude} & \textbf{GPT-5} & \textbf{Avg.} \\
\hline
Gold Reference & Human & 79.0 & 84.3 & 88.5 & 82.6 & \textsf{83.6} & 81.6 & 85.7 & 80.6 & \textsf{82.6} & 68.9 & 69.7 & \colorbox{mlo}{\textbf{76.0}} & \colorbox{mlo}{\textbf{\textsf{71.5}}} \\
\hdashline
VideoDubber & \multirow{4}{*}{ST} & 40.9 & 47.5 & 64.1 & 35.0 & \textsf{46.9} & 52.4 & 61.3 & 42.0 & \textsf{51.9} & 49.7 & 52.2 & 47.1 & \textsf{49.7} \\
NLLB-3.3B & ~ & 61.9 & 65.2 & 70.4 & 48.2 & \textsf{61.4} & 59.5 & 56.8 & 45.6 & \textsf{54.0} & 46.3 & 45.2 & 39.5 & \textsf{43.7} \\
MADLAD-10B & ~ & 53.5 & 61.0 & 79.5 & 44.7 & \textsf{59.7} & 58.3 & 65.5 & 42.7 & \textsf{55.5} & 47.4 & 46.7 & 44.8 & \textsf{46.3} \\
Google Translate & ~ & 79.3 & 81.4 & \colorbox{mlb}{\textbf{96.1}} & 80.0 & \textsf{84.2} & 79.4 & 85.1 & 74.6 & \textsf{79.7} & 52.3 & 51.8 & 59.0 & \textsf{54.4} \\
\hdashline
GPT-4o & \multirow{3}{*}{ICL (C)} & 84.1 & 92.1 & 92.6 & 88.6 & \textsf{89.3} & 83.0 & 84.2 & 79.6 & \textsf{82.3} & 56.3 & 57.7 & 65.3 & \textsf{59.8} \\
Qwen-Max & ~ & \colorbox{mlo}{\textbf{88.3}} & 95.3 & 93.5 & 90.3 & \colorbox{mlo}{\textbf{\textsf{91.9}}} & 85.8 & 85.8 & 81.6 & \textsf{84.4} & 59.1 & 61.7 & 63.1 & \textsf{61.3} \\
DeepSeek-V3.1 & ~ & 85.9 & \colorbox{mlo}{\textbf{95.6}} & \colorbox{mlo}{\textbf{93.6}} & 89.5 & \textsf{91.2} & 86.9 & \colorbox{mlo}{\textbf{86.2}} & 82.8 & \textsf{85.3} & 59.5 & 61.8 & 69.2 & \textsf{63.5} \\
\hdashline
DeepSeek-R1 & \multirow{2}{*}{ICL (R)} & 86.9 & 92.3 & 92.9 & 90.0 & \textsf{90.5} & \colorbox{mlo}{\textbf{87.2}} & 86.0 & \colorbox{mlo}{\textbf{83.9}} & \colorbox{mlo}{\textbf{\textsf{85.7}}} & 68.1 & \colorbox{mlo}{\textbf{70.0}} & 74.3 & \textsf{70.8} \\
GPT-5 & ~ & \colorbox{mlb}{\textbf{88.4}} & \colorbox{mlb}{\textbf{96.7}} & 93.6 & \colorbox{mlb}{\textbf{90.8}} & \colorbox{mlb}{\textbf{\textsf{92.4}}} & \colorbox{mlb}{\textbf{88.4}} & \colorbox{mlb}{\textbf{87.1}} & \colorbox{mlb}{\textbf{85.4}} & \colorbox{mlb}{\textbf{\textsf{87.0}}} & \colorbox{mlo}{\textbf{69.2}} & 69.9 & 74.3 & \textsf{71.1} \\
\hdashline
Qwen2.5-14B & SFT & 82.4 & 89.9 & 90.1 & 83.4 & \textsf{86.4} & 83.3 & 83.3 & 79.5 & \textsf{82.0} & 57.0 & 56.0 & 64.2 & \textsf{59.1} \\
Qwen2.5-14B & \textbf{ALPO} & 84.8 & 93.9 & 93.1 & \colorbox{mlo}{\textbf{90.8}} & \textsf{90.6} & 84.2 & 85.0 & 83.8 & \textsf{84.3} & \colorbox{mlb}{\textbf{73.6}} & \colorbox{mlb}{\textbf{76.9}} & \colorbox{mlb}{\textbf{79.2}} & \colorbox{mlb}{\textbf{\textsf{76.6}}} \\
\hline
\rowcolor{gray!20}
~ & ~ & \multicolumn{13}{c}{\texttt{ko}$\Rightarrow$\texttt{zh}} \\
\cline{3-15}
\rowcolor{gray!20}
~ & ~ & \multicolumn{5}{c}{\textbf{Accuracy}} & \multicolumn{4}{c}{\textbf{Naturalness}} & \multicolumn{4}{c}{\textbf{Vividness}} \\
\cline{3-15}
\rowcolor{gray!20}
\multirow{-3}{*}{\textbf{Models}} & \multirow{-3}{*}{\textbf{Training}} & \textbf{XCOMET} & \textbf{DeepSeek} & \textbf{Claude} & \textbf{GPT-5} & \textbf{Avg.} & \textbf{DeepSeek} & \textbf{Claude} & \textbf{GPT-5} & \textbf{Avg.} & \textbf{DeepSeek} & \textbf{Claude} & \textbf{GPT-5} & \textbf{Avg.} \\
\hline
Gold Reference & Human & 68.6 & 80.4 & 88.4 & 74.6 & \textsf{78.0} & 78.8 & 79.9 & 74.7 & \textsf{77.8} & \colorbox{mlo}{\textbf{65.6}} & 58.4 & \colorbox{mlo}{\textbf{73.5}} & \colorbox{mlo}{\textbf{\textsf{65.8}}} \\
\hdashline
VideoDubber & \multirow{4}{*}{ST} & 35.4 & 47.0 & 56.8 & 19.3 & \textsf{39.6} & 48.9 & 58.2 & 28.5 & \textsf{45.2} & 47.4 & 56.0 & 41.1 & \textsf{48.2} \\
NLLB-3.3B & ~ & 30.9 & 39.8 & 40.1 & 21.8 & \textsf{33.1} & 30.3 & 29.5 & 18.5 & \textsf{26.1} & 27.6 & 30.6 & 17.9 & \textsf{25.4} \\
MADLAD-10B & ~ & 37.2 & 47.6 & 66.1 & 28.8 & \textsf{44.9} & 44.7 & 57.3 & 26.8 & \textsf{42.9} & 44.4 & 55.6 & 40.2 & \textsf{46.7} \\
Google Translate & ~ & 42.4 & 50.6 & 76.9 & 49.6 & \textsf{54.9} & 49.5 & 61.8 & 47.2 & \textsf{52.8} & 46.3 & 57.3 & 52.4 & \textsf{52.0} \\
\hdashline
GPT-4o & \multirow{3}{*}{ICL (C)} & 70.5 & 87.5 & 86.5 & 75.4 & \textsf{80.0} & 81.7 & 81.7 & 76.4 & \textsf{79.9} & 55.5 & 51.8 & 67.0 & \textsf{58.1} \\
Qwen-Max & ~ & 74.5 & 89.7 & \colorbox{mlo}{\textbf{88.8}} & \colorbox{mlo}{\textbf{81.7}} & \textsf{83.7} & 83.4 & \colorbox{mlo}{\textbf{83.7}} & \colorbox{mlo}{\textbf{80.5}} & \textsf{82.5} & 57.1 & 57.6 & 70.7 & \textsf{61.8} \\
DeepSeek-V3.1 & ~ & 73.5 & \colorbox{mlo}{\textbf{91.9}} & 87.7 & 79.1 & \textsf{83.1} & \colorbox{mlo}{\textbf{84.7}} & 82.2 & 79.8 & \textsf{82.2} & 55.6 & 49.6 & 66.4 & \textsf{57.2} \\
\hdashline
DeepSeek-R1 & \multirow{2}{*}{ICL (R)} & 71.1 & 86.0 & 86.4 & 75.9 & \textsf{79.8} & 84.0 & 82.3 & 78.6 & \textsf{81.6} & 64.0 & \colorbox{mlo}{\textbf{60.0}} & 72.7 & \textsf{65.6} \\
GPT-5 & ~ & \colorbox{mlb}{\textbf{76.3}} & \colorbox{mlb}{\textbf{93.5}} & 88.7 & 79.5 & \colorbox{mlb}{\textbf{\textsf{84.5}}} & \colorbox{mlb}{\textbf{85.1}} & 83.2 & 79.6 & \colorbox{mlo}{\textbf{\textsf{82.6}}} & 63.1 & 59.3 & 72.5 & \textsf{65.0} \\
\hdashline
Qwen2.5-14B & SFT & 72.9 & 84.8 & 88.1 & 77.7 & \textsf{80.9} & 78.5 & 78.7 & 71.2 & \textsf{76.1} & 53.8 & 47.0 & 61.0 & \textsf{53.9} \\
Qwen2.5-14B & \textbf{ALPO} & \colorbox{mlo}{\textbf{76.0}} & 89.8 & \colorbox{mlb}{\textbf{89.0}} & \colorbox{mlb}{\textbf{82.3}} & \colorbox{mlo}{\textbf{\textsf{84.3}}} & 83.6 & \colorbox{mlb}{\textbf{84.7}} & \colorbox{mlb}{\textbf{81.7}} & \colorbox{mlb}{\textbf{\textsf{83.3}}} & \colorbox{mlb}{\textbf{67.6}} & \colorbox{mlb}{\textbf{67.5}} & \colorbox{mlb}{\textbf{76.3}} & \colorbox{mlb}{\textbf{\textsf{70.5}}} \\
\hline
\rowcolor{gray!20}
~ & ~ & \multicolumn{13}{c}{\texttt{zh}$\Rightarrow$\texttt{en}} \\
\cline{3-15}
\rowcolor{gray!20}
~ & ~ & \multicolumn{5}{c}{\textbf{Accuracy}} & \multicolumn{4}{c}{\textbf{Naturalness}} & \multicolumn{4}{c}{\textbf{Vividness}} \\
\cline{3-15}
\rowcolor{gray!20}
\multirow{-3}{*}{\textbf{Models}} & \multirow{-3}{*}{\textbf{Training}} & \textbf{XCOMET} & \textbf{DeepSeek} & \textbf{Claude} & \textbf{GPT-5} & \textbf{Avg.} & \textbf{DeepSeek} & \textbf{Claude} & \textbf{GPT-5} & \textbf{Avg.} & \textbf{DeepSeek} & \textbf{Claude} & \textbf{GPT-5} & \textbf{Avg.} \\
\hline
Gold Reference & Human & 78.0 & 87.8 & 86.6 & 79.4 & \textsf{83.0} & 82.0 & 81.9 & 77.1 & \textsf{80.3} & 71.9 & 71.1 & 76.9 & \textsf{73.3} \\
\hdashline
VideoDubber & \multirow{4}{*}{ST} & 54.1 & 54.3 & 66.5 & 39.5 & \textsf{53.6} & 59.2 & 60.5 & 44.6 & \textsf{54.8} & 53.9 & 52.8 & 43.6 & \textsf{50.1} \\
NLLB-3.3B & ~ & 35.9 & 30.8 & 36.8 & 12.7 & \textsf{29.1} & 26.4 & 26.1 & 12.5 & \textsf{21.7} & 26.4 & 24.4 & 11.5 & \textsf{20.8} \\
MADLAD-10B & ~ & 41.6 & 49.6 & 56.0 & 33.3 & \textsf{45.1} & 45.7 & 40.9 & 30.2 & \textsf{38.9} & 43.5 & 38.2 & 31.2 & \textsf{37.6} \\
Google Translate & ~ & 77.2 & 75.6 & 88.1 & 78.2 & \textsf{79.8} & 70.0 & 67.6 & 61.3 & \textsf{66.3} & 56.2 & 45.4 & 49.1 & \textsf{50.2} \\
\hdashline
GPT-4o & \multirow{3}{*}{ICL (C)} & 84.8 & 92.3 & 90.3 & 86.7 & \textsf{88.5} & 85.2 & 83.9 & 79.9 & \textsf{83.0} & 61.8 & 63.4 & 68.6 & \textsf{64.6} \\
Qwen-Max & ~ & \colorbox{mlb}{\textbf{86.4}} & \colorbox{mlb}{\textbf{93.2}} & \colorbox{mlb}{\textbf{92.6}} & 87.9 & \colorbox{mlb}{\textbf{\textsf{90.0}}} & \colorbox{mlo}{\textbf{87.0}} & 85.2 & 82.7 & \textsf{85.0} & 63.7 & 67.2 & 69.5 & \textsf{66.8} \\
DeepSeek-V3.1 & ~ & \colorbox{mlo}{\textbf{85.3}} & \colorbox{mlo}{\textbf{92.9}} & \colorbox{mlo}{\textbf{91.4}} & 88.3 & \colorbox{mlo}{\textbf{\textsf{89.5}}} & 85.8 & 84.9 & 81.5 & \textsf{84.1} & 59.0 & 63.8 & 66.2 & \textsf{63.0} \\
\hdashline
DeepSeek-R1 & \multirow{2}{*}{ICL (R)} & 83.2 & 91.6 & 89.6 & \colorbox{mlb}{\textbf{89.8}} & \textsf{88.5} & 85.6 & 83.9 & \colorbox{mlb}{\textbf{87.4}} & \textsf{85.6} & 69.5 & 71.3 & \colorbox{mlo}{\textbf{79.7}} & \textsf{73.5} \\
GPT-5 & ~ & 84.9 & 92.0 & 90.9 & \colorbox{mlo}{\textbf{88.5}} & \textsf{89.1} & \colorbox{mlb}{\textbf{87.2}} & \colorbox{mlo}{\textbf{86.8}} & 84.3 & \colorbox{mlo}{\textbf{\textsf{86.1}}} & \colorbox{mlo}{\textbf{72.7}} & \colorbox{mlo}{\textbf{75.3}} & 77.5 & \colorbox{mlo}{\textbf{\textsf{75.2}}} \\
\hdashline
Qwen2.5-14B & SFT & 80.4 & 88.7 & 88.2 & 83.5 & \textsf{85.2} & 81.9 & 82.0 & 76.3 & \textsf{80.1} & 51.5 & 52.3 & 60.7 & \textsf{54.8} \\
Qwen2.5-14B & \textbf{ALPO} & 85.1 & 90.9 & 89.8 & 87.6 & \textsf{88.3} & 86.1 & \colorbox{mlb}{\textbf{87.1}} & \colorbox{mlo}{\textbf{87.2}} & \colorbox{mlb}{\textbf{\textsf{86.8}}} & \colorbox{mlb}{\textbf{79.2}} & \colorbox{mlb}{\textbf{83.0}} & \colorbox{mlb}{\textbf{82.8}} & \colorbox{mlb}{\textbf{\textsf{81.7}}} \\
\hline
\rowcolor{gray!20}
~ & ~ & \multicolumn{13}{c}{\texttt{zh}$\Rightarrow$\texttt{th}} \\
\cline{3-15}
\rowcolor{gray!20}
~ & ~ & \multicolumn{5}{c}{\textbf{Accuracy}} & \multicolumn{4}{c}{\textbf{Naturalness}} & \multicolumn{4}{c}{\textbf{Vividness}} \\
\cline{3-15}
\rowcolor{gray!20}
\multirow{-3}{*}{\textbf{Models}} & \multirow{-3}{*}{\textbf{Training}} & \textbf{XCOMET} & \textbf{DeepSeek} & \textbf{Claude} & \textbf{GPT-5} & \textbf{Avg.} & \textbf{DeepSeek} & \textbf{Claude} & \textbf{GPT-5} & \textbf{Avg.} & \textbf{DeepSeek} & \textbf{Claude} & \textbf{GPT-5} & \textbf{Avg.} \\
\hline
Gold Reference & Human & 70.5 & 77.9 & 78.0 & 80.0 & \textsf{76.6} & 74.3 & 74.7 & 76.3 & \textsf{75.1} & 60.8 & 67.5 & 70.6 & \textsf{66.3} \\
\hdashline
VideoDubber & \multirow{4}{*}{ST} & 40.2 & 38.2 & 42.5 & 15.7 & \textsf{34.1} & 39.0 & 48.7 & 17.0 & \textsf{34.9} & 43.7 & 49.8 & 31.0 & \textsf{41.5} \\
NLLB-3.3B & ~ & 41.8 & 50.2 & 59.8 & 18.7 & \textsf{42.6} & 48.9 & 41.1 & 11.7 & \textsf{33.9} & 49.3 & 41.7 & 30.5 & \textsf{40.5} \\
MADLAD-10B & ~ & 45.6 & 51.4 & 69.0 & 25.7 & \textsf{47.9} & 54.7 & 64.7 & 33.0 & \textsf{50.8} & 52.7 & 58.1 & 42.2 & \textsf{51.0} \\
Google Translate & ~ & 52.5 & 50.1 & 71.4 & 46.7 & \textsf{55.2} & 53.6 & 65.9 & 49.2 & \textsf{56.2} & 53.5 & 57.6 & 52.5 & \textsf{54.5} \\
\hdashline
GPT-4o & \multirow{3}{*}{ICL (C)} & 80.9 & 90.8 & 89.6 & 90.7 & \textsf{88.0} & \colorbox{mlo}{\textbf{85.1}} & 84.7 & 83.5 & \textsf{84.4} & 62.6 & 67.1 & 73.9 & \textsf{67.9} \\
Qwen-Max & ~ & 84.3 & \colorbox{mlb}{\textbf{95.1}} & \colorbox{mlo}{\textbf{91.4}} & \colorbox{mlo}{\textbf{94.5}} & \colorbox{mlo}{\textbf{\textsf{91.3}}} & \colorbox{mlb}{\textbf{85.9}} & \colorbox{mlb}{\textbf{86.1}} & \colorbox{mlb}{\textbf{85.5}} & \colorbox{mlb}{\textbf{\textsf{85.8}}} & 63.8 & 70.0 & 73.6 & \textsf{69.1} \\
DeepSeek-V3.1 & ~ & \colorbox{mlo}{\textbf{84.8}} & 92.6 & 89.3 & 93.1 & \textsf{89.9} & 85.0 & 85.2 & 83.5 & \textsf{84.6} & 61.8 & 66.6 & 72.8 & \textsf{67.1} \\
\hdashline
DeepSeek-R1 & \multirow{2}{*}{ICL (R)} & 82.2 & 89.2 & 88.1 & 90.8 & \textsf{87.6} & 83.5 & 84.4 & 84.0 & \textsf{84.0} & 64.4 & 72.3 & 76.2 & \textsf{71.0} \\
GPT-5 & ~ & 83.3 & 92.9 & 88.1 & 90.5 & \textsf{88.7} & 85.1 & 82.4 & 84.1 & \textsf{83.9} & \colorbox{mlo}{\textbf{67.3}} & \colorbox{mlo}{\textbf{73.7}} & \colorbox{mlb}{\textbf{78.1}} & \colorbox{mlo}{\textbf{\textsf{73.0}}} \\
\hdashline
Qwen2.5-14B & SFT & 81.7 & 90.0 & 88.8 & 88.7 & \textsf{87.3} & 82.6 & 83.8 & 81.4 & \textsf{82.6} & 59.5 & 66.3 & 72.2 & \textsf{66.0} \\
Qwen2.5-14B & \textbf{ALPO} & \colorbox{mlb}{\textbf{85.0}} & \colorbox{mlo}{\textbf{95.0}} & \colorbox{mlb}{\textbf{92.7}} & \colorbox{mlb}{\textbf{95.1}} & \colorbox{mlb}{\textbf{\textsf{91.9}}} & 84.3 & \colorbox{mlo}{\textbf{85.3}} & \colorbox{mlo}{\textbf{84.4}} & \colorbox{mlo}{\textbf{\textsf{84.7}}} & \colorbox{mlb}{\textbf{69.0}} & \colorbox{mlb}{\textbf{75.4}} & \colorbox{mlo}{\textbf{78.1}} & \colorbox{mlb}{\textbf{\textsf{74.2}}}\\
\Xhline{1.0pt}
\end{tabular}
}
\label{tab:fulleval}
\end{table*}

\subsection{Translation Demonstration}
\label{sec:demo}

In Table~\ref{tab:demo}, we present the translation demonstrations for the \texttt{en}$\Rightarrow$\texttt{zh} and \texttt{zh}$\Rightarrow$\texttt{en} directions including human translations, chat model GPT-4o, reason model GPT-5 Thinking, and our ALPO model. It can be intuitively observed that the translations by GPT-4o tend more towards literal translation and fail to convey the emotional ambiance and stylistic information of the original subtitle through liberal translation like the other translations do. The model trained with ALPO aligns better with the subtitle translation scenario, which includes:
\begin{itemize}
\item The translations are more conversational, resembling human character dialogues rather than formal translations.
\item The wording and phrasing are more vivid, offering expressive articulation.
\item It does not excessively pursue accuracy, allowing for free translation based on context.
\end{itemize}
Furthermore, Table~\ref{tab:demo} illustrates some limitations of LLM translations. LLM translation relies solely on subtitle text and cannot reference information from video context and scenes as human translators do. For instance, LLM translates "nature" to "\begin{CJK}{UTF8}{gkai}自然\end{CJK}", whereas based on the video context, it should be translated to "\begin{CJK}{UTF8}{gkai}天性\end{CJK}". This issue surpasses what preference optimization techniques can address. Therefore, we believe multimodal machine translation might be a valuable direction for advancing subtitle translation research.

% 我们在Table~\ref{tab:demo}中展示了\texttt{en}$\Rightarrow$\texttt{zh}和\texttt{zh}$\Rightarrow$\texttt{en}方向上human、chat model GPT-4o、reason model DeepSeek-R1以及我们的ALPO模型的译文demonstration。可以直观地看出GPT-4o的译文更偏literal translation，不能像其他译文一样能够以liberal translation的方式传达出原台词的情感、氛围等风格信息。经过ALPO训练的模型更能契合subtitle translation场景，这包括：
% \begin{itemize}
% \item 译文更加地口语化，更像是人类角色的对话，而非书面翻译。
% \item 遣词造句更加地生动，表达更具表现力。
% \item 不过分追求准确性，会根据语境自由地翻译。
% \end{itemize}
% 另外，从Table~\ref{tab:demo}中也可看到LLM译文的一些局限性。LLM翻译仅参考字幕文本而不能像人类翻译那样参考视频中的情节、场景等信息，例如LLM将"Nature"翻译为"\begin{CJK}{UTF8}{gkai}自然\end{CJK}"，而按照视频情节信息则应译为"\begin{CJK}{UTF8}{gkai}天性\end{CJK}"，这超出了preference optimization技术所能应对的范畴。因此，我们认为多模态的机器翻译或许是subtitle translation研究的一个有价值的前进方向。

\begin{table*}[hp]
\centering
\caption{Multiple translation demonstration of \texttt{en}$\Rightarrow$\texttt{zh} and \texttt{zh}$\Rightarrow$\texttt{en}. High-quality translations and inadequate translations are marked in \colorbox{mlb}{\textbf{blue}} and \colorbox{mlo}{\textbf{orange}}, respectively.}
\resizebox{\textwidth}{!}{
\begin{tabular}{lllll}
\Xhline{1.0pt}
\rowcolor{gray!20}
 ~ & \multicolumn{4}{c}{\textbf{Translation}} \\
\cline{2-5}
\rowcolor{gray!20}
\multirow{-2}{*}{\textbf{Source}} & \textbf{Human} & \textbf{GPT-4o} & \textbf{GPT-5} & \textbf{ALPO (Ours)} \\
\hline
\texttt{\makecell[l]{Piltover was founded to escape\\the warmongering of mages.}} & \begin{CJK}{UTF8}{gkai}\makecell[l]{当初建立皮尔特沃夫就是\\为了避开那群好战的魔法\\师}\end{CJK} & \begin{CJK}{UTF8}{gkai}\makecell[l]{建立皮尔特沃夫正是为了\\逃离法师的战争}\end{CJK} & \begin{CJK}{UTF8}{gkai}\makecell[l]{皮尔特沃夫是为了逃避法\\师的\colorbox{mlb}{战争狂澜}而建立的}\end{CJK} & \begin{CJK}{UTF8}{gkai}\makecell[l]{皮尔特沃夫的创立是为了\\逃离术士们的\colorbox{mlb}{穷兵黩武}}\end{CJK} \\
\hdashline
\texttt{\makecell[l]{The Arcane is the curse of our\\world. My race was nearly\\destroyed by it.}} & \begin{CJK}{UTF8}{gkai}\makecell[l]{奥术对我们的世界来说就\\是诅咒 我的族人几乎因此\\而灭绝}\end{CJK} & \begin{CJK}{UTF8}{gkai}\makecell[l]{奥术是我们世界的诅咒\\我的族群几乎被它摧毁}\end{CJK} & \begin{CJK}{UTF8}{gkai}\makecell[l]{奥术是我们世界的诅咒\\我的族人几乎因此灭绝}\end{CJK} & \begin{CJK}{UTF8}{gkai}\makecell[l]{奥术是世间的诅咒 我的族\\人曾险些被它\colorbox{mlb}{摧毁殆尽}}\end{CJK}\\
\hdashline
\texttt{\makecell[l]{Councilors Hoskel and Salo.\\Don't they hate each other?}} & \begin{CJK}{UTF8}{gkai}\makecell[l]{霍斯卡尔议员和萨罗议员\\他俩不是\colorbox{mlb}{死对头}吗}\end{CJK} & \begin{CJK}{UTF8}{gkai}\makecell[l]{\colorbox{mlo}{议员们}霍斯克和萨罗\\他们不是互相讨厌吗}\end{CJK} & \begin{CJK}{UTF8}{gkai}\makecell[l]{霍斯克尔议员和萨罗议员\\他俩不是\colorbox{mlb}{水火不容}吗}\end{CJK} & \begin{CJK}{UTF8}{gkai}\makecell[l]{霍斯卡尔议员和萨罗议员\\他们不是\colorbox{mlb}{水火不容}吗}\end{CJK}\\
\hdashline
\texttt{\makecell[l]{His entire life, he's chased an\\impossible dream.}} & \begin{CJK}{UTF8}{gkai}\makecell[l]{他这辈子都在追逐一个不\\可能实现的梦想}\end{CJK} & \begin{CJK}{UTF8}{gkai}\makecell[l]{他一生都在追逐一个不可\\能的梦想}\end{CJK} & \begin{CJK}{UTF8}{gkai}\makecell[l]{他这辈子都在追逐\colorbox{mlb}{不切实}\\\colorbox{mlb}{际}的幻想}\end{CJK} & \begin{CJK}{UTF8}{gkai}\makecell[l]{他一生都在追逐一个\colorbox{mlb}{遥不}\\\colorbox{mlb}{可及}的梦}\end{CJK}\\
\hdashline
\texttt{\makecell[l]{It was the ship of dreams to\\everyone else. To me, it was a\\slave ship.}} & \begin{CJK}{UTF8}{gkai}\makecell[l]{它是所有人心目中的梦想\\之船 对我 它只是一条贩\\奴船}\end{CJK} & \begin{CJK}{UTF8}{gkai}\makecell[l]{对其他人来说 这是一艘梦\\想之船 对我而言 它是一\\艘奴隶船}\end{CJK} & \begin{CJK}{UTF8}{gkai}\makecell[l]{在别人眼中 这是艘梦想之\\船 但对我来说 它却是艘\\奴隶船}\end{CJK} & \begin{CJK}{UTF8}{gkai}\makecell[l]{对旁人来说 这艘船\colorbox{mlb}{承载着}\\\colorbox{mlb}{梦想} 而于我而言 它不过\\是艘奴隶船}\end{CJK}\\
\hdashline
\texttt{\makecell[l]{I would have gone overboard but\\Mr.Dawson here saved me.}} & \begin{CJK}{UTF8}{gkai}\makecell[l]{我差点儿掉下去\\是道森先生救了我}\end{CJK} & \begin{CJK}{UTF8}{gkai}\makecell[l]{如果不是道森先生救了我\\我早就掉下水了}\end{CJK} & \begin{CJK}{UTF8}{gkai}\makecell[l]{要不是这位道森先生救我\\我差点就掉下去了}\end{CJK} & \begin{CJK}{UTF8}{gkai}\makecell[l]{要不是道森先生及时相救\\我差点\colorbox{mlb}{葬身大海}}\end{CJK}\\
\hdashline
\texttt{\makecell[l]{Nature has made us intolerant\\to change, but fortunately, we\\have the capacity to change our\\nature.}} & \begin{CJK}{UTF8}{gkai}\makecell[l]{天性会让我们排斥异变 但\\好在 我们有能力改变自己\\的天性}\end{CJK} & \begin{CJK}{UTF8}{gkai}\makecell[l]{\colorbox{mlo}{大自然}让我们对变化难以\\接受 但幸运的是 我们有能\\力改变我们的本性}\end{CJK} & \begin{CJK}{UTF8}{gkai}\makecell[l]{\colorbox{mlo}{自然}让我们无法适应剧变\\但幸运的是 我们拥有改变\\天性的能力}\end{CJK} & \begin{CJK}{UTF8}{gkai}\makecell[l]{\colorbox{mlo}{自然造物}使我们抗拒改变\\但我们幸而拥有改变天性\\的能力}\end{CJK}\\
\hdashline
\texttt{\makecell[l]{You made me want to see the\\world in more than just black\\and white.}} & \begin{CJK}{UTF8}{gkai}\makecell[l]{你让我超越黑白分明的视\\角看世界}\end{CJK} & \begin{CJK}{UTF8}{gkai}\makecell[l]{你让我想要从不是非黑即\\白的角度看世界}\end{CJK} & \begin{CJK}{UTF8}{gkai}\makecell[l]{你让我学会用更丰富的眼\\光看世界}\end{CJK} & \begin{CJK}{UTF8}{gkai}\makecell[l]{你让我渴望用\colorbox{mlb}{缤纷的色彩}\\而不只是黑白两色来观照\\世界}\end{CJK}\\
\hdashline
\texttt{\makecell[l]{Don't be scared. Think of\\yourself as a caterpillar about\\to transform into a butterfly.}} & \begin{CJK}{UTF8}{gkai}\makecell[l]{别害怕 想象自己是即将变\\成蝴蝶的毛毛虫}\end{CJK} & \begin{CJK}{UTF8}{gkai}\makecell[l]{别害怕 想象你自己是一只\\即将\colorbox{mlb}{化蛹为蝶}的毛毛虫}\end{CJK} & \begin{CJK}{UTF8}{gkai}\makecell[l]{别害怕 把自己想象成即将\\化身成蝶的毛毛虫}\end{CJK} & \begin{CJK}{UTF8}{gkai}\makecell[l]{别害怕 把自己想象成一只\\即将\colorbox{mlb}{破茧成蝶}的毛毛虫}\end{CJK}\\
\hdashline
\texttt{\makecell[l]{As long as we protected each\\other, we would stay immortal…\\and young.}} & \begin{CJK}{UTF8}{gkai}\makecell[l]{只要我们互相保护 就会永\\远年轻}\end{CJK} & \begin{CJK}{UTF8}{gkai}\makecell[l]{只要我们保护彼此 就可以\\\colorbox{mlo}{保持不死} 且年轻}\end{CJK} & \begin{CJK}{UTF8}{gkai}\makecell[l]{只要互相守护 就能\colorbox{mlb}{永葆青}\\\colorbox{mlb}{春}...与不朽}\end{CJK} & \begin{CJK}{UTF8}{gkai}\makecell[l]{只要我们彼此守护 就能\colorbox{mlb}{长}\\\colorbox{mlb}{生不老} \colorbox{mlb}{永葆青春}}\end{CJK}\\
\hdashline
\texttt{\makecell[l]{You two think you're better\\than us, but really, you're\\just a prig and a pig.}} & \begin{CJK}{UTF8}{gkai}\makecell[l]{你们俩以为自己比我们强\\但你们只是猪猡}\end{CJK} & \begin{CJK}{UTF8}{gkai}\makecell[l]{你们俩以为自己比我们好\\但事实上 你们不过就是一\\个\colorbox{mlb}{牛鼻子老学究}和\colorbox{mlo}{一个小}\\\colorbox{mlo}{猪}}\end{CJK} & \begin{CJK}{UTF8}{gkai}\makecell[l]{你们俩自以为高人一等 其\\实一个\colorbox{mlb}{假清高} 一个\colorbox{mlb}{真肥猪}}\end{CJK} & \begin{CJK}{UTF8}{gkai}\makecell[l]{你们俩自以为高人一等 其\\实啊 你俩一个\colorbox{mlb}{道貌岸然}\\一个\colorbox{mlb}{猪狗不如}}\end{CJK}\\
\hdashline
\texttt{\makecell[l]{One day you're on top...\\the next, you're a clown.}} & \begin{CJK}{UTF8}{gkai}\makecell[l]{今日\colorbox{mlb}{风光无限}\\明朝小丑相见}\end{CJK} & \begin{CJK}{UTF8}{gkai}\makecell[l]{有一天你\colorbox{mlo}{在上面}...\\下一天你就是个小丑}\end{CJK} & \begin{CJK}{UTF8}{gkai}\makecell[l]{今天你\colorbox{mlb}{风光无限}...\\明天就沦为小丑}\end{CJK} & \begin{CJK}{UTF8}{gkai}\makecell[l]{你今天还\colorbox{mlb}{高高在上}\\明日就沦为小丑一枚}\end{CJK}\\
\hdashline
\texttt{\makecell[l]{Gotham loves a comeback story.}} & \begin{CJK}{UTF8}{gkai}\makecell[l]{哥谭最喜欢的就是\colorbox{mlb}{英雄回}\\\colorbox{mlb}{归}}\end{CJK} & \begin{CJK}{UTF8}{gkai}\makecell[l]{哥谭市喜欢\colorbox{mlb}{东山再起}的故\\事}\end{CJK} & \begin{CJK}{UTF8}{gkai}\makecell[l]{哥谭市最爱看逆袭故事}\end{CJK} & \begin{CJK}{UTF8}{gkai}\makecell[l]{哥谭人最爱看\colorbox{mlb}{咸鱼翻身}的\\戏码}\end{CJK}\\
\hdashline
\texttt{\makecell[l]{You're the de facto head of\\the Council. People notice\\where you go.}} & \begin{CJK}{UTF8}{gkai}\makecell[l]{你是议会真正的领袖\\行踪备受关注}\end{CJK} & \begin{CJK}{UTF8}{gkai}\makecell[l]{你是议会的实际领导\\人们\colorbox{mlo}{注意你去了哪里}}\end{CJK} & \begin{CJK}{UTF8}{gkai}\makecell[l]{身为议会实际上的话事人\\行踪自然会受关注}\end{CJK} & \begin{CJK}{UTF8}{gkai}\makecell[l]{你是议会事实上的领袖\\一举一动自然引人瞩目}\end{CJK}\\
\hline
\begin{CJK}{UTF8}{gkai}\makecell[l]{我那个时候\\就有一种强烈的预感\\那个球啊 我一定能投进}\end{CJK} & \texttt{\makecell[l]{At that moment, I\\had a really strong\\\colorbox{mlb}{intuition}. That I\\would be able to\\score the shot.}} & \texttt{\makecell[l]{At that moment, I\\had a strong feeling\\that I'd definitely\\make that shot.}} & \texttt{\makecell[l]{Back then, I just\\had this strong\\feeling—that shot,\\you know, I was\\gonna make it.}} & \texttt{\makecell[l]{At that moment, I\\had a strong \colorbox{mlb}{hunch}\\that I'd make that\\shot.}}\\
\hdashline
\begin{CJK}{UTF8}{gkai}\makecell[l]{霜华清光 凝息止气}\end{CJK} & \texttt{\makecell[l]{The luminous frost\\drives away all\\impurities. Seal her\\aura.}} & \texttt{\makecell[l]{Frost's \colorbox{mlo}{pure light},\\calming breath and\\energy.}} & \texttt{\makecell[l]{Frostlight gathers,\\breath stills.}} & \texttt{\makecell[l]{Frost Illume,\\still the breath,\\quell the air.}}\\
\hdashline
\begin{CJK}{UTF8}{gkai}\makecell[l]{人生很多重要的决定\\都是要在这种\\稀里糊涂的时候马上做的}\end{CJK} & \texttt{\makecell[l]{There are many\\important decisions\\in life that we have\\to make immediately\\when the situation\\is still \colorbox{mlb}{blurry}.}} & \texttt{\makecell[l]{Many important\\decisions in life\\have to be made\\in these \colorbox{mlb}{muddled}\\moments.}} & \texttt{\makecell[l]{Many crucial life\\decisions must be\\made precisely in\\these \colorbox{mlb}{muddled} moments\\immediately.}} & \texttt{\makecell[l]{Many crucial\\decisions in life\\must be made\\\colorbox{mlb}{on the spot}, in\\moments of confusion\\and \colorbox{mlb}{bewilderment}.}}\\
\hdashline
\begin{CJK}{UTF8}{gkai}\makecell[l]{现在恐怕还不是\\谈情说爱的时候}\end{CJK} & \texttt{\makecell[l]{I'm afraid this\\isn't the time for\\you two to be\\\colorbox{mlb}{lovey-dovey}.}} & \texttt{\makecell[l]{Now's probably not\\the time for romance.}} & \texttt{\makecell[l]{It's probably\\not the time yet\\to \colorbox{mlo}{talk about love}.}} & \texttt{\makecell[l]{I'm afraid it's\\not the right time\\for \colorbox{mlb}{lovey-dovey} stuff\\just yet.}}\\
\hdashline
\begin{CJK}{UTF8}{gkai}\makecell[l]{你我皆是\\被命运捉弄之人\\为何不与我一起\\颠覆了这世间}\end{CJK} & \texttt{\makecell[l]{Both you and I\\were toyed by fate.\\Why don't you join\\me in \colorbox{mlb}{oversetting}\\this world?}} & \texttt{\makecell[l]{You and I are both\\playthings of fate.\\Why not join me to\\\colorbox{mlb}{overthrow} this world?}} & \texttt{\makecell[l]{Both of us are\\victims of fate's\\cruelty. Why not\\join me to \colorbox{mlb}{overthrow}\\this world?}} & \texttt{\makecell[l]{Both of us are\\pawns in the whims\\of fate. Why not\\join forces with me\\and turn this world\\upside down together?}}\\
\hdashline
\begin{CJK}{UTF8}{gkai}\makecell[l]{只可惜\\这和煦的暖阳\\以后再也看不到了}\end{CJK} & \texttt{\makecell[l]{It's a pity that\\I won't be able\\to feel the warm\\sunshine anymore.}} & \texttt{\makecell[l]{It's such a pity\\that this warm\\sunlight, I will\\never see again.}} & \texttt{\makecell[l]{It’s just a pity\\this warm, \colorbox{mlb}{gentle}\\sunlight—I’ll never\\see it again.}} & \texttt{\makecell[l]{What a shame, I'll\\never get to \colorbox{mlb}{bask}\\\colorbox{mlb}{in} the warm sunshine\\ever again...}}\\
\hdashline
\begin{CJK}{UTF8}{gkai}\makecell[l]{你不要看我赌运差\\但是我看人的眼光\\可是很准的}\end{CJK} & \texttt{\makecell[l]{I might be unlucky\\when it comes to\\gambling, but I'm\\pretty good at\\judging characters.}} & \texttt{\makecell[l]{Don't judge me\\by my bad luck at\\gambling, but my\\eye for people\\is pretty sharp.}} & \texttt{\makecell[l]{Don't judge me by\\my bad gambling\\luck—my judgment\\of people is\\\colorbox{mlb}{razor-sharp}.}} & \texttt{\makecell[l]{Just because I'm\\not a lucky gambler\\doesn't mean I'm bad\\at judging people.\\My instincts are\\\colorbox{mlb}{spot-on}.}}\\
\hdashline
\begin{CJK}{UTF8}{gkai}\makecell[l]{青山常在 绿水长流}\end{CJK} & \texttt{\makecell[l]{The world will\\\colorbox{mlb}{remain the same}.}} & \texttt{\makecell[l]{The mountains \colorbox{mlo}{are}\\\colorbox{mlo}{green}, and the\\waters \colorbox{mlo}{flow long}.}} & \texttt{\makecell[l]{The green hills\\\colorbox{mlb}{endure}, the blue\\rivers flow.}} & \texttt{\makecell[l]{May the mountains\\stand tall and the\\waters flow on.}}\\
\hdashline
\begin{CJK}{UTF8}{gkai}\makecell[l]{她为了保护你被千夫所指\\因为担心你而食不下咽}\end{CJK} & \texttt{\makecell[l]{She's facing a\\\colorbox{mlb}{backlash} for trying\\to keep you safe.\\And she's too\\worried about you\\to eat.}} & \texttt{\makecell[l]{She took all the\\blame to protect\\you, worried sick,\\unable to eat.}} & \texttt{\makecell[l]{She \colorbox{mlb}{endured public}\\\colorbox{mlb}{scorn} to protect you,\\lost her appetite\\worrying about you.}} & \texttt{\makecell[l]{She's been \colorbox{mlb}{vilified}\\for protecting you,\\and she's lost her\\appetite out of\\worry for you.}}\\
\hdashline
\begin{CJK}{UTF8}{gkai}\makecell[l]{她一定有三头六臂}\end{CJK} & \texttt{\makecell[l]{I'm sure she has\\many arms and legs.}} & \texttt{\makecell[l]{She must have\\three heads and\\six arms.}} & \texttt{\makecell[l]{She must have three\\heads and six arms\\or something.}} & \texttt{\makecell[l]{She must be a\\\colorbox{mlb}{superwoman}.}}\\
\hdashline
\begin{CJK}{UTF8}{gkai}\makecell[l]{青丘军队全部停止前行\\原地扎寨 不得违令}\end{CJK} & \texttt{\makecell[l]{All the Qingqiu\\troops shall stop\\advancing and camp\\where you are.\\The order must\\be obeyed.}} & \texttt{\makecell[l]{The Qingqiu army\\must halt and set\\up camp immediately,\\no \colorbox{mlb}{disobedience}\\allowed.}} & \texttt{\makecell[l]{All Qingqiu troops\\halt your advance,\\set camp immediately.\\\colorbox{mlb}{Disobedience} forbidden.}} & \texttt{\makecell[l]{Qingqiu's armies,\\halt your advance\\at once. Set up\\camp where you stand.\\No \colorbox{mlb}{disobedience}\\allowed.}}\\
\hdashline
\begin{CJK}{UTF8}{gkai}\makecell[l]{人间的桂花开了}\end{CJK} & \texttt{\makecell[l]{The osmanthus in\\the human realm\\is blooming.}} & \texttt{\makecell[l]{The \colorbox{mlo}{osmanthus flowers}\\are blooming in\\the Mortal Realm.}} & \texttt{\makecell[l]{The osmanthus in the\\Mortal Realm is\\blooming.}} & \texttt{\makecell[l]{The osmanthus blooms\\in the Mortal Realm.}}\\
\Xhline{1.0pt}
\end{tabular}
}
\label{tab:demo}
\end{table*}

\subsection{Extended Ablation Study}
\label{sec:furtherab}

\subsubsection{Alignment Loss}

When simplified, ALPO can be viewed as a framework compatible with various plug-and-play preference optimization losses. Specifically, the ALPO loss can be expressed as:
\begin{equation}
\mathcal{L}_{\text{alpo}}(\pi _{\theta };\pi _{\text{ref}})=-\mathbb{E}_{(x,\mathcal{S}(x))\sim \mathbb{D}_{\text{alpo}}}\left (\sum_{i=1}^{n}w(s_{i})\cdot \mathcal{L}_{\text{po}}(s_{i})\right ),
\end{equation}
where $\mathcal{L}_{\text{po}}$ can adopt multiple types of preference optimization losses, such as DPO \citep{dpo}, SimPO \citep{po17}, and GRPO \citep{grpo}. For the advantage-based GRPO loss, we compute it as:
\begin{equation}
\begin{split}
&\mathcal{L}_{\text{po}}(s_{i})=\\&\frac{1}{|\mathcal{T}_{i}|}\sum _{j}\Bigg[\min \Bigg(\frac{\pi _{\theta }(t_{i}^{j}\mid p_{i})}{\pi _{\text{ref}}(t_{i}^{j}\mid p_{i})}A_{j},\text{clip}\bigg(\frac{\pi _{\theta }(t_{i}^{j}\mid p_{i})}{\pi _{\text{ref}}(t_{i}^{j}\mid p_{i})},1-\varepsilon ,1+\varepsilon \bigg)A_{j}\Bigg)-\eta \text{KL}(\pi _{\theta }||\pi _{\text{ref}})\Bigg],
\end{split}
\end{equation}
\begin{equation}
A_{j}=\frac{\mathcal{E}_{i}^{j}-\text{mean}(\mathcal{E}_{i})}{\text{std}(\mathcal{E}_{i})},\; \; \; \; \text{KL}(\pi _{\theta }||\pi _{\text{ref}})=\frac{\pi _{\text{ref}}(t_{i}^{j}\mid p_{i})}{\pi _{\theta }(t_{i}^{j}\mid p_{i})}-\text{log}\frac{\pi _{\text{ref}}(t_{i}^{j}\mid p_{i})}{\pi _{\theta }(t_{i}^{j}\mid p_{i})}-1,
\end{equation}
where $\varepsilon$ and $\eta$ are adjustable hyperparameters, and $A_{j}$ denotes the advantage estimated from the score sequence $\mathcal{E}_{i}$.

% 当我们简化一下ALPO的损失，则可以将其作为一种兼容多种plug-and-play偏好优化损失的框架。具体的，我们将ALPO损失写成以下形式：
% \begin{equation}
% \mathcal{L}_{\text{alpo}}(\pi _{\theta };\pi _{\text{ref}})=-\mathbb{E}_{(x,\mathcal{S}(x))\sim \mathbb{D}_{\text{alpo}}}\left (\sum_{i=1}^{n}w(s_{i})\cdot \mathcal{L}_{\text{po}}(s_{i})\right ),
% \end{equation}
% 这里的$\mathcal{L}_{\text{po}}$可以采用多种类型的损失，例如DPO \citep{dpo}、SimPO \citep{po17}以及GRPO \citep{grpo}。对于基于advantage的GRPO损失，我们如下计算其损失：
% \begin{equation}
% \begin{split}
% &\mathcal{L}_{\text{po}}(s_{i})=\\&\frac{1}{|\mathcal{T}_{i}|}\sum _{j}\Bigg[\min \Bigg(\frac{\pi _{\theta }(t_{i}^{j}\mid p_{i})}{\pi _{\text{ref}}(t_{i}^{j}\mid p_{i})}A_{j},\text{clip}\bigg(\frac{\pi _{\theta }(t_{i}^{j}\mid p_{i})}{\pi _{\text{ref}}(t_{i}^{j}\mid p_{i})},1-\varepsilon ,1+\varepsilon \bigg)A_{j}\Bigg)-\eta \text{KL}(\pi _{\theta }||\pi _{\text{ref}})\Bigg],
% \end{split}
% \end{equation}
% \begin{equation}
% A_{j}=\frac{\mathcal{E}_{i}^{j}-\text{mean}(\mathcal{E}_{i})}{\text{std}(\mathcal{E}_{i})},\; \; \; \; \text{KL}(\pi _{\theta }||\pi _{\text{ref}})=\frac{\pi _{\text{ref}}(t_{i}^{j}\mid p_{i})}{\pi _{\theta }(t_{i}^{j}\mid p_{i})}-\text{log}\frac{\pi _{\text{ref}}(t_{i}^{j}\mid p_{i})}{\pi _{\theta }(t_{i}^{j}\mid p_{i})}-1,
% \end{equation}
% where $\varepsilon$ and $\eta$为hyperparameters，$A_{j}$是基于得分序列$\mathcal{E}_{i}$估计的advantage。

We compare the performance of ALPO with different preference alignment losses $\mathcal{L}_{\text{po}}$ in Table~\ref{tab:loss}. The experimental results demonstrate that all loss variants achieve substantial improvements over the SFT model, validating ALPO's broad compatibility. The DPO loss exhibits stable and superior performance across multiple translation directions. While GRPO effectively utilizes all sampled outputs rather than focusing solely on preference pairs, it only achieves optimal performance in the \texttt{en}$\Rightarrow$\texttt{zh} direction. 

% 我们在Table~\ref{tab:loss}中对比了ALPO采用不同的偏好对齐loss $\mathcal{L}_{\text{po}}$的性能对比。结果显示不同的损失均相比SFT模型取得了可观的提升，这验证了ALPO广泛的兼容性。DPO损失在多个方向上均展示出稳定且优越的性能表现，GRPO能够有效地利用所有采样结果而非仅关注偏好pair，尽管如此其也只是在\texttt{en}$\Rightarrow$\texttt{zh}一个方向上取得了最优性能。

\begin{table*}[!h]
\centering
\caption{Impact of $\mathcal{L}_{\text{po}}$. The 1st and 2nd best results are denoted as \colorbox{mlb}{\textbf{blue}} and \colorbox{mlo}{\textbf{orange}}.}
\resizebox{\textwidth}{!}{
\begin{tabular}{llccc|ccc|ccc}
\Xhline{1.0pt}
\rowcolor{gray!20}
~ & ~ & \multicolumn{3}{c}{\texttt{en}$\Rightarrow$\texttt{de}} & \multicolumn{3}{c}{\texttt{en}$\Rightarrow$\texttt{zh}} & \multicolumn{3}{c}{\texttt{zh}$\Rightarrow$\texttt{en}} \\
\cline{3-11}
\rowcolor{gray!20}
\multirow{-2}{*}{\textbf{Method}} & \multirow{-2}{*}{$\mathcal{L}_{\text{po}}$} & \textbf{Accuracy} & \textbf{Naturalness} & \textbf{Vividness} & \textbf{Accuracy} & \textbf{Naturalness} & \textbf{Vividness} & \textbf{Accuracy} & \textbf{Naturalness} & \textbf{Vividness} \\
\hline
SFT & - & 87.7 & 83.4 & 64.4 & 86.5 & 82.1 & 59.2 & 85.2 & 80.1 & 54.9 \\
\hdashline
\multirow{3}{*}{ALPO} & DPO & \colorbox{mlb}{\textbf{95.2}} & \colorbox{mlb}{\textbf{88.3}} & \colorbox{mlb}{\textbf{74.8}} & 90.6 & \colorbox{mlo}{\textbf{84.2}} & \colorbox{mlo}{\textbf{76.6}} & \colorbox{mlb}{\textbf{88.3}} & \colorbox{mlb}{\textbf{86.8}} & \colorbox{mlb}{\textbf{81.6}} \\
~ & SimPO & 93.2 & 87.4 & 73.9 & \colorbox{mlo}{\textbf{90.7}} & 83.9 & 74.3 & 86.2 & 84.2 & 80.9 \\
~ & GRPO & \colorbox{mlo}{\textbf{94.6}} & \colorbox{mlo}{\textbf{87.9}} & \colorbox{mlo}{\textbf{74.3}} & \colorbox{mlb}{\textbf{91.2}} & \colorbox{mlb}{\textbf{85.0}} & \colorbox{mlb}{\textbf{77.1}} & \colorbox{mlo}{\textbf{87.2}} & \colorbox{mlo}{\textbf{84.9}} & \colorbox{mlo}{\textbf{81.3}} \\
\Xhline{1.0pt}
\end{tabular}
}
\label{tab:loss}
\end{table*}

\subsubsection{Sampling Size}

\begin{wrapfigure}{r}{0.4\textwidth}
    \vspace{-1.0em}
    \includegraphics[width=1.0\linewidth]{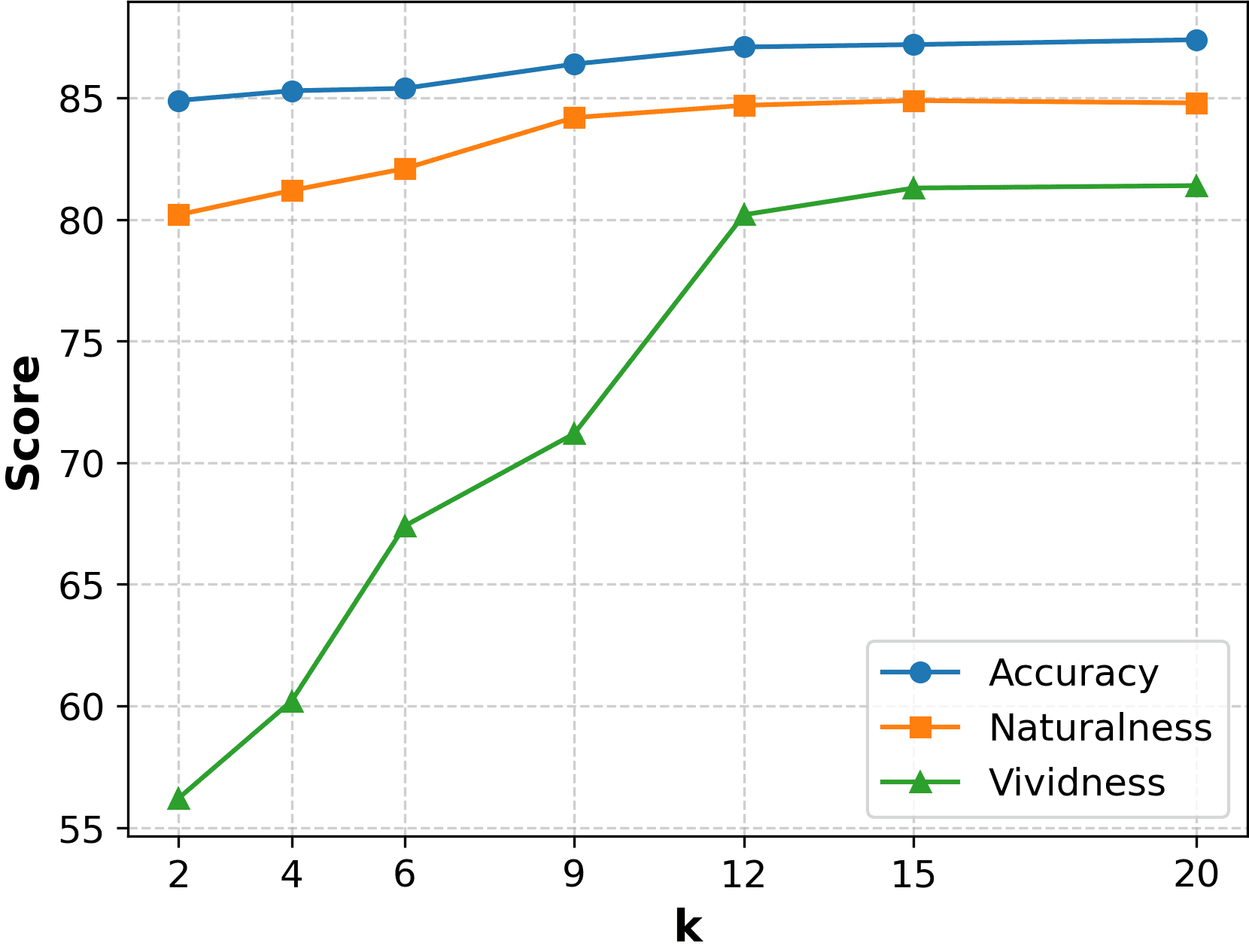}
    \caption{Impact of sampling size on translation quality.}
    \vspace{-1em}
    \label{fig:sampingsize}
\end{wrapfigure}

The sampling size of ALPO directly affects the diversity of sampled translations. In Figure~\ref{fig:sampingsize}, we examine the impact of sampling size $k$ on translation quality in the \texttt{zh}$\Rightarrow$\texttt{en} direction. Results show that as $k$ increases, translation quality improves across multiple dimensions, with the most significant gain observed in vividness, and stabilizes around $k=12$. The computation of adaptive hyperparameters $w(s_{i})$ and $\beta _{i}$ during ALPO training depends on the number and quality variation of sampled translations. A larger sampling size directly improves the quality of training translations and leads to more suitable $w(s_{i})$ and $\beta _{i}$. Since ALPO loss relies on the relative quality of chosen and rejected translations, performance stabilizes once the sampling size is sufficient. Based on these results, we set $k=15$ for all other experiments.

% ALPO采样的size直接影响采样译文的丰富度，我们在Figure~\ref{fig:sampingsize}中探究了\texttt{zh}$\Rightarrow$\texttt{en}方向上sampling size $k$对翻译质量的影响。结果表明随着$k$的提升，多个维度上的翻译质量都有所提升，其中vividness维度提升幅度最明显，到$k=12$趋于稳定。ALPO训练过程中的$w(s_{i})$和$\beta _{i}$这些自适应参数的计算依赖于采样译文的数量和质量差异，更大的采样size会直接影响到用来训练的译文的质量并且能够获得更合适的$w(s_{i})$和$\beta _{i}$。ALPO loss依赖于采样的chosen译文和rejected译文的相对质量，因此当采样数量充足时，其性能会趋于稳定。根据实验结果，我们在其他实验中统一设置$k=15$。

\subsubsection{Model Size}

% \begin{wrapfigure}{r}{0.6\textwidth}
%     \vspace{-1.2em}
%     \includegraphics[width=1.0\linewidth]{img/model_size.png}
%     \caption{Impact of model size on translation quality.}
%     \vspace{-1em}
%     \label{fig:modelsize}
% \end{wrapfigure} 

\begin{figure*}[!h]
  \centering
  \includegraphics[width=0.6\textwidth]{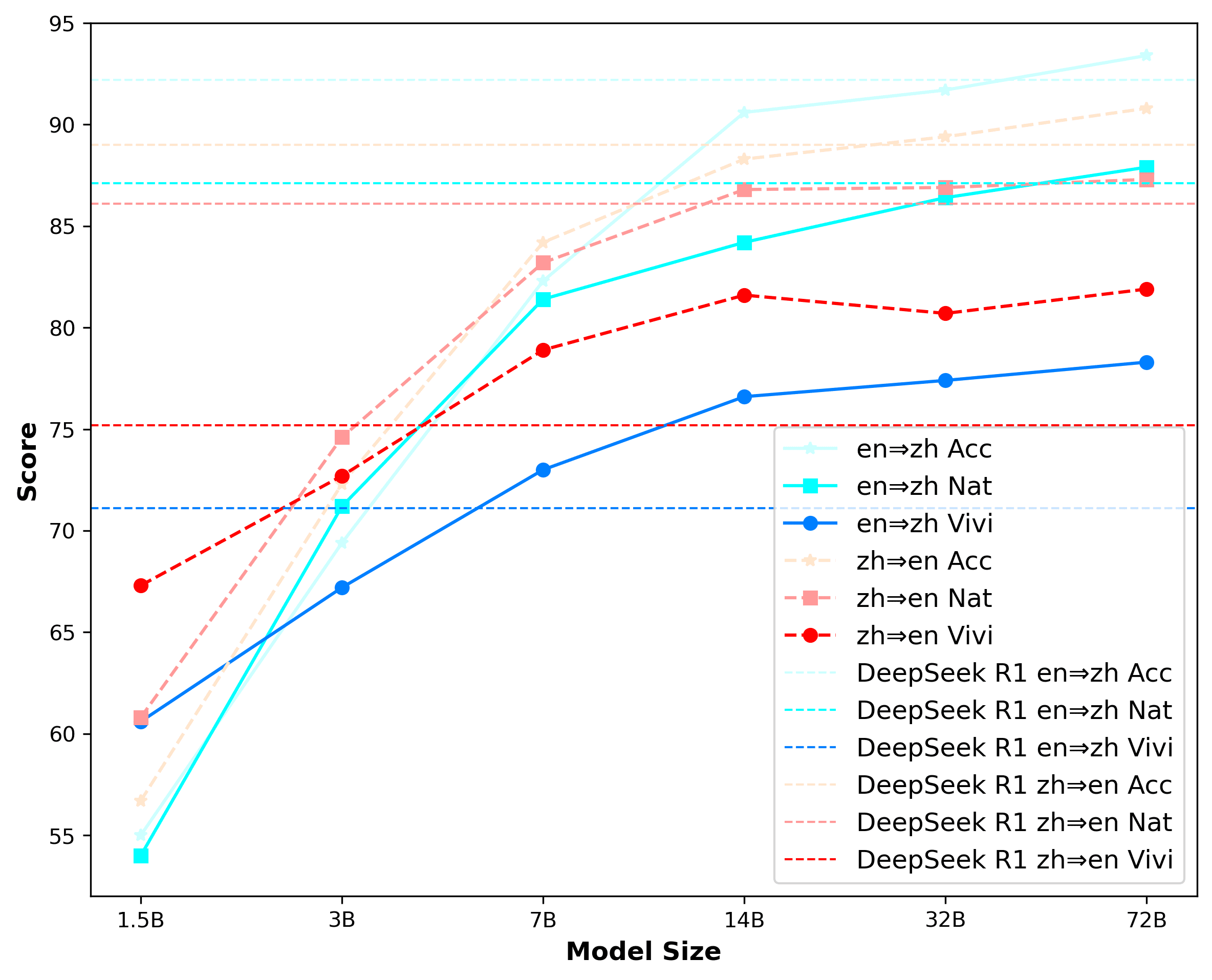}
  \caption{Impact of model size on translation quality.}
  \label{fig:modelsize}
\end{figure*}

In Figure~\ref{fig:modelsize}, we investigate the impact of ALPO's backbone model size on translation quality. Using the Qwen2.5 series models ranging from 1.5B to 72B as ALPO's base models, we evaluate translation performance in both \texttt{en}$\Rightarrow$\texttt{zh} and \texttt{zh}$\Rightarrow$\texttt{en} directions. Experimental results demonstrate that as model size increases, scores across multiple dimensions consistently improve, though the growth rate gradually diminishes. The 14B or 32B model generally achieves performance comparable to DeepSeek-R1, further validating ALPO's significance in enhancing translation quality. ALPO employs a fully autonomous approach to preference labeling of alignment data, enabling cost-effective training of high-quality subtitle translation models. Overall, the 14B base model strikes a balanced trade-off between cost and performance. However, as subtitle translation is an offline task, larger-scale models may be adopted when pursuing SOTA performance.

% 我们在Figure~\ref{fig:modelsize}中探究了ALPO的backbone模型size对翻译质量的影响。我们利用从1.5B到72B的Qwen2.5系列模型来作为ALPO的base model，并在\texttt{en}$\Rightarrow$\texttt{zh}和\texttt{zh}$\Rightarrow$\texttt{en}方向上进行翻译质量evaluation。实验结果表明随着model size的逐渐扩大，模型在多个维度上的得分均逐渐提高，但增长态势逐渐放缓。采用14B or 32B模型时一般可以达到DeepSeek-R1模型的水平，这进一步验证了ALPO对提升翻译质量的重要性。ALPO通过一种fully autonomous的方式实现了对alignment数据的偏好标注，从而低成本地训练出高质量的subtitle translation模型。综合来看，选用14B大小的base model是一种成本和性能更平衡的选择，不过subtitle translation作为一种离线任务，如果追求极致性能的话可以选用更大规模的模型。

\subsubsection{Data Volume}

% \begin{wrapfigure}{r}{0.6\textwidth}
%     \vspace{-0.7em}
%     \includegraphics[width=1.0\linewidth]{img/step.png}
%     \caption{Impact of training data volume on translation.}
%     \vspace{-1em}
%     \label{fig:step}
% \end{wrapfigure} 

\begin{figure*}[!h]
  \centering
  \includegraphics[width=0.6\textwidth]{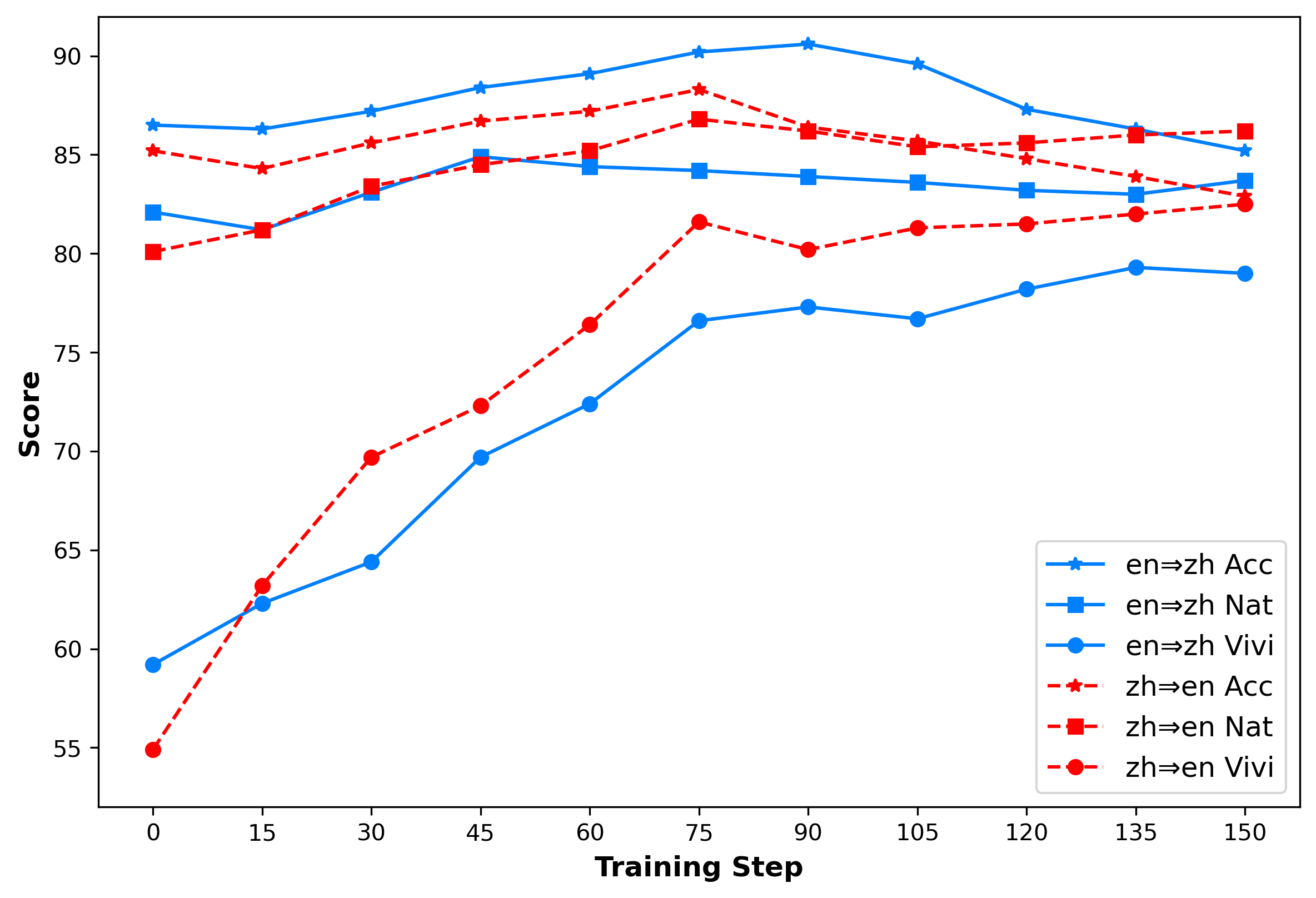}
  \caption{Impact of training data volume on translation quality.}
  \label{fig:step}
\end{figure*}

We present the translation quality scores across training steps in Figure~\ref{fig:step} to investigate the impact of training data scale. The results demonstrate that as training data volume increases, model accuracy initially rises before declining, naturalness shows modest improvement, while vividness exhibits substantial yet gradually decelerating growth. Although enhanced training data yields higher vividness scores, we observe pronounced hallucination phenomena in the model outputs, which explains the accuracy deterioration. Comprehensive analysis indicates that training for 75 steps (corresponding to 7,000 prompts with a batch size of 96) achieves optimal balance, which serves as the default configuration for all our experiments.

% 我们在Figure~\ref{fig:step}中展示了翻译质量得分随训练step的变化，从而探究训练数据的规模对翻译质量的影响。结果显示随着训练数据的使用量增大，模型的accuracy呈现出先增大后下降的趋势，naturalness会有微弱地增长，而vividness呈现明显的增长态势但增长趋势逐渐放缓。虽然使用更多的训练数据可以获取更高的vividness得分，但我们在如此训练的模型译文中观察到了明显的幻觉现象，这也是accuracy出现下降的原因。综合来看，训练75个step是最合适的配置，在batch size设置为96的情况下这对应于使用7000个prompt进行训练，这也是我们所有实验使用的默认设置。

\subsection{Further Exploration}
\label{sec:furtherexp}

\subsubsection{Vanilla DPO Training}

Training expressive subtitle translation LLM as a preference optimization problem, we want to explore \textbf{whether vanilla DPO can directly solve this issue.} Motivated by this inquiry, we designed and conducted relevant experiments. ALPO achieves significant improvements in translation quality through its segment-wise sampling strategy and fine-grained alignment loss, and we aim to verify the impact of these two components. Following the standard DPO training protocol \citep{dpo}, we perform post-training on the SFT model $\pi_{\text{sft}}$. Specifically, we employ either coarse-grained or fine-grained sampling to generate a chosen response $y^{(\text{c})}$ and a rejected response $y^{(\text{r})}$ for each sample $x \in \mathbb{D}_{\text{alpo}}$ in the alignment dataset, and subsequently optimize the policy model using the standard DPO loss.

% 训练expressive subtitle translation LLM作为一个preference optimization问题，我们想要探究一下“vanilla DPO算法能否直接解决这个问题？”。基于这个疑问我们进行了相关设计与实验。ALPO利用利用segment-wise sampling strategy和fine-grained alignment loss实现了翻译质量的显著提升，我们希望能够验证这两个component的影响。我们利用标准的DPO训练流程 \citep{dpo} 来对SFT模型$\pi _{\text{sft}}$进行post-training。具体的，我们利用coarse-grained或者fine-grained采样来为alignment数据集中的每个样本$x\in \mathbb{D}_{\text{alpo}}$采样一个chosen response $y^{(\text{c})}$和rejected response $y^{(\text{r})}$，然后利用标准的DPO loss来优化policy模型。

The coarse-grained and fine-grained sampling procedures are illustrated in Algorithm~\ref{alg:dpocoarse} and Algorithm~\ref{alg:dpofine}. In the coarse-grained sampling approach, we directly sample k complete responses for a given prompt x. Subsequently, a Qwen2-14B-Instruct model is utilized to score all subtitle lines across these k responses. The summation of scores for n subtitle lines within each response is calculated, with the highest-scoring response selected as the chosen response $y^{(\text{c})}$ and the lowest-scoring as the rejected response $y^{(\text{r})}$. For fine-grained sampling, we perform two separate segment-wise sampling iterations similar to Algorithm~\ref{alg:sampling} for each prompt x. During these iterations, the sampling of subsequent subtitle lines employs either the highest-scored or lowest-scored existing line as the prefix, ultimately yielding the segment-wise sampled chosen response $y^{(\text{c})}$ and rejected response $y^{(\text{r})}$.

% 采用的coarse-grained和fine-grained采样流程如Algorithm~\ref{alg:dpocoarse}和Algorithm~\ref{alg:dpofine}所示。其中coarse-grained sampling直接采样一个prompt$x$的$k$个完整response，然后利用一个Qwen2.5-14B-Instruct模型为$k$个response中的所有台词进行评分，并计算每个response中$n$句台词的评分的和，取最高的作为chosen response $y^{(\text{c})}$，最低的作为rejected response $y^{(\text{r})}$。fine-grained采样需要对一个prompt$x$进行两次类似于Algorithm~\ref{alg:sampling}中的segment-wise采样循环，在两次循环中分别以评分最高的和最低的台词作为下一句台词采样的prefix，最终获得segment-wise采样的chosen response $y^{(\text{c})}$和rejected response $y^{(\text{r})}$。

\begin{algorithm*}[!h]
  \caption{Coarse-grained Sampling for Vanilla DPO.}
  \label{alg:dpocoarse}
  \begin{algorithmic}[1]
    \Require SFT model $\pi _{\text{sft}}$, evaluation LLM $\pi _{\text{e}}$, alignment dataset $\mathbb{D}_{\text{alpo}}$, sample number $k$.
    \Ensure sampled response pairs set $\mathcal{S}(x)$.
    \For {any $x\in \mathbb{D}_{\text{alpo}}$} \quad $//$ Iterate through the alignment dataset $\mathbb{D}_{\text{alpo}}$.
    \For {$i=1$ to $k$} \quad $//$ Sample multiple candidate responses.
    \State Sample $\pi _{\text{sft}}(y\mid x)$.
    \EndFor
    \State Measure the sum of $\pi _{\text{e}}$ score $\mathcal{E}$ for each line of $y^{i}$ in the candidate set $\left \{y^{i}|i=1,2,\dots ,k\right \}$.
    \State Select chosen $y^{(\text{c})}$ and rejected $y^{(\text{r})}$.
    \EndFor \\
    \Return $\mathcal{S}(x)\equiv \left \{y^{(\text{c})},y^{(\text{r})}\right \}$.
  \end{algorithmic}
\end{algorithm*}

\begin{algorithm*}[!h]
  \caption{Fine-grained Sampling for Vanilla DPO.}
  \label{alg:dpofine}
  \begin{algorithmic}[1]
    \Require SFT model $\pi _{\text{sft}}$, evaluation LLM $\pi _{\text{e}}$, alignment dataset $\mathbb{D}_{\text{alpo}}$, sample number $k$.
    \Ensure sampled response pairs set $\mathcal{S}(x)$.
    \For {any $x\in \mathbb{D}_{\text{alpo}}$} \quad $//$ Iterate through the alignment dataset $\mathbb{D}_{\text{alpo}}$.
    \For {$i=1$ to $n$} \quad $//$ The first sampling cycle used to obtain $y^{(\text{c})}$.
    \For {$j=1$ to $k$}
    \State Sample $\pi _{\text{sft}}(t_{i}^{j}\mid x,t_{1}^{\text{(c)}},\dots ,t_{i-1}^{\text{(c)}})$.
    \EndFor
    \State Deduplicate the candidate set  $\{t_{i}^{j}\mid j=1,2,\dots ,k\}$.
    \State Add human reference $t_{i}^{0}$ to the candidate set, get $\{t_{i}^{j}\mid j=0,1,\dots \}$.
    \State Measure $\{t_{i}^{j}\mid j=0,1,\dots \}$ by $\pi _{\text{e}}$, get score sequence $\mathcal{E}_{i}$.
    \State Select a chosen translation $t_{i}^{\text{(c)}}$ (random from \textbf{top 3} of $\mathcal{E}_{i}$).
    \EndFor 
    \State Concatenate $\{t_{i}^{\text{(c)}}|i=1,2,\dots ,n\}$ yields $y^{(\text{c})}$.
    \For {$i=1$ to $n$} \quad $//$ The second sampling cycle used to obtain $y^{(\text{r})}$.
    \For {$j=1$ to $k$}
    \State Sample $\pi _{\text{sft}}(t_{i}^{j}\mid x,t_{1}^{\text{(r)}},\dots ,t_{i-1}^{\text{(r)}})$.
    \EndFor
    \State Deduplicate the candidate set  $\{t_{i}^{j}\mid j=1,2,\dots ,k\}$.
    \State Add human reference $t_{i}^{0}$ to the candidate set, get $\{t_{i}^{j}\mid j=0,1,\dots \}$.
    \State Measure $\{t_{i}^{j}\mid j=0,1,\dots \}$ by $\pi _{\text{e}}$, get score sequence $\mathcal{E}_{i}$.
    \State Select a rejected translation $t_{i}^{\text{(r)}}$ (random from \textbf{bottom 3} of $\mathcal{E}_{i}$).
    \EndFor 
    \State Concatenate $\{t_{i}^{\text{(r)}}|i=1,2,\dots ,n\}$ yields $y^{(\text{r})}$.
    \EndFor \\
    \Return $\mathcal{S}(x)\equiv \{y^{(\text{c})},y^{(\text{r})}\}$.
  \end{algorithmic}
\end{algorithm*}

\subsubsection{Advantage-based PPO Training}

In the literature, it is commonly acknowledged that in preference optimization techniques, RLHF (primarily based on PPO), despite its intricate procedures, outperforms DPO \citep{po1,po2}. This motivates our investigation into \textbf{whether PPO can achieve superior performance to ALPO.} To address this, we designed and conducted relevant experiments. We adopt an advantage-based PPO training pipeline \citep{po13}. Specifically, we implement the PPO training process through the following steps:  

\begin{enumerate}[label=\arabic*.]  
\item \textit{Rollout} — Use Algorithm~\ref{alg:ppo} to sample a trajectory $\tau$ for each input $x \in \mathbb{D}_{\text{alpo}}$.  
\item \textit{Computing Returns and Advantages} — Perform Generalized Advantage Estimation (GAE) on $\tau$ using the value network $V_{\phi}$ (base model: Qwen2.5-7B-Instruct \citep{qwen25}):  
\begin{equation}
\begin{split}
\delta _{i}&=-\mathcal{E}_{i}+\gamma V_{\phi }(p_{i+1})-V_{\phi }(p_{i}),\\
A_{i}&=\displaystyle\sum_{l=0}^{n-i-1}(\gamma \lambda )^{l}\delta _{i+l}.
\end{split}
\end{equation}
\item \textit{Updating Policy Network} — Let $\pi_{\text{old}}$ denote the fixed old policy during sampling. Minimize the clipped loss function (KL divergence constraint between $\pi_{\theta}$ and $\pi_{\text{old}}$ is omitted below):  
\begin{equation}
\mathcal{L}_{\text{clip}}(\theta )=-\mathbb{E}_{x\sim \mathbb{D}_{\text{alpo}}}\left [\displaystyle\sum_{i=1}^{n}\min \left (\frac{\pi _{\theta }(t_i\mid p_i)}{\pi _{\text{old}}(t_i\mid p_i)}A_i,\text{clip}\left (\frac{\pi _{\theta }(t_i\mid p_i)}{\pi _{\text{old}}(t_i\mid p_i)},1-\epsilon ,1+\epsilon \right )A_i\right )\right ].
\end{equation}
\item \textit{Updating Value Network} — Fit $V_{\phi}(p_{i})$ to the GAE-based estimate $\hat{V}_{i} = A_{i} + V_{\phi}(p_{i})$ via mean squared error:  
\begin{equation}
\mathcal{L}_{V}(\phi )=\mathbb{E}_{x\sim \mathbb{D}_{\text{alpo}}}\left [\displaystyle\sum_{i=1}^{n}(V_{\phi }(p_{i})-\hat{V}_{i})^{2}\right ].
\end{equation} 
\item \textit{Repeat for Multiple Epochs} — Continuously collect new data and update the value and policy networks until convergence.  
\end{enumerate}

% In the literature，通常认为在preference optimization技术中RLHF（主要是基于Proximal Policy Optimization (PPO)）虽然步骤繁琐但性能要优于DPO\citep{po1,po2}，因此我们想要探究一下“PPO技术能否比ALPO在DA中取得更优的效果？”。基于这个疑问我们进行了相关设计与实验。我们采用基于advantage的PPO训练流程\citep{16-2}。具体的，我们按照以下步骤执行PPO训练流程：
% \begin{enumerate}[label=\arabic*.]
% \item \textit{Rollout} - 利用Algorithm~\ref{alg:ppo}为每个样本$x\in \mathbb{D}_{\text{alpo}}$采样一个trajectory $\tau$。
% \item \textit{计算回报与优势} - 在$\tau$上利用价值网络$V_{\phi }$（基座为Qwen2.5-7B-Instruct \citep{qwen25}）进行Generalized Advantage Estimation (GAE)：
% \begin{equation}
% \begin{split}
% \delta _{i}&=-\mathcal{E}_{i}+\gamma V_{\phi }(p_{i+1})-V_{\phi }(p_{i}),\\
% A_{i}&=\displaystyle\sum_{l=0}^{n-i-1}(\gamma \lambda )^{l}\delta _{i+l}.
% \end{split}
% \end{equation}
% \item \textit{更新策略网络} - 令$\pi _{\text{old}}$为固定的采样时的旧策略，最小化损失函数（下式省略了$\pi _{\theta }$和$\pi _{\text{old}}$的KL散度约束）：
% \begin{equation}
% \mathcal{L}_{\text{clip}}(\theta )=-\mathbb{E}_{x\sim \mathbb{D}_{\text{alpo}}}\left [\displaystyle\sum_{i=1}^{n}\min \left (\frac{\pi _{\theta }(t_i\mid p_i)}{\pi _{\text{old}}(t_i\mid p_i)}A_i,\text{clip}\left (\frac{\pi _{\theta }(t_i\mid p_i)}{\pi _{\text{old}}(t_i\mid p_i)},1-\epsilon ,1+\epsilon \right )A_i\right )\right ].
% \end{equation}
% \item \textit{更新价值网络} - 利用均方误差来拟合$V_{\phi }(p_{i})$到GAE-based估计值$\hat{V}_{i}=A_{i}+V_{\phi }(p_{i})$：
% \begin{equation}
% \mathcal{L}_{V}(\phi )=\mathbb{E}_{x\sim \mathbb{D}_{\text{alpo}}}\left [\displaystyle\sum_{i=1}^{n}(V_{\phi }(p_{i})-\hat{V}_{i})^{2}\right ].
% \end{equation}
% \item \textit{重复多轮} - 不断收集新的数据、更新价值网络与策略网络，直到收敛。
% \end{enumerate}

\begin{algorithm*}[!h]
  \caption{Sampling for PPO solution.}
  \label{alg:ppo}
  \begin{algorithmic}[1]
    \Require SFT model $\pi _{\text{sft}}$, evaluation LLM $\pi _{\text{e}}$, alignment dataset $\mathbb{D}_{\text{alpo}}$.
    \Ensure sampled trajectory set $\mathcal{S}(x)\equiv \{\tau \}$.
    \For {any $x\in \mathbb{D}_{\text{alpo}}$} \quad $//$ Iterate through the DA dataset $\mathbb{D}_{\text{alpo}}$.
    \For {$i=1$ to $n$} \quad $//$ Iterate through the subtitle lines in $x$.
    \State Sample $\pi _{\text{sft}}(t_{i}\mid x,t_{1},\dots ,t_{i-1})$.
    \State Measure $t_{i}$ by $\pi _{\text{e}}$, get score $\mathcal{E}_{i}$.
    \EndFor
    \State Obtain trajectory $\tau =\{(p_i,t_i,\mathcal{E}_{i})\mid i=1,2,\cdots ,n\}$.
    \EndFor \\
    \Return $\mathcal{S}(x)\equiv \{\tau \}$.
  \end{algorithmic}
\end{algorithm*}

\subsubsection{Evaluation}

We evaluated the performance of \texttt{en}$\Rightarrow$\texttt{zh} and \texttt{zh}$\Rightarrow$\texttt{en} translation under "Vanilla DPO Training" and "Advantage-based PPO Training" configurations using the Qwen2.5-14B model. In addition, we also compared two token-level DPO-based methods, TDPO \citep{tdpo} and TIS-DPO \citep{tisdpo}. Experimental results presented in Table~\ref{tab:further-exp} demonstrate that while DPO and PPO training with segment-level sampling strategy achieved moderate performance improvements compared to the SFT model, they still exhibited significant gaps compared to ALPO with fine-grained alignment loss. ALPO also significantly outperformed the two token-level methods. This validates the effectiveness of both segment-wise sampling strategy and fine-grained alignment loss for the local preference optimization task.

% 我们利用Qwen2.5-14B模型验证了“Vanilla DPO Training”和“Advantage-based PPO Training”配置下\texttt{en}$\Rightarrow$\texttt{zh}和\texttt{zh}$\Rightarrow$\texttt{en}翻译的性能表现。另外，我们也对比了两个token-level DPO-based methods, TDPO\citep{tdpo} and TIS-DPO\citep{tisdpo}。实验结果展示在Table~\ref{tab:further-exp}中。结果表明采用segment-level sampling策略的DPO和PPO training相比SFT模型取得了一定的性能提升，但仍与采用fine-grained alignment loss的ALPO有较大差距。ALPO也显著超越了两个token-level的方法。这验证了segment-wise sampling strategy和fine-grained alignment loss对local preference optimization task的有效性。

\begin{table*}[!h]
\centering
\caption{Experimental evaluation results of alternative solutions. C for Coarse-grained, while F for Fine-grained. The 1st and 2nd best results are denoted as \colorbox{mlb}{\textbf{blue}} and \colorbox{mlo}{\textbf{orange}}.}
\resizebox{\textwidth}{!}{
\begin{tabular}{lccccc|ccc}
 \Xhline{1.0pt}
 \rowcolor{gray!20}
 ~ & ~ & ~ & \multicolumn{3}{c}{\textbf{\texttt{en}$\Rightarrow$\texttt{zh}}} & \multicolumn{3}{c}{\textbf{\texttt{zh}$\Rightarrow$\texttt{en}}}\\
 \cline{4-9}
 \rowcolor{gray!20}
 \multirow{-2}{*}{\textbf{Model}} & \multirow{-2}{*}{\textbf{Train}} & \multirow{-2}{*}{\textbf{Sampling}} & \textbf{Accuracy} & \textbf{Naturalness} & \textbf{Vividness} & \textbf{Accuracy} & \textbf{Naturalness} & \textbf{Vividness} \\
 \hline
 Gold Reference & - & - & \textit{83.6} & \textit{82.9} & \textit{71.6} & \textit{82.9} & \textit{80.2} & \textit{73.2} \\
 GPT-4o & - & - & \textit{89.5} & \textit{82.4} & \textit{59.5} & \textit{88.6} & \textit{82.9} & \textit{64.6} \\
 DeepSeek-R1 & - & - & \textit{90.4} & \textit{85.6} & \textit{70.7} & \textit{88.5} & \textit{85.6} & \textit{73.6} \\
 \hdashline
 \multirow{7}{*}{Qwen2.5-14B} & SFT & - & 86.5 & 82.1 & 59.2 & 85.2 & 80.1 & 54.9 \\
 ~ & \multirow{2}{*}{DPO} & C & 87.1 & 82.2 & 63.3 & 86.1 & 81.1 & 60.2 \\
 ~ & ~ & F & 88.2 & 82.5 & 69.4 & \colorbox{mlo}{\textbf{86.3}} & 83.1 & 68.4 \\
 ~ & PPO & F & 88.3 & \colorbox{mlo}{\textbf{83.2}} & \colorbox{mlo}{\textbf{70.1}} & 85.9 & 82.3 & \colorbox{mlo}{\textbf{70.1}} \\
 ~ & TDPO & F & 87.4 & 82.7 & 64.0 & 86.0 & 81.3 & 61.2 \\
 ~ & TIS-DPO & F & \colorbox{mlo}{\textbf{88.5}} & 83.1 & 65.9 & 86.2 & \colorbox{mlo}{\textbf{83.3}} & 62.4 \\
 ~ & \textbf{ALPO} & F & \colorbox{mlb}{\textbf{90.6}} & \colorbox{mlb}{\textbf{84.2}} & \colorbox{mlb}{\textbf{76.6}} & \colorbox{mlb}{\textbf{88.3}} & \colorbox{mlb}{\textbf{86.8}} & \colorbox{mlb}{\textbf{81.6}} \\
 \Xhline{1.0pt}
\end{tabular}
}
\label{tab:further-exp}
\end{table*}

\subsection{Further Application of ALPO in Multi-turn Interaction}
\label{sec:multiturn}

\subsubsection{Overview}

To validate the generality of the ALPO method, we conducted experiments on another local preference optimization task: social language agent multi-turn interaction \citep{sotopia,sotopiapi,selfalign,sdpo}. This task requires LLM agents to dynamically adjust generation strategies through multi-turn interactions in an interactive environment, guided by character profiles, contextual scenarios, and private social objectives. The objective is to enable the agent to more effectively accomplish predefined goals (e.g., persuasion, agreement attainment) within specific social scenarios (e.g., negotiation, collaboration, competition), while simultaneously maintaining or improving the relationship between both dialogue parties.

% 为了验证ALPO方法的通用性，我们在另一个local preference optimization task——social language agent multi-turn interaction \citep{sotopia,sotopiapi,selfalign,sdpo}上面进行了实验。该任务要求LLM agent在交互式环境中根据角色设定、场景背景和私密社交目标，通过多轮对话动态调整生成策略，使其能够在特定社交场景（如谈判、合作、竞争）中更有效地完成预设目标（例如说服对方、达成协议），同时维护或改善对话双方的关系。

\subsubsection{Method}

For the multi-turn interaction task, we primarily complete preference alignment data sampling through two steps: 1) Error Location; 2) Preference Data Sampling. For a multi-turn dialogue in the training set, we first employ GPT-4o to identify the starting position of erroneous turns, following the same criteria as Kong et al. \citep{sdpo}:
1) The turn is critical to achieving the role's goal;
2) There remains room for improvement in the relationship between goal completion and the role.

If both criteria are satisfied, we take the dialogue history preceding this turn as the initial sampling prefix $p_1$, then utilize our LLM agent to sequentially sample 5 single-turn responses for each turn. During the $i$-th sampling iteration, we employ GPT-4o to determine whether the current turn qualifies as crucial based on the sampled response: responses containing primarily pleasantries or similar content are deemed non-crucial, with their indicator function $\mathbf{1}(p_{i})$ set to 0; otherwise, it is set to 1. We then select the response with the highest combined goal and relationship score as the preferred response $y_i^w$, and the lowest as the dispreferred response $y_i^l$, with goal completion prioritized over relationship. The prefix $p_i$ for the $i$-th iteration becomes $p_{i-1}, y_{i-1}^w$ (including interlocutor behaviors, omitted here). Finally, we apply the following simplified ALPO loss function for alignment training:
\begin{equation}
\mathcal{L}_{\text{alpo}}(\pi _{\theta };\pi _{\text{ref}})=-\mathbb{E}_{(x,y_{1:n}^{w},y_{1:n}^{l})\sim \mathbb{D}}\left [\displaystyle\sum_{i=1}^{n}\mathbf{1}(p_{i})\cdot \text{log}\, \sigma \left (\beta \, \text{log}\frac{\pi _{\theta }(y_i^{w}\mid p_i)}{\pi _{\text{ref}}(y_i^{w}\mid p_i)}-\beta \, \text{log}\frac{\pi _{\theta }(y_i^{l}\mid p_i)}{\pi _{\text{ref}}(y_i^{l}\mid p_i)}\right )\right ],
\end{equation}
where $\mathbb{D}$ denotes the dataset, and both $\pi _{\theta }$ and $\pi _{\text{ref}}$ are initialized from behavioral cloning (BC) models.

% 对于multi-turn interaction任务，我们主要分两步来完成偏好对齐数据采样：1) Error Location；2) Preference Data Samping。对于训练集中的一个multi-turn dialogue，我们首先利用GPT-4o来识别erroneous turn的起始位置，根据与Kong et al. \citep{sdpo}相同的标准：
% \begin{itemize}
% \item 该turn对于达成role的goal是关键的；
% \item 对于goal completion和role之间的relationship仍然有提升的空间。
% \end{itemize}
% 如果符合这两个标准，我们则以该turn前的dialogue历史作为最初的采样prefix$p_1$，然后利用我们的LLM agent来依次为每个turn采样5个单turn response。对于第$i$次采样，我们利用GPT-4o来根据采样的response来判断当前turn是否为crucial turn，如果主要为pleasantries等内容则为不crucial，其indicator function $\mathbf{1}(p_{i})$为0，否则为1。然后从采样的5个response中选择goal和relationship score最高的作为preferred response $y_i^w$，最低的作为dispreferred response $y_i^l$，且goal completion prioritized over relationship。第$i$次采样的prefix$p_i$为$p_{i-1},y_{i-1}^w$（也包含interlocutor的行为，已省略）。最终我们应用以下简化的ALPO损失损失函数来进行对齐训练：
% \begin{equation}
% \mathcal{L}_{\text{alpo}}(\pi _{\theta };\pi _{\text{ref}})=-\mathbb{E}_{(x,y_{1:n}^{w},y_{1:n}^{l})\sim \mathbb{D}}\left [\displaystyle\sum_{i=1}^{n}\mathbf{1}(p_{i})\cdot \text{log}\, \sigma \left (\beta \, \text{log}\frac{\pi _{\theta }(y_i^{w}\mid p_i)}{\pi _{\text{ref}}(y_i^{w}\mid p_i)}-\beta \, \text{log}\frac{\pi _{\theta }(y_i^{l}\mid p_i)}{\pi _{\text{ref}}(y_i^{l}\mid p_i)}\right )\right ],
% \end{equation}
% where $\mathbb{D}$为训练数据集，$\pi _{\theta }$和$\pi _{\text{ref}}$均初始化自behavioral cloning (BC)模型。

\subsubsection{Experiments}

We utilize 100 out of the 410 scenarios from SOTOPIA-$\pi$ \citep{sotopiapi} for behavioral cloning (10 role pairs per scenario) and 310 scenarios for alignment (8 role pairs per scenario). We adopt SOTOPIA \citep{sotopia} as the test set, containing 90 scenarios with 5 role pairs per scenario, totaling 450 self-chat tasks and 900 non-self-chat tasks. We evaluate both goal and relationship dimensions on the same baselines, with experimental results presented in Table~\ref{tab:multiturn}. Baseline results are sourced from Kong et al. \citep{sdpo}.

% 我们将SOTOPIA-$\pi$ \citep{sotopiapi}的410个scenario中的100个用于behavioral cloning（10 role pairs per scenario），310个用于对齐（8 role pairs per scenario）。我们以SOTOPIA\citep{sotopia}作为测试集，包含90个scenario，每个scenario 5个role pair，总计450个self-chat任务和900个non-self-chat任务。我们同样在相同的baseline上评估goal和relationship dimensions，实验结果展示在Table~\ref{tab:multiturn}中，baseline结果来自Kong et al. \citep{sdpo}。

\begin{table*}[!h]
\centering
\caption{The performance of various methods on SOTOPIA across the goal and relationship dimensions. The 1st and 2nd best results are denoted as \colorbox{mlb}{\textbf{blue}} and \colorbox{mlo}{\textbf{orange}}, respectively.}
% \resizebox{\textwidth}{!}{
\begin{tabular}{lcccc}
 \Xhline{1.0pt}
 \rowcolor{gray!20}
 ~ & \multicolumn{2}{c}{\textbf{Self-Chat}} & \multicolumn{2}{c}{\textbf{GPT-4o}} \\
 \cline{2-5}
 \rowcolor{gray!20}
 \multirow{-2}{*}{\textbf{Methods}} & \textbf{Goal} & \textbf{Rel} & \textbf{Goal} & \textbf{Rel} \\
 \hline
 GPT-3.5-turbo & 6.38 & 1.36 & 7.19 & 2.05 \\
 GPT-4o-mini & 6.98 & 2.11 & 7.44 & 2.36 \\
 GPT-4-turbo & 8.18 & 2.96 & 7.92 & 2.79 \\
 GPT-4o & 7.90 & 2.67 & 7.90 & 2.67 \\
 \hdashline
 LLaMA-8B & 7.24 & 1.94 & 7.70 & 2.49 \\
 LLaMA-8B+BC & 7.81 & 3.05 & 7.53 & 2.78 \\
 LLaMA-8B+BC+Preferred-SFT & 7.76 & 3.05 & 7.65 & 2.88 \\
 LLaMA-8B+BC+DPO \citep{dpo} & 7.95 & 3.28 & 7.80 & 2.97 \\
 LLaMA-8B+BC+ETO \citep{eto} & 8.29 & 3.39 & 8.02 & 3.03 \\
 LLaMA-8B+BC+DMPO \citep{dmpo} & 8.28 & 3.37 & 8.00 & 2.98 \\
 LLaMA-8B+BC+SDPO \citep{sdpo} & \colorbox{mlb}{\textbf{8.56}} & \colorbox{mlb}{\textbf{3.69}} & \colorbox{mlo}{\textbf{8.13}} & \colorbox{mlo}{\textbf{3.16}} \\
 \hdashline
 LLaMA-8B+BC+ALPO & \colorbox{mlo}{\textbf{8.41}} & \colorbox{mlo}{\textbf{3.56}} & \colorbox{mlb}{\textbf{8.21}} & \colorbox{mlb}{\textbf{3.19}} \\
 \Xhline{1.0pt}
\end{tabular}
% }
\label{tab:multiturn}
\end{table*}

The experimental results demonstrate the effectiveness of ALPO on other application tasks. ALPO employs a progressive optimization strategy transitioning from local to global optima, while utilizing an indicator function to adaptively determine the participation of each segment in the loss computation. This approach enables effective handling of multi-segment local optimization tasks.

% 实验结果验证了ALPO在其他应用任务上的有效性。ALPO采用一种从局部最优过度到全局最优的优化方式，并且利用indicator function来自适应地决定每个segment是否参与损失计算，从而实现了对multi-segment local优化任务的有效处理。

\section{Theory: Adaptive Local Preference Optimization}
\label{sec:theory}

In this section, we formalize the local preference optimization problem and validate the effectiveness of ALPO and the limitations of general preference optimization approaches. Note that for the sake of expository convenience, the notations used in this section may differ from those in previous sections.

% In this section，我们将local preference optimization problem形式化，并验证ALPO的有效性和通用preference optimization方法的局限性。注意，为了表述的便利性，本section使用的符号与前文含义可能略有不同。

\subsection{Local Preference Optimization Problem}

For a given input $x \in \mathcal{X}$ ($\mathcal{X}$ denotes the input space) to the language model $\pi_{\theta}$, assume it consists of $n$ interrelated segments, i.e., $x = (x_1, \cdots, x_n)$. Correspondingly, its output $y \in \mathcal{Y}$ ($\mathcal{Y}$ being the output space) is also composed of $n$ interrelated segments, i.e., $y = (y_1, \cdots, y_n)$, where $x_i$ and $y_i$ exhibit a one-to-one correspondence. Additionally, the generation of $y_i$ is influenced by $y_1, \cdots, y_{i-1}$ (note that this influence arises not only from the autoregressive nature of language model but also potentially from semantic dependencies between output segments, among other factors), i.e.:  
\begin{equation}
\pi _{\theta }(y\mid x)=\displaystyle\prod_{i=1}^{n}\pi _{\theta }(y_{i}\mid x,y_1,\cdots ,y_{i-1}).
\end{equation}

% 对于语言模型$\pi _{\theta }$的一个输入$x\in \mathcal{X}$（$\mathcal{X}$为输入空间），假设其由相互关联的$n$个segment组成，即$x=(x_1,x_2,\cdots ,x_n)$，相应的其输出$y\in \mathcal{Y}$（$\mathcal{Y}$为输出空间）也由相互关联的$n$个segment组成，即$y=(y_1,y_2,\cdots ,y_n)$，并且$x_i$和$y_i$具有一一对应性。另外，$y_i$的生成受到$y_1,y_2,\cdots ,y_{i-1}$的影响（注意这种影响并非仅来自于语言模型的自回归属性，也可能来自输出segment之间的语义依赖等），即：
% \begin{equation}
% \pi _{\theta }(y\mid x)=\displaystyle\prod_{i=1}^{n}\pi _{\theta }(y_{i}\mid x,y_1,\cdots ,y_{i-1}).
% \end{equation}

Consider a segment-level preference metric $r(x_i, y_i)$ (a reward signal or other quantitative measure) that evaluates how well the segment output $y_i$ aligns with the optimization objective for its corresponding segment input $x_i$. The goal of the local preference optimization problem is to adjust the parameters $\theta$ such that $\pi_{\theta}$ is optimized to maximize $r(x_i, y_i)$. This objective is formally expressed as:
\begin{equation}
\theta^{*}=\underset{\theta}{\arg\max}\;\mathbb{E}_{x \sim p(x)}\Bigl[\mathbb{E}_{y\sim \pi_{\theta}(\cdot \mid x)}\bigl[\sum_{i=1}^{n} r(x_i,\,y_i)\bigr]\Bigr],
\end{equation}
where $p(x)$ represents the true distribution of the input $x$.

% 现存在一个segment-level的preference度量$r(x_i,y_i)$（reward信号或者其他量化标准）能够衡量segment输入$x_i$对应的segment输出$y_i$相对于优化方向preference的契合程度。local preference optimization problem的目标是调整参数$\theta$使得$\pi _{\theta }$向最大化$r(x_i,y_i)$的方向优化，形式化为：
% \begin{equation}
% \theta^{*}=\underset{\theta}{\arg\max}\;\mathbb{E}_{x \sim p(x)}\Bigl[\mathbb{E}_{y\sim \pi_{\theta}(\cdot \mid x)}\bigl[\sum_{i=1}^{n} r(x_i,\,y_i)\bigr]\Bigr],
% \end{equation}
% 其中$p(x)$为输入$x$的真实分布。

\subsection{Ineffectiveness of Outcome-supervised Preference Optimization Methods}

We take DPO \citep{dpo} as an example to illustrate the limitations of general preference optimization methods when addressing local preference optimization problems. When performing local preference optimization, DPO first annotates preferred responses $y^w=(y_1^w,\cdots ,y_n^w)$ and non-preferred responses $y^l=(y_1^l,\cdots ,y_n^l)$ corresponding to $x=(x_1,\cdots ,x_n)$ based on $r(x_i, y_i)$. These responses inherently satisfy:
\begin{equation}
\pi _{\theta }(y^w\mid x)=\displaystyle\prod_{i=1}^{n}\pi _{\theta }(y_i^w\mid x,y_1^w,\cdots ,y_{i-1}^w), \pi _{\theta }(y^l\mid x)=\displaystyle\prod_{i=1}^{n}\pi _{\theta }(y_i^l\mid x,y_1^l,\cdots ,y_{i-1}^l).
\end{equation}

% 我们以DPO\citep{dpo}为例，阐述general preference optimization methods在面对local preference optimization problem时的局限性。DPO在进行local偏好优化时首先根据$r(x_i, y_i)$分别标注对应于$x=(x_1,x_2,\cdots ,x_n)$的完整的偏好的response$y^w=(y_1^w,y_2^w,\cdots ,y_n^w)$和不偏好的response$y^l=(y_1^l,y_2^l,\cdots ,y_n^l)$，$y^w$和$y^l$必然地满足：
% \begin{equation}
% \pi _{\theta }(y^w\mid x)=\displaystyle\prod_{i=1}^{n}\pi _{\theta }(y_i^w\mid x,y_1^w,\cdots ,y_{i-1}^w), \pi _{\theta }(y^l\mid x)=\displaystyle\prod_{i=1}^{n}\pi _{\theta }(y_i^l\mid x,y_1^l,\cdots ,y_{i-1}^l).
% \end{equation}

Then, DPO optimizes $\pi_{\theta}$ through a single-stage policy optimization approach:
\begin{equation}
\mathcal{L}_{\text{DPO}}(\pi _{\theta };\pi _{\text{ref}})=-\mathbb{E}_{(x,y^{w},y^{l})\sim \mathbb{D}}\left [\text{log}\, \sigma \left (\beta \, \text{log}\frac{\pi _{\theta }(y^{w}\mid x)}{\pi _{\text{ref}}(y^{w}\mid x)}-\beta \, \text{log}\frac{\pi _{\theta }(y^{l}\mid x)}{\pi _{\text{ref}}(y^{l}\mid x)}\right )\right ],
\end{equation}
where $\pi_{\text{ref}}$ denotes a frozen reference model. For the contrastive terms in the loss function, compute:
\begin{equation}
\text{log}\frac{\pi _{\theta }(y^{w}\mid x)}{\pi _{\text{ref}}(y^{w}\mid x)}-\text{log}\frac{\pi _{\theta }(y^{l}\mid x)}{\pi _{\text{ref}}(y^{l}\mid x)}=\displaystyle\sum_{i=1}^{n}\left [\text{log}\frac{\pi _{\theta }(y_i^w\mid x,y_{1:i-1}^w)}{\pi _{\text{ref}}(y_i^w\mid x,y_{1:i-1}^w)}-\text{log}\frac{\pi _{\theta }(y_i^l\mid x,y_{1:i-1}^l)}{\pi _{\text{ref}}(y_i^l\mid x,y_{1:i-1}^l)}\right ],
\end{equation}
where $y_{1:i-1}^w=(y_1^w,\cdots,y_{i-1}^w)$ and $y_{1:i-1}^l=(y_1^l,\cdots,y_{i-1}^l)$. The limitations of this approach are:
\begin{itemize}
\item \textbf{Lack of Fine-Grained Contrast on Aligned Prefixes}: The generation probability of the $i$-th segment is conditioned on its own preferred or dispreferred prefix ($y_{1:i-1}^w$ or $y_{1:i-1}^l$), rather than a shared prefix. If $y_i^w$ and $y_i^l$ exhibit significant divergence in their prefixes (e.g., differing styles or topic branches), the contrast at the $i$-th segment ceases to be a fair comparison under identical conditions.
\item \textbf{Inability to Apply Training Signals Per Segment}: DPO performs a holistic preference judgment over complete sequences $(y^w, y^l)$, depriving the model of segment-level corrective guidance. Specifically, while the model recognizes the global superiority of $y^w$ over $y^l$, it lacks localized feedback to optimize decisions at individual segments.
\item \textbf{Gradient Dilution and Noise}: When the divergence between $y^w$ and $y^l$ occurs only at specific segments, the global contrast aggregates gradients from numerous irrelevant segments (nearly identical in generation). This drowns critical alignment signals, hindering effective identification and correction of misaligned segments.
\end{itemize}

% 然后，DPO以单阶段策略优化的方式优化$\pi _{\theta }$：
% \begin{equation}
% \mathcal{L}_{\text{DPO}}(\pi _{\theta };\pi _{\text{ref}})=-\mathbb{E}_{(x,y^{w},y^{l})\sim \mathbb{D}}\left [\text{log}\, \sigma \left (\beta \, \text{log}\frac{\pi _{\theta }(y^{w}\mid x)}{\pi _{\text{ref}}(y^{w}\mid x)}-\beta \, \text{log}\frac{\pi _{\theta }(y^{l}\mid x)}{\pi _{\text{ref}}(y^{l}\mid x)}\right )\right ],
% \end{equation}
% 这里$\pi _{\text{ref}}$是一个参数固定的reference模型。对于损失函数中的对比项，计算：
% \begin{equation}
% \text{log}\frac{\pi _{\theta }(y^{w}\mid x)}{\pi _{\text{ref}}(y^{w}\mid x)}-\text{log}\frac{\pi _{\theta }(y^{l}\mid x)}{\pi _{\text{ref}}(y^{l}\mid x)}=\displaystyle\sum_{i=1}^{n}\left [\text{log}\frac{\pi _{\theta }(y_i^w\mid x,y_{1:i-1}^w)}{\pi _{\text{ref}}(y_i^w\mid x,y_{1:i-1}^w)}-\text{log}\frac{\pi _{\theta }(y_i^l\mid x,y_{1:i-1}^l)}{\pi _{\text{ref}}(y_i^l\mid x,y_{1:i-1}^l)}\right ],
% \end{equation}
% 其中$y_{1:i-1}^w=(y_1^w,\cdots ,y_{i-1}^w),\, y_{1:i-1}^l=(y_1^l,\cdots ,y_{i-1}^l)$。这样做的问题是：
% \begin{itemize}
% \item \textbf{缺失对应prefix的精细对比}：第$i$个segment的生成概率是条件在自身自身偏好/不偏好的prefix上（即$y_{1:i-1}^w$或$y_{1:i-1}^l$），并不是条件在同一个prefix上。若$y_i^w$与$y_i^l$在prefix部分已经显著分歧（如不同风格、不同话题分支），那么第$i$个segment的对比就不会是“同一起跑线”上的公平对比。
% \item \textbf{无法逐段施加训练信号}：DPO只对完整序列$(y^w, y^l)$做一次整体的偏好判断，模型无法获知每一个segment应当如何修正。换言之，模型只知道整条序列$y^w$优于整条序列$y^l$，但具体到第$i$个segment时，模型并不清楚如何在局部做更优的选择。
% \item \textbf{梯度稀释与噪声}：当$y^w$与$y^l$的差异仅在个别的segment时，整体对比会将绝大多数无关segment (两者生成几乎相同) 混在一起，导致梯度信号被淹没，难以有效学到哪个segment不对齐、哪个segment应如何改进。
% \end{itemize}

In summary, DPO underperforms in local preference optimization tasks due to misaligned generative conditions caused by prefix divergence and the absence of stepwise "correct vs. incorrect" decision learning (a limitation shared by other outcome-supervised preference optimization losses).

% 综上，DPO在面对local preference optimization problem时，由于prefix分支的差异没有被对齐到同一个生成条件，且模型也无法逐段学习每一步的“正确 vs. 不正确”决策，因此DPO效果不理想（其他outcome-supervised preference optimization损失类似）。

\subsection{Adaptive Local Preference Optimization}

Unlike vanilla DPO, ALPO labels the preferred response $y_i^w$ and dispreferred response $y_i^l$ for each segment using $y_{1:i-1}^w$ as the prefix, instead of generating the entire response sequence. During training, the prefix for the $i$-th segment is fixed as $p_i=(x,\hat{y}_{1},\cdots ,\hat{y}_{i-1})$, where $\hat{y}_i$ is obtained through the scheduled prefix mixing strategy. This ensures that comparisons for each segment are made under the same preferred prefix, eliminating unfair competition between “preferred prefix vs. dispreferred prefix”. Under the same prefix, the preference alignment loss contrasts $\pi _{\theta }(y_i^w\mid p_i)$ and $\pi _{\theta }(y_i^l\mid p_i)$ conditioned on $p_i$. Finally, the summation of segment-wise losses imposes preference constraints at each segment:
\begin{equation}
\mathcal{L}_{\text{alpo}}(\pi _{\theta };\pi _{\text{ref}})=-\mathbb{E}_{(x,y_{1:n}^{w},y_{1:n}^{l})\sim \mathbb{D}}\Bigg[\displaystyle\sum_{i=1}^{n}w(p_{i})\cdot \text{log}\, \sigma \bigg(\beta _{i}\, \text{log}\frac{\pi _{\theta }(y_i^{w}\mid p_i)}{\pi _{\text{ref}}(y_i^{w}\mid p_i)}-\beta _{i}\, \text{log}\frac{\pi _{\theta }(y_i^{l}\mid p_i)}{\pi _{\text{ref}}(y_i^{l}\mid p_i)}\bigg)\Bigg],
\end{equation}
where $w(p_{i})$ and $\beta _{i}$ are adaptive hyperparameters that determine the contribution of each segment to the optimization process based on specific criteria.

% 与vanilla DPO不同，ALPO逐segment地以$y_{1:i-1}^w$为prefix标注下一个segment的preferred response$y_i^w$和dispreferred response$y_i^l$，而非完整地获取整个response序列。在训练时，对于第$i$个segment，其prefix为固定的prefix$p_i=(x,\hat{y}_{1},\cdots ,\hat{y}_{i-1})$，$\hat{y}_i$为通过scheduled prefix mixing strategy获取的prefix。这样每一个segment的对比都在相同且preferred的prefix上进行，排除了‘偏好prefix vs. 不偏好prefix’的不公平竞争。在相同的prefix下，应用偏好对齐损失将同一条件$p_i$下的$\pi _{\theta }(y_i^w\mid p_i)$和$\pi _{\theta }(y_i^l\mid p_i)$进行对比。最终，累加所有segment的损失即可在每一个segment上施加偏好约束：
% \begin{equation}
% \mathcal{L}_{\text{alpo}}(\pi _{\theta };\pi _{\text{ref}})=-\mathbb{E}_{(x,y_{1:n}^{w},y_{1:n}^{l})\sim \mathbb{D}}\Bigg[\displaystyle\sum_{i=1}^{n}w(p_{i})\cdot \text{log}\, \sigma \bigg(\beta _{i}\, \text{log}\frac{\pi _{\theta }(y_i^{w}\mid p_i)}{\pi _{\text{ref}}(y_i^{w}\mid p_i)}-\beta _{i}\, \text{log}\frac{\pi _{\theta }(y_i^{l}\mid p_i)}{\pi _{\text{ref}}(y_i^{l}\mid p_i)}\bigg)\Bigg],
% \end{equation}
% 这里$w(p_{i})$和$\beta _{i}$为自适应hyperparameter，用于根据特定的criterion来决定当前segment对优化过程的影响。

The reason for using scheduled prefix mixing to obtain $\hat{y}_{1:i-1}$ as the prefix is that, as optimization proceeds, the model tends to generate $y_i^w$, making the post-training distribution more consistent with the $y_{1:i-1}^w$ path. Since $y_{1:i-1}^w$ is also the prefix path most likely to be followed, this ensures that the model learns to generate the preferred segment $y_i^w$ rather than $y_i^l$ under the prefixes it visits most frequently. This aligns with a common principle in reinforcement learning and preference optimization: \textit{the state (or prefix) distribution during training should match the distribution most likely to be visited by the final policy (on-policy)} \citep{po8,onpolicy,po1}. Formally, during training we aim to minimize:
\begin{equation}
\mathbb{E}_{p \sim d_{\theta^*}}\Bigl[\mathcal{L}_{\text{alpo}}\bigl(\pi_\theta(\cdot \mid p)\bigr)\Bigr],
\end{equation}
where $d_{\theta^*}$ denotes the prefix distribution induced by the final model. Since $d_{\theta^*}$ is likely to include preferred prefixes upon convergence, selecting $\hat{y}_{1:i-1}$ as the prefix essentially samples (or directly uses) states from the preferred path, ensuring consistency with the final distribution $d_{\theta^*}$.

% 之所以采用scheduled prefix mixing获取$\hat{y}_{1:i-1}$作为prefix，其原因在于随着优化进行，模型会倾向于生成$y_i^w$，从而在训练后模型分布就与$y_{1:i-1}^w$路径保持较高一致性。$y_{1:i-1}^w$也是模型最倾向走到的prefix路径。这样做可以确保模型在自己最常访问的prefix上学习到如何在相同前缀下继续生成preferred segment$y_i^w$而非$y_i^l$。这也符合强化学习和偏好优化中常见的一个原则：\textit{训练时的状态(或prefix)分布，应该与最终策略最可能访问的分布相匹配(on-policy)} \citep{po8,onpolicy,po1}。形式化地，在训练过程中，我们希望最小化：
% \begin{equation}
% \mathbb{E}_{p \sim d_{\theta^*}}\Bigl[\mathcal{L}_{\text{alpo}}\bigl(\pi_\theta(\cdot \mid p)\bigr)\Bigr],
% \end{equation}
% 其中$d_{\theta^*}$表示最终模型下可能出现的前缀分布。因为当模型收敛时，$d_{\theta^*}$将较大概率地包含那些preferred prefix。于是选择用$\hat{y}_{1:i-1}$来作为prefix，本质上就是采样（或直接使用）该preferred路径所带来的状态，从而与最终分布$d_{\theta^*}$保持一致。

\section{Discussion}
\label{sec:discussion}

In this section, we provide further discussion on the ALPO method.

% In this section，我们将进行更多关于ALPO方法的讨论。

\subsection{Conclusion}

The visual media industry, serving as a pivotal medium for human cultural exchange and dissemination, is witnessing automation and industrialization as critical future trends. Against this backdrop, we propose ALPO, a novel training paradigm for translation models, by employing techniques including LLM-as-a-Judge and preference alignment. Through comprehensive experimentation and theoretical analysis, we validate the effectiveness of ALPO. We believe this study will not only positively contribute to the advancement of visual media technologies but also promote research on domain-specific translation models in other fields.

% visual media行业作为人类文明交流和传播的重要媒介，其自动化与工业化是未来发展的重要趋势。在这一发展背景下，in this study，我们应用了LLM-as-a-Judge、偏好对齐等技术提出了ALPO这一novel的翻译模型训练范式。我们进行了充分的实验与理论分析验证了ALPO的有效性。我们相信this study将对于未来的visual media的发展起到积极的促进作用，同时也能促进其他领域专有翻译模型的研究。

\subsection{Recommendations}

We would like to offer some developer suggestions for technicians using ALPO:
\begin{itemize}
\item When employing LLMs for preference annotation, shuffle the order of translations across multiple inference passes and average the results to obtain robust evaluation outcomes.
\item The training prompt should utilize the language best suited to the selected backbone, e.g., Chinese prompts for Qwen-series models and English prompts for LLaMA-series models.
% \item We primarily recommend using DPO as the loss function for ALPO, which demonstrates the most stable configuration based on our experimental results and empirical observations.
\item Our open-source implementation reveals additional processing details, including handling of terminology translation, subtitle segmentation, and semantic integrity preservation. Refer to the source code for specific implementations.
\end{itemize}

% 我们希望为使用ALPO的技术人员提供一些开发者建议：
% \begin{itemize}
% \item 利用LLM进行偏好标注时可以打乱译文顺序多次推理并取平均值以获取鲁棒评估结果。
% \item 训练模型的prompt可以使用最适合选取的backbone模型的语言，例如Qwen系列模型使用中文prompt，LLaMA系列模型使用英文prompt。
% \item 我们首要推荐使用DPO作为ALPO的损失函数，其在我们的实验和经验上都是效果最稳定的配置。
% \item 我们开源的代码中展示了更多处理细节，例如对术语翻译的处理、对台56TTTTT9O0\subsection{Future Research}

Subtitle texts are inherently multimodal content closely intertwined with the visual and auditory elements of media programs. Videos provide supplementary information about characters, scenes, and plot progression, while audio conveys emotional and tonal cues. Future research should therefore focus on harnessing these multimodal data resources to enhance subtitle translation, e.g., leveraging video and audio modalities to enable speaker annotation for subtitle lines, and utilizing audiovisual context to facilitate subtitle translation.

% 字幕文本实际上是与visual media programs的视频和音频内容紧密联系的内容，视频能够补充角色、场景、情节等信息，音频能够补充情感、语气等信息。因此未来的研究应该集中于如何利用这些多模态的数据资源以训练更好的subtitle translation，例如基于视频和音频模态实现对字幕台词的说话人标注，以及利用视频与音频信息辅助完成字幕翻译等。

\subsection{Limitations}
\label{sec:limitations}

ALPO has the following limitations:  
\begin{itemize}  
    \item The visual media data and associated copyrights required for subtitle translation are predominantly concentrated within large enterprises, creating data accessibility barriers.  
    \item Similar to other SOTA LLMs, ALPO lacks the capability to integrate multimodal information (e.g., video) during the translation process.  
\end{itemize}

% ALPO存在以下limitations：
% \begin{itemize}
% \item Subtitle translation依赖的visual media数据及其版权通常集中于大型企业手中，存在数据壁垒。
% \item ALPO与其他SOTA LLM一样，不具备结合视频等其他模态信息进行翻译的能力。
% \end{itemize}

% \subsection{LLM Usage Statement}

% LLMs were employed exclusively for polishing the presentation of the manuscript (e.g., grammar, style, and readability). All scientific content was produced entirely by the authors.

\section{Prompts and Instructions}
\label{sec:pi}

In this section, we present the input and output formats used for LLM and human evaluators.

% In this section，我们展示了我们采用的LLM的输入与输出格式以及用于人工评测翻译质量的evaluation instruction等文本内容。

\subsection{Input and Output of Translation LLM}
\label{sec:iosft}

We illustrate the prompt and response format of the \texttt{zh}$\Rightarrow$\texttt{en} translation model below (similar formats apply to other languages). The red text appears only in LLM ICL. The original and translated subtitles of the program are formatted as shown in the text boxes for use in SFT and ALPO model training, as well as other baseline LLM ICL.

\begin{tcolorbox}[
    title={The \texttt{zh}$\Rightarrow$\texttt{en} translation prompt demonstration of translation model.},
    colback=blue!5!white,       % 背景：极淡的蓝色
    colframe=blue!75!black,     % 边框：深蓝色
    coltitle=white,             % 标题文字：白色
    breakable,
    fontupper=\small
]
\texttt{<INSTRUCTION> \newline Please translate the following multiple Chinese movie/TV lines into English according to the following requirements: \newline 1. The English translation should be colloquial, easily understandable, and that the language style is consistent with the Chinese lines. \newline 2. The English translation must be expressive and vivid, effectively conveying the atmosphere, emotions, and tone of the original Chinese lines. \newline 3. Output the translation with its line number, ensuring that the number of lines in the translation matches the number of lines in the Chinese original, without merging any lines. \newline }\\
\texttt{\textcolor{red}{<EXAMPLE>} \newline \textcolor{red}{Original text:}\newline \textcolor{red}{1.\begin{CJK}{UTF8}{gkai}请给你我一点和平相处的时间\end{CJK}} \newline \textcolor{red}{2.\begin{CJK}{UTF8}{gkai}我们必须谈谈\end{CJK}} \newline \textcolor{red}{...}  \newline \textcolor{red}{34.\begin{CJK}{UTF8}{gkai}奥术在你体内涌动\end{CJK}} \newline \textcolor{red}{35.\begin{CJK}{UTF8}{gkai}请你离海克斯飞门远点\end{CJK}} \newline \newline \textcolor{red}{Translation results:} \newline \textcolor{red}{1.Allow us a moment of civility.} \newline \textcolor{red}{2.We must talk.} \newline \textcolor{red}{...} \newline \textcolor{red}{34.The Arcane stirs within you.} \newline \textcolor{red}{35.Stay away from the Hexgates.} \newline }\\
\texttt{<TASK> \newline According to the requirements, complete the following translation. \newline Original text: \newline 1.\begin{CJK}{UTF8}{gkai}你的病情我都听说了\end{CJK} \newline 2.\begin{CJK}{UTF8}{gkai}我真的很遗憾 孩子\end{CJK} \newline ... \newline 34.\begin{CJK}{UTF8}{gkai}令人心痛的是 越是光芒万丈的人\end{CJK} \newline 35.\begin{CJK}{UTF8}{gkai}越会更快地燃尽生命的能量\end{CJK} \newline \newline Translation results:}
\end{tcolorbox}

\begin{tcolorbox}[
    title={The \texttt{zh}$\Rightarrow$\texttt{en} translation response demonstration of SFT model.},
    colback=blue!5!white,       % 背景：极淡的蓝色
    colframe=blue!75!black,     % 边框：深蓝色
    coltitle=white,             % 标题文字：白色
    breakable,
    fontupper=\small
]
\texttt{1.I heard your prognosis. \newline 2.I’m terribly sorry, my boy. \newline ... \newline 34.It’s a sad truth that \newline 35.those who shine brightest often burn fastest.}
\end{tcolorbox}

\subsection{Input of Evaluation LLM}
\label{sec:inputjudge}

We present the prompt format for LLM evaluation of vividness in \texttt{zh}$\Rightarrow$\texttt{en} translation below (similar formats apply to other languages). When using LLM to assess the vividness of a line, we provide the contextual lines surrounding it.

\begin{tcolorbox}[
    title={The \texttt{zh}$\Rightarrow$\texttt{en} prompt demonstration of evaluation LLM.},
    colback=blue!5!white,       % 背景：极淡的蓝色
    colframe=blue!75!black,     % 边框：深蓝色
    coltitle=white,             % 标题文字：白色
    breakable,
    fontupper=\small
]
\texttt{<INSTRUCTION> \newline Please assign a vividness score (0-100, integer) to multiple English translations of a Chinese subtitle line. The original Chinese line and its surrounding context will be provided for reference, marked with [To be evaluated] and [Context] respectively. The scoring principles for translation vividness are as follows: \newline Principle 1 (Accuracy): The translation can be moderately liberal but must maintain reasonable accuracy in conveying the original meaning without additions or omissions (Weight:30\%); \newline Principle 2 (Colloquial Appropriateness): Evaluate whether the translation uses natural spoken language suitable for the subtitle context and effectively conveys the original emotion, atmosphere, and tone (Weight:30\%); \newline Principle 3 (Expressive Power): Assess whether the translation is emotionally resonant, employs vivid phrasing, and has literary qualities that engage the audience (Weight:40\%). \newline \newline Scoring Criteria (Vividness): \newline 100: Exceptionally expressive and impactful translation with rich emotional layers that deeply resonate with the audience. \newline 50: Moderately expressive translation with subdued emotional delivery that partially conveys the intended feelings. \newline 0: Inaccurate translation lacking emotional depth or expressive qualities, failing to evoke any audience connection. \newline }\\
\texttt{<EXAMPLE> \newline Chinese original text: \newline [Context] \begin{CJK}{UTF8}{gkai}被你到处践踏的遗迹 可是无价之宝\end{CJK} \newline [Context] \begin{CJK}{UTF8}{gkai}是无法用价值来衡量的珍宝\end{CJK} \newline [To be evaluated] \begin{CJK}{UTF8}{gkai}历史虽然会重演 但是人类是无法回到过去的\end{CJK} \newline [Context] \begin{CJK}{UTF8}{gkai}看来你是不会明白的\end{CJK} \newline [Context] \begin{CJK}{UTF8}{gkai}明...明白...我明白了\end{CJK} \newline \newline English translation: \newline Translation A: History repeats, but we can’t go back to what was. \newline \newline Translation B: History often echoes, yet there’s no way for us to turn back the clock. \newline \newline ... \newline \newline Evaluation score: \newline \{"A": 70, "B": 92, "C": 85, ...\} \newline }\\
\texttt{<TASK> \newline According to the criterion, complete the following evaluation. \newline Chinese original text: \newline [Context] \begin{CJK}{UTF8}{gkai}我的贡献注定是转瞬即逝 连您都会很快遗忘\end{CJK} \newline [Context] \begin{CJK}{UTF8}{gkai}我见过的学生很多\end{CJK} \newline [To be evaluated] \begin{CJK}{UTF8}{gkai}令人心痛的是 越是光芒万丈的人 越会更快地燃尽生命的能量\end{CJK} \newline [Context] \begin{CJK}{UTF8}{gkai}我不知道你还是个艺术家\end{CJK} \newline [Context] \begin{CJK}{UTF8}{gkai}我的事你不知道的多着呢\end{CJK} \newline \newline English translation: \newline Translation A: It's a sad truth that those who shine brightest often burn fastest. \newline \newline Translation B: The heartbreaking truth is that those who shine the brightest tend to burn through their life's energy all the faster. \newline \newline ... \newline \newline Note, you need to output the ratings in JSON format: \{"A": score, "B": score, "C": score, ...\} \newline Evaluation score (only output the rating, no other content): }
\end{tcolorbox}

\subsection{Instruction and Prompt for Quality Evaluation}
\label{sec:ioqe}

For human evaluation of translation quality, it is essential to provide evaluators with clear instructions specifying the evaluation perspectives, criteria, and format. These instructions directly influence the focus and emphasis of evaluators during the quality assessment process. The instructions provided to evaluators are shown in the first text box, while the prompts used for multidimensional automated evaluation with LLMs are presented in the three subsequent boxes.

\begin{tcolorbox}[
    title={Instruction for human evaluation of translation quality.},
    colback=blue!5!white,       % 背景：极淡的蓝色
    colframe=blue!75!black,     % 边框：深蓝色
    coltitle=white,             % 标题文字：白色
    breakable,
    fontupper=\small
]
\texttt{\textbf{[Evaluation Criteria]} \newline \newline \textbf{1. Accuracy} \newline When evaluateing the accuracy of audiovisual subtitle translation, consider the following dimensions: \newline \hspace*{0.4cm} \textbullet \textbf{Semantic Equivalence}: Evaluate whether the meaning of the original subtitle is accurately conveyed in the translated version, and if the semantic content of the source subtitle is precisely expressed in the target subtitle. \newline \hspace*{0.4cm} \textbullet \textbf{Grammatical Correctness}: Evaluate the grammatical accuracy of the translated subtitle, including sentence structure, tense, voice, and other grammatical aspects. \newline \hspace*{0.4cm} \textbullet \textbf{Terminology Translation}: Evaluate whether proper nouns have been accurately translated, maintaining the semantics and context of the original terms. \newline \textbf{2. Naturalness} \newline When evaluateing the naturalness of audiovisual subtitle translation, consider the following dimensions: \newline \hspace*{0.4cm} \textbullet \textbf{Coherence}: Evaluate whether the translated subtitle reads as if written by a native speaker of the target language, and evaluate the logical relationships between sentences. \newline \hspace*{0.4cm} \textbullet \textbf{Readability}: Evaluate whether the translated subtitle is easy to read and understand, and if the word choice and expressions conform to the conventions of the target language. \newline \hspace*{0.4cm} \textbullet \textbf{Fluency}: Evaluate whether the translated subtitle flows smoothly, if sentences are well-constructed, and if there are any obvious grammatical errors or unnatural expressions. \newline \textbf{3. Vividness} \newline When evaluateing the vividness of audiovisual subtitle translation, consider the following dimensions:  \newline \hspace*{0.4cm} \textbullet \textbf{Stylistic Consistency}: Evaluate whether the translated subtitle maintains the style and characteristics of the original, including consistency in character tone and emotional nuances. \newline \hspace*{0.4cm} \textbullet \textbf{Expressiveness}: Evaluate whether the translation conveys the essence and atmosphere of the original lines, avoiding mechanical literal translation, thereby making it easier for the audience to understand and find engaging. \newline \hspace*{0.4cm} \textbullet \textbf{Emotion}: Assess whether the translation faithfully conveys the character's emotions, aligns with the scene and character context, and resonates emotionally with the target language audience. \newline }\\
\texttt{\textbf{[Task]} \newline \newline For each set of original subtitles, two different translations (A and B) are provided. Please refer to the multiple evaluation dimensions specified in the [Evaluation Criteria] to evaluate the two translations for each set of original subtitles. Indicate your evaluation results for translations A and B by marking [A is better], [B is better], or [No significant difference between A and B]. Note that you only need to evaluate the overall performance of each set of subtitles, not each individual line of subtitle.}
\end{tcolorbox}

\begin{tcolorbox}[
    title={The prompt of \texttt{en}$\Rightarrow$\texttt{zh} subtitle translation accuracy evaluation.},
    colback=blue!5!white,       % 背景：极淡的蓝色
    colframe=blue!75!black,     % 边框：深蓝色
    coltitle=white,             % 标题文字：白色
    breakable,
    fontupper=\small
]
\texttt{[English to Chinese Subtitle Translation Accuracy Evaluation] \newline \newline Please rate the accuracy of the following English to Chinese subtitle translation, using integer scores from 0 to 100. The translation accuracy rating criteria include evaluating whether the Chinese translation accurately conveys the original meaning of the English subtitle. Additionally, pay attention to whether the terminology (e.g., names of people, places, organizations, items, etc.) in the English original subtitle is accurately translated in the Chinese translation. \newline \newline Scoring Criteria (Accuracy): \newline 100: The translation is completely accurate, covers all information, provides a coherent translation, and the proper nouns are accurately translated. \newline 50: The translation is mostly accurate, with only minor omissions or unclear context, and proper nouns are partially inaccurately translated, but the overall meaning is still understandable. \newline 0: The translation is severely distorted, misinterprets the main meaning of the original text, and the translation of proper nouns is poor. \newline }\\
\texttt{Original English Line: \newline A crime like this can't be overlooked. The boy must be punished. \newline A violation of the Ethos calls for banishment, \newline but I can sympathize with a young man's dream to change the world. \newline ... \newline Yeah, I must admit, his theory intrigues. \newline If dangerous ideas didn't excite the imagination, \newline we would never wander astray. \newline  \newline Chinese Translation: \newline \begin{CJK}{UTF8}{gkai}这样严重的罪行不能轻易放过\end{CJK} \newline \begin{CJK}{UTF8}{gkai}违反社会共识的人确实应该遭到驱逐\end{CJK} \newline \begin{CJK}{UTF8}{gkai}但我也能体会一个年轻人梦想改变世界的雄心\end{CJK} \newline ... \newline \begin{CJK}{UTF8}{gkai}我不得不说 他那套学说非常有意思\end{CJK} \newline \begin{CJK}{UTF8}{gkai}如果危险的念头不曾引人遐想\end{CJK} \newline \begin{CJK}{UTF8}{gkai}也就没有误入歧途一说了\end{CJK} \newline \newline Note: You need to output in JSON format: \{"Score": evaluation score\} \newline Score (only output the score, no further explanation required):}
\end{tcolorbox}

\begin{tcolorbox}[
    title={The prompt of \texttt{en}$\Rightarrow$\texttt{zh} subtitle translation naturalness evaluation.},
    colback=blue!5!white,       % 背景：极淡的蓝色
    colframe=blue!75!black,     % 边框：深蓝色
    coltitle=white,             % 标题文字：白色
    breakable,
    fontupper=\small
]
\texttt{[English to Chinese Subtitle Translation Naturalness Evaluation] \newline \newline Please rate the naturalness of the following English to Chinese subtitle translation using an integer score from 0 to 100. The naturalness score of the subtitle translation should consider whether the translated text adequately takes into account contextual factors, including cultural background and context, and whether it ensures natural and fluent language expression that conforms to Chinese grammatical structures and word usage habits. It should be easy to understand and close to the culture and expression habits of the Chinese audience. \newline \newline Scoring Criteria (Naturalness): \newline 100: The translation fully considers context and cultural background, with smooth and natural language that aligns with Chinese usage habits, and contains no grammatical or lexical errors. \newline 50: The translation considers the context and is basically fluent, but some expressions may be slightly awkward or unnatural, with potential minor grammatical errors. \newline 0: The translation does not effectively consider the context or culture, is not fluent, has many grammatical errors, is rigid, and difficult to understand. \newline }\\
\texttt{Original English Line: \newline A crime like this can't be overlooked. The boy must be punished. \newline A violation of the Ethos calls for banishment, \newline but I can sympathize with a young man's dream to change the world. \newline ... \newline Yeah, I must admit, his theory intrigues. \newline If dangerous ideas didn't excite the imagination, \newline we would never wander astray. \newline  \newline Chinese Translation: \newline \begin{CJK}{UTF8}{gkai}这样严重的罪行不能轻易放过\end{CJK} \newline \begin{CJK}{UTF8}{gkai}违反社会共识的人确实应该遭到驱逐\end{CJK} \newline \begin{CJK}{UTF8}{gkai}但我也能体会一个年轻人梦想改变世界的雄心\end{CJK} \newline ... \newline \begin{CJK}{UTF8}{gkai}我不得不说 他那套学说非常有意思\end{CJK} \newline \begin{CJK}{UTF8}{gkai}如果危险的念头不曾引人遐想\end{CJK} \newline \begin{CJK}{UTF8}{gkai}也就没有误入歧途一说了\end{CJK} \newline \newline Note: You need to output in JSON format: \{"Score": evaluation score\} \newline Score (only output the score, no further explanation required):}
\end{tcolorbox}

\begin{tcolorbox}[
    title={The prompt of \texttt{en}$\Rightarrow$\texttt{zh} subtitle translation vividness evaluation.},
    colback=blue!5!white,       % 背景：极淡的蓝色
    colframe=blue!75!black,     % 边框：深蓝色
    coltitle=white,             % 标题文字：白色
    breakable,
    fontupper=\small
]
\texttt{[English to Chinese Subtitle Translation Vividness Evaluation] \newline \newline Please rate the vividness of the English to Chinese subtitle translation below, using an integer score from 0 to 100. The score for translation vividness does not take into account the accuracy of the translation; it only evaluates whether the translation is expressive, emotionally rich, and more capable of engaging the audience. \newline \newline Scoring Criteria (Vividness): \newline 100: The translation is highly expressive and impactful, with rich emotions, capable of strongly moving the audience. \newline 50: The translation has some expressiveness, and the emotional expression is relatively flat but still conveys some emotion. \newline 0: The translation lacks expressiveness and emotion, failing to evoke any emotional resonance from the audience. \newline }\\
\texttt{Original English Line: \newline A crime like this can't be overlooked. The boy must be punished. \newline A violation of the Ethos calls for banishment, \newline but I can sympathize with a young man's dream to change the world. \newline ... \newline Yeah, I must admit, his theory intrigues. \newline If dangerous ideas didn't excite the imagination, \newline we would never wander astray. \newline  \newline Chinese Translation: \newline \begin{CJK}{UTF8}{gkai}这样严重的罪行不能轻易放过\end{CJK} \newline \begin{CJK}{UTF8}{gkai}违反社会共识的人确实应该遭到驱逐\end{CJK} \newline \begin{CJK}{UTF8}{gkai}但我也能体会一个年轻人梦想改变世界的雄心\end{CJK} \newline ... \newline \begin{CJK}{UTF8}{gkai}我不得不说 他那套学说非常有意思\end{CJK} \newline \begin{CJK}{UTF8}{gkai}如果危险的念头不曾引人遐想\end{CJK} \newline \begin{CJK}{UTF8}{gkai}也就没有误入歧途一说了\end{CJK} \newline \newline Note: You need to output in JSON format: \{"Score": evaluation score\} \newline Score (only output the score, no further explanation required):}
\end{tcolorbox}

\end{document}